\documentclass[11pt, a4paper, logo, copyright, nonumbering]{deepseek}
\usepackage[authoryear, sort&compress, round]{natbib}
\usepackage{dblfloatfix}
\usepackage{ulem}
\usepackage{caption}
\usepackage{dramatist}
\usepackage{xspace}
\usepackage{pifont} % http://ctan.org/pkg/pifont
\usepackage{multirow}
\usepackage{tcolorbox}
\usepackage{xltabular}
\usepackage{diagbox}
\usepackage{longtable}
\usepackage{hyperref}
\usepackage{algorithm}
\usepackage{algpseudocode}
\usepackage{amsfonts}
\usepackage{amsmath}
\usepackage{amssymb}
\usepackage{lineno}
\usepackage{multirow}
\usepackage{adjustbox}
\usepackage{listings}
\usepackage[bottom]{footmisc}
\usepackage{CJKutf8}
\usepackage{setspace}

\usepackage{dsfont}
\usepackage{array}  % 用于tabularx环境
\usepackage{tabularx}  % 用于tabularx环境
\usepackage{subfigure}  % 用于创建子图
\usepackage{xcolor}  % 用于画有颜色的框
\usepackage{listings}

\usepackage{lipsum}  % 用于生成示例文本
\usepackage{multicol} % 引入 multicol 宏包

\makeatletter
\def\@BTrule[#1]{%
  \ifx\longtable\undefined
    \let\@BTswitch\@BTnormal
  \else\ifx\hline\LT@hline
    \nobreak
    \let\@BTswitch\@BLTrule
  \else
     \let\@BTswitch\@BTnormal
  \fi\fi
  \global\@thisrulewidth=#1\relax
  \ifnum\@thisruleclass=\tw@\vskip\@aboverulesep\else
  \ifnum\@lastruleclass=\z@\vskip\@aboverulesep\else
  \ifnum\@lastruleclass=\@ne\vskip\doublerulesep\fi\fi\fi
  \@BTswitch}
\makeatother

\addto\extrasenglish{
}

 {\begin{list}{}%
         {\setlength{\leftmargin}{#1}}%
         \item[]%
 }
 {\end{list}}
 
\reportnumber{001} % Leave blank if n/a

\renewcommand{\today}{}
\newcommand{\dsmoe}{DeepSeekMoE}

\newcommand{\dsviii}{DeepSeek-V3}
\newcommand{\dsviiiII}{DeepSeek-V3.2}
\newcommand{\dsviv}{DeepSeek-V4}
\newcommand{\dsvivp}{DeepSeek-V4-Pro}
\newcommand{\dsvivpm}{DeepSeek-V4-Pro-Max}

\newcommand{\dsvivf}{DeepSeek-V4-Flash}
\newcommand{\dsvivfm}{DeepSeek-V4-Flash-Max}

\newcommand{\dsvivptp}{1.6T}
\newcommand{\dsvivpap}{49B}
\newcommand{\dsvivptoken}{33T}
\newcommand{\dsvivftp}{284B}
\newcommand{\dsvivfap}{13B}
\newcommand{\dsvivftoken}{32T}
\newcommand{\mhcfull}{Manifold-Constrained Hyper-Connections}
\newcommand{\mhc}{\textit{m}HC}
\newcommand{\hcafull}{Heavily Compressed Attention}
\newcommand{\hca}{HCA}
\newcommand{\csafull}{Compressed Sparse Attention}
\newcommand{\csa}{CSA}
\newcommand{\dsafull}{DeepSeek Sparse Attention}
\newcommand{\dsa}{DSA}

\title{\centering \dsviv{}: \\ Towards Highly Efficient Million-Token Context Intelligence}

\author[*]{
\vspace{-0.8cm}
DeepSeek-AI
\\
\small
\vspace{-0.3cm}
\texttt{research@deepseek.com}
\vspace{-0.9cm}
}

\renewcommand{\phi}{\varphi}

\renewcommand{\leq}{\leqslant}
\renewcommand{\geq}{\geqslant}

\renewcommand{\epsilon}{\varepsilon}
\renewcommand{\imath}{\mathrm{i}}

\newlength{\restsubwidth}
\newlength{\restsubheight}
\newlength{\restsubmoreheight}
\newcommand{\rest}[2]{%
        \settowidth{\restsubwidth}{\ensuremath{#2}}
        \settoheight{\restsubheight}{\ensuremath{{}_{#2}}}
        \ensuremath{{#1\hskip 0.5pt}_{\vrule\kern2pt\parbox[b][%
        4pt][b]{\the\restsubwidth}{%
                        \ensuremath{{}_{#2}}}}}
        }

\begin{abstract}
\vspace{-0.5cm}

We present a preview version of \dsviv{} series, including two strong Mixture-of-Experts~(MoE) language models --- \dsvivp{} with \dsvivptp{} parameters (\dsvivpap{} activated) and \dsvivf{} with \dsvivftp{} parameters (\dsvivfap{} activated) --- both supporting a context length of one million tokens.
\dsviv{} series incorporate several key upgrades in architecture and optimization: 
(1) a hybrid attention architecture that combines \csafull{}~(\csa{}) and \hcafull{}~(\hca{}) to improve long-context efficiency; 
(2) \mhcfull{} (\mhc{}) that enhance conventional residual connections;
(3) and the Muon optimizer for faster convergence and greater training stability.
We pre-train both models on more than 32T diverse and high-quality tokens, followed by a comprehensive post-training pipeline that unlocks and further enhances their capabilities.
\dsvivpm{}, the maximum reasoning effort mode of \dsvivp{}, redefines the state-of-the-art for open models, outperforming its predecessors in core tasks.
Meanwhile, \dsviv{} series are highly efficient in long-context scenarios. 
In the one-million-token context setting, \dsvivp{} requires only 27\% of single-token inference FLOPs and 10\% of KV cache compared with \dsviiiII{}. 
This enables us to routinely support one-million-token contexts, thereby making long-horizon tasks and further test-time scaling more feasible.
The model checkpoints are available at \url{https://huggingface.co/collections/deepseek-ai/deepseek-v4}. 

\end{abstract}

\begin{document}
\begin{CJK*}{UTF8}{gbsn}

\maketitle

\begin{figure}[h]
\centering
\includegraphics[width=1.0\textwidth]{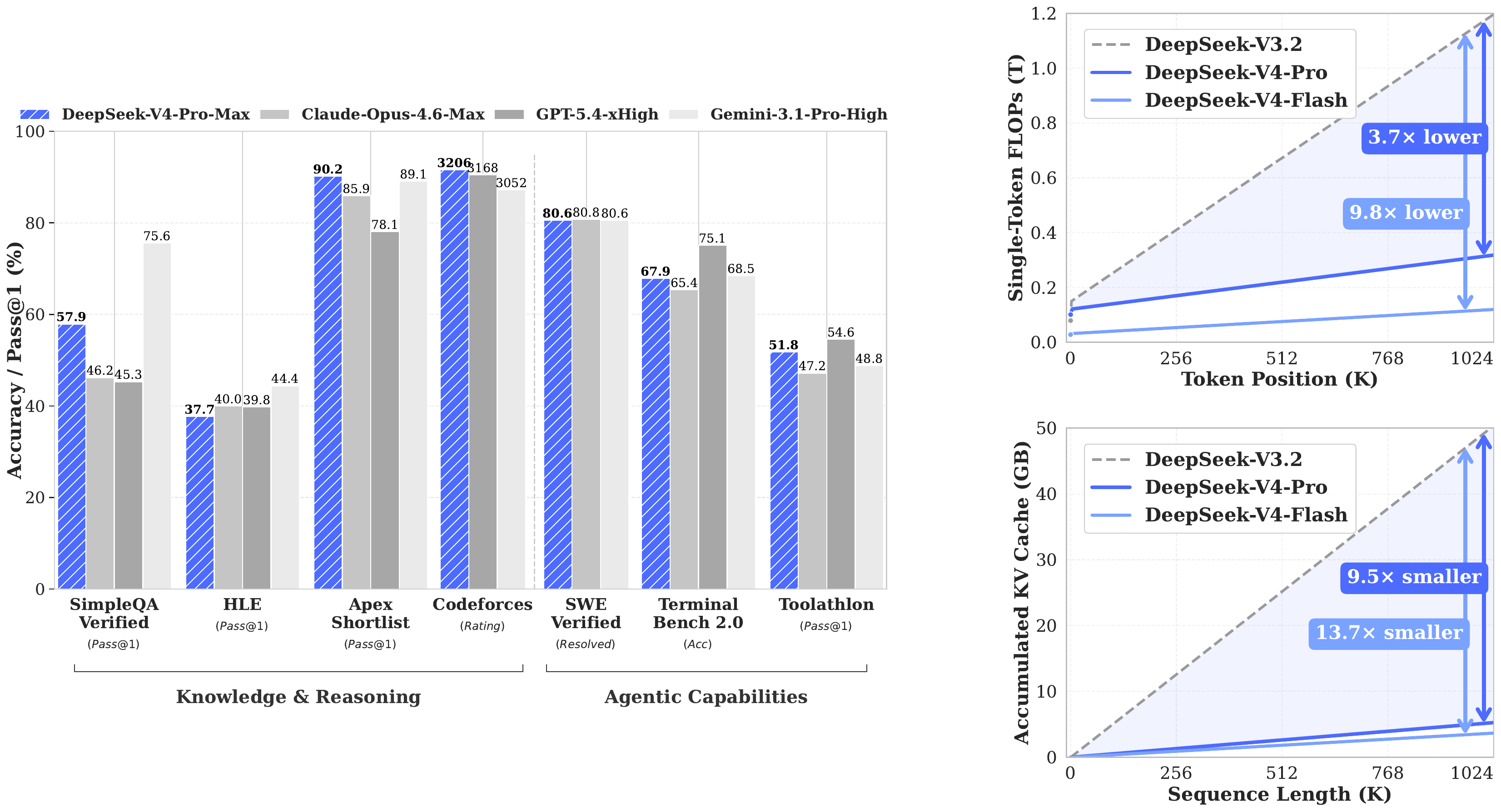}
\caption{
    \textbf{Left}: benchmark performance of \dsvivpm{} and its counterparts.
    \textbf{Right}: inference FLOPs and KV cache size of \dsviv{} series and \dsviiiII. 
}
\label{fig:dsv4_performance_and_cost}
\end{figure}

\newpage

\begin{spacing}{0.9}
\tableofcontents
\end{spacing}

\newpage

\section{Introduction}

% background
The emergence of reasoning models~\citep{o1,dsr1} has established a new paradigm of test-time scaling, driving substantial performance gains for Large Language Models~(LLMs). 
However, this scaling paradigm is fundamentally constrained by the quadratic computational complexity of the vanilla attention mechanism~\citep{transformer}, which creates a prohibitive bottleneck for ultra-long contexts and reasoning processes. 
Concurrently, the emergence of long-horizon scenarios and tasks --- from complex agentic workflows to massive cross-document analysis --- has also made efficient support for ultra-long contexts critical for future progress. 
While recent open-source efforts~\citep{dsv3,qwen3,minimax_m2,kimi_k2} have advanced general capabilities, this core architectural inefficiency in handling ultra-long sequences remains a key impediment, limiting further gains from test-time scaling and hindering further exploration into long-horizon scenarios and tasks.

% brief intro to dsv4
In order to break the efficiency barrier in ultra-long contexts, we develop the \dsviv{} series, including the preview versions of \dsvivp{} with \dsvivptp{} parameters (\dsvivpap{} activated) and \dsvivf{} with \dsvivftp{} parameters (\dsvivfap{} activated). 
Through architectural innovations, \dsviv{} series achieve a dramatic leap in computational efficiency for processing ultra-long sequences. 
This breakthrough enables efficient support for a context length of one million tokens, ushering in a new era of million-length contexts for next-generation LLMs.
We believe our capability to efficiently handle ultra-long sequences unlocks the next frontier of test-time scaling, paves the way for deeper research into long-horizon tasks, and establishes a necessary foundation for exploring future paradigms like online learning.

% brief intro to arch
Compared with the \dsviii{} architecture~\citep{dsv3}, \dsviv{} series retain the DeepSeekMoE framework~\citep{deepseekmoe} and Multi-Token Prediction (MTP) strategy, while introducing several key innovations in architecture and optimization.  
To enhance long-context efficiency, we design a hybrid attention mechanism combining \csafull{}~(\csa{}) and \hcafull{}~(\hca{}). 
\csa{} compresses the KV caches along the sequence dimension and then performs DeepSeek Sparse Attention (DSA)~\citep{dsv32}, whereas \hca{} applies more aggressive compression to the KV caches but keeps dense attention.
To strengthen modeling capability, we incorporate \mhcfull{} (\mhc{})~\citep{mhc} that upgrade conventional residual connections.  
Additionally, we introduce the Muon~\citep{muon,muon_kimi} optimizer to the training of \dsviv{} series, leading to faster convergence and improved training stability.

% brief intro to infra
To enable efficient training and inference for \dsviv{} series as well as productive development, we introduce several infrastructure optimizations.
First, we design and implement a single fused kernel for MoE modules that fully overlaps computation, communication, and memory access. 
Second, we employ TileLang~\citep{wangtilelang}, a Domain-Specific Language (DSL) to balance development productivity and runtime efficiency. 
Third, we provide efficient batch-invariant and deterministic kernel libraries to ensure bitwise reproducibility across training and inference. 
Fourth, for the training framework, we extend the autograd framework with tensor-level checkpointing for fine-grained recomputation control; and we enhance training efficiency with a hybrid ZeRO strategy for the Muon optimizer, cost-effective \mhc{} implementations via recomputation and fused kernels, and two-stage contextual parallelism to manage compressed attention.
Fifth, for the inference framework, we design a heterogeneous KV cache structure with on-disk storage strategies to enable efficient shared-prefix reuse.
In addition, during the post-training stage, we incorporate FP4 quantization-aware training for MoE expert weights and the indexer QK path to reduce memory and computation. 

% efficiency and cost
By employing hybrid \csa{} and \hca{}, along with precision optimizations on computation and storage, \dsviv{} series achieve significantly lower inference FLOPs and a substantially reduced KV cache size compared with \dsviiiII{}, especially in long-context settings. 
The right part of Figure~\ref{fig:dsv4_performance_and_cost} demonstrates the estimated single-token inference FLOPs and accumulated KV cache size of \dsviiiII{} and \dsviv{} series.
In the scenario of 1M-token context, even \dsvivp{}, which has a larger number of activated parameters, attains only 27\% of the single-token FLOPs (measured in equivalent FP8 FLOPs) and 10\% of the KV cache size relative to \dsviiiII{}. 
Furthermore, \dsvivf{}, with its smaller number of activated parameters, pushes efficiency even further: in the 1M-token context setting, it achieves only 10\% of the single-token FLOPs and 7\% of the KV cache size compared with \dsviiiII{}. 
Additionally, for \dsviv{} series, the routed expert parameters utilize FP4 precision. 
While the peak FLOPs for FP4 $\times$ FP8 operations are currently the same as FP8 $\times$ FP8 on existing hardware, they can theoretically be implemented to be $1/3$ more efficient on future hardware, which will further enhance the efficiency of \dsviv{} series.

% brief intro to pre-training
During pre-training, we train \dsvivf{} on \dsvivftoken{} tokens and \dsvivp{} on \dsvivptoken{} tokens, respectively. 
After pre-training, these two models can natively and efficiently support 1M-length contexts. 
In our internal evaluations, \dsvivf{}-Base already surpasses \dsviiiII{}-Base across a majority of benchmarks with its more parameter-efficient design. 
\dsvivp{}-Base further extends this advantage to set a new performance standard among DeepSeek foundation models, achieving comprehensive superiority across reasoning, coding, long-context, and world knowledge tasks.

% brief intro to post-training
The post-training pipeline of \dsviv{} series features a two-stage paradigm: the independent cultivation of domain-specific experts, followed by unified model consolidation via on-policy distillation~\citep{lu2025onpolicydistillation,minillm}. 
Initially, for each target domain --- such as mathematics, coding, agent, and instruction following --- a separate expert model is trained independently. 
The base model first undergoes Supervised Fine-Tuning (SFT) on high-quality, domain-specific data to establish foundational capabilities. 
Subsequently, Reinforcement Learning (RL) is applied using Group Relative Policy Optimization (GRPO)~\citep{dsr1}, which further optimizes the model for domain-aligned behaviors guided by reward models tailored to specific success criteria. 
This phase yields a diverse set of specialized experts, each excelling in its respective field. 
Finally, to integrate these distinct proficiencies, a single unified model is trained through on-policy distillation, wherein the unified model acts as the student learning to optimize the reverse KL loss with teacher models.

\noindent
\textbf{Summary of Core Evaluation Results}
\begin{itemize}[topsep=0pt]
    \item \textbf{Knowledge}:
    In assessments of broad world knowledge, \dsvivpm{}, the maximum reasoning effort mode of \dsvivp{}, significantly outperforms leading open-source models on the SimpleQA~\citep{simpleqa} and Chinese-SimpleQA~\citep{csimpleqa} benchmarks. 
    Regarding educational knowledge --- evaluated via MMLU-Pro~\citep{mmlu_pro}, HLE~\citep{hle}, and GPQA~\citep{gpqa} --- \dsvivpm{} shows a marginal lead over its open-source counterparts. 
    \dsvivpm{} has significantly closed the gap with the leading proprietary model, Gemini-3.1-Pro, despite still trailing it in these knowledge-based evaluations.
    
    \item \textbf{Reasoning}: 
    Through the expansion of reasoning tokens, \dsvivpm{} demonstrates superior performance relative to GPT-5.2 and Gemini-3.0-Pro on standard reasoning benchmarks. 
    Nevertheless, its performance falls marginally short of GPT-5.4 and Gemini-3.1-Pro, suggesting a developmental trajectory that trails state-of-the-art frontier models by approximately 3 to 6 months.
    Furthermore, \dsvivfm{} achieves comparable performance to GPT-5.2 and Gemini-3.0-Pro, establishing itself as a highly cost-effective architecture for complex reasoning tasks.

    \item \textbf{Agent}: 
    On public benchmarks, \dsvivpm{} is on par with leading open-source models, such as Kimi-K2.6 and GLM-5.1, but slightly worse than frontier closed models. 
    In our internal evaluation, \dsvivpm{} outperforms Claude Sonnet 4.5 and approaches the level of Opus 4.5. 

    \item \textbf{Long-Context}: \dsvivpm{} delivers strong results on synthetic and real use cases with a 1-million-token context window, surpassing even Gemini-3.1-Pro on academic benchmarks.

    \item \textbf{\dsvivp{} v.s. \dsvivf{}}: 
    \dsvivfm{} exhibits lower performance in knowledge evaluations due to its smaller parameter scale. 
    However, it achieves comparable results on reasoning tasks when allocated a larger thinking budget. 
    In agent evaluations, while \dsvivfm{} matches the performance of \dsvivpm{} on several benchmarks, it still trails its larger counterpart on more complex, high-difficulty tasks.

\end{itemize}

\begin{figure}[t]
\centering
\includegraphics[width=0.6\linewidth]{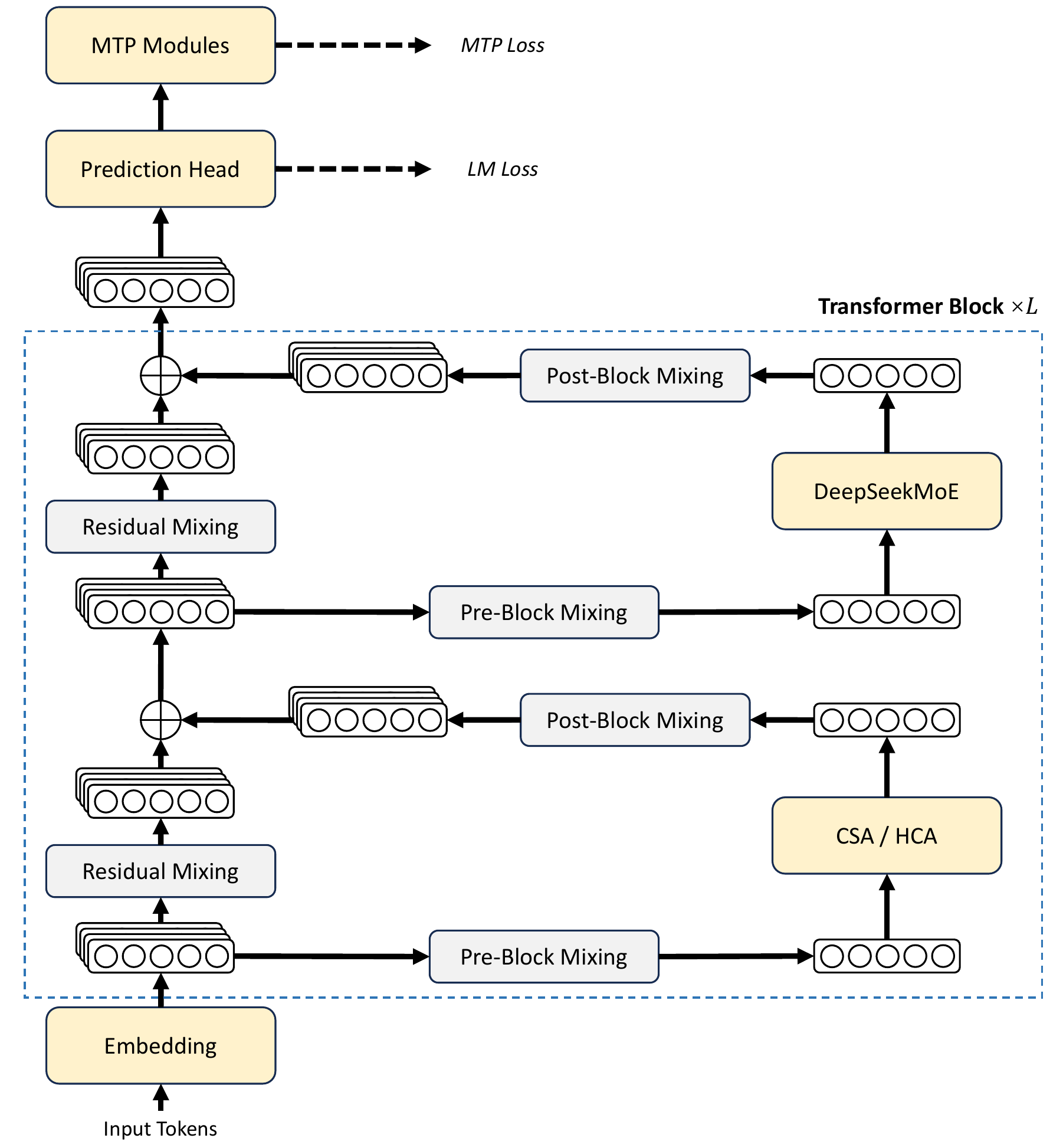}
\caption{
    Overall architecture of \dsviv{} series. 
    We use hybrid \csa{} (\csafull{}) and \hca{} (\hcafull{}) for attention layers, DeepSeekMoE for feed-forward layers, and strengthen conventional residual connections with \mhc{}.
}
\label{fig:dsv4_arch}
\end{figure}

\section{Architecture}
\label{sec:arch}

Overall, \dsviv{} series retain the Transformer~\citep{transformer} architecture and Multi-Token Prediction (MTP) modules~\citep{meta_mtp,dsv3}, while introducing several key upgrades over \dsviii{}:
(1) firstly, we introduce the \mhcfull{} (\mhc{})~\citep{mhc} to strengthen conventional residual connections; 
(2) secondly, we design a hybrid attention architecture, which greatly improves long-context efficiency through \csafull{} and \hcafull{}. 
(3) thirdly, we employ Muon~\citep{muon,muon_kimi} as the optimizer.
For the Mixture-of-Experts~(MoE) components, we still adopt the \dsmoe{}~\citep{deepseekmoe} architecture, with only minor adjustments from \dsviii{}.
The Multi-Token Prediction (MTP)~\citep{ms_mtp,meta_mtp,eagle,dsv3} configuration remains identical to that of \dsviii{}. 
All other unspecified details follow the settings established in \dsviii{}~\citep{dsv3}. 
Figure~\ref{fig:dsv4_arch} illustrates the overall architecture of \dsviv{}, and the details are described below.

\subsection{Designs Inherited from \dsviii{}}

\paragraph{Mixture-of-Experts.}

As previous DeepSeek-series models~\citep{dsvii,dsv3}, \dsviv{} series also adopt the \dsmoe{} paradigm~\citep{deepseekmoe} for Feed-Forward Networks~(FFNs), which sets fine-grained routed experts and shared experts.
Different from \dsviii{}, we change the activation function that computes the affinity scores from $\operatorname{Sigmoid}(\cdot)$ into $\operatorname{Sqrt}(\operatorname{Softplus}(\cdot))$.
For load balancing, we also employ the auxiliary-loss-free strategy~\citep{noaux_tc,dsv3}, augmented by a slight sequence-wise balance loss that prevents extreme imbalance within individual sequences.
For \dsviv{}, we remove the constraint on the number of routing target nodes, and carefully redesign the parallelism strategy to maintain training efficiency.
Furthermore, compared with \dsviii{}, we replace the dense FFN layers in the initial several Transformer blocks with MoE layers that employ Hash routing~\citep{hash_layer}. 
The Hash routing strategy determines the target experts of each token according to a predefined hash function with regard to the input token ID.

\paragraph{Multi-Token Prediction.}

As \dsviii{}, \dsviv{} series also set MTP modules and objectives. 
Given that the MTP strategy has been validated in \dsviii{}, we adopt the same strategy for \dsviv{} series without modification.

\subsection{\mhcfull{}}

As shown in Figure~\ref{fig:dsv4_arch}, \dsviv{} series incorporate \mhcfull{} (\mhc{})~\citep{mhc} to strengthen the conventional residual connections between adjacent Transformer blocks. 
Compared with naive Hyper-Connections (HC)~\citep{hc}, the core idea of \mhc{} is to constrain the residual mapping onto a specific manifold, and thus enhance the stability of signal propagation across layers while preserving model expressivity.
This subsection briefly introduces the standard HC and describes how we design \mhc{} for stable training. 

\paragraph{Standard Hyper-Connections.}
The standard HC expands the width of the residual stream by a factor of $n_{\text{hc}}$. 
Specifically, the shape of the residual stream is expanded from $\mathbb{R}^{d}$ to $\mathbb{R}^{n_{\text{hc}} \times d}$, where $d$ is the hidden size of the actual layer input. 
Let $X_l = [\mathbf{x}_{l,1}; \ldots; \mathbf{x}_{l,n_{\text{hc}}}]^T \in \mathbb{R}^{n_{\text{hc}} \times d}$ be the residual state before the $l$-th layer. 
HC introduces three linear mappings: an input mapping $A_{l} \in \mathbb{R}^{1 \times n_{\text{hc}}}$, a residual transformation $B_{l} \in \mathbb{R}^{n_{\text{hc}} \times n_{\text{hc}}}$, and an output mapping $C_{l} \in \mathbb{R}^{n_{\text{hc}} \times 1}$. 
The update of the residual state is then formulated as:
\begin{equation}
    X_{l+1} = B_{l} X_l + C_{l} \mathcal{F}_{l}(A_{l} X_l),
\end{equation}
where $\mathcal{F}_{l}$ denotes the $l$-th layer (e.g., an MoE layer), whose input and output shapes are both $\mathbb{R}^{d}$.
Note that the actual layer input $A_{l} X_l \in \mathbb{R}^{d}$ is also $d$-dimensional, so the expanded residual width does not influence the design of the inner layers. 
HC decouples the residual width from the actual hidden size, offering a complementary scaling axis with minimal computational overhead, as $n_{\text{hc}}$ is typically much smaller than the hidden size $d$.
However, even though HC has demonstrated potential in improving model performance, we find that the training will frequently exhibit numerical instability when stacking multiple layers, which hinders the scaling of HC.

\paragraph{Manifold-Constrained Residual Mapping.}
The core innovation of \mhc{} is to constrain the residual mapping matrix $B_l$ to the manifold of doubly stochastic matrices (the Birkhoff polytope) $\mathcal{M}$, and thus enhance the stability of signal propagation across layers:
\begin{equation}
    B_l \in \mathcal{M} \coloneq \{ M \in \mathbb{R}^{n \times n} \mid M\mathbf{1}_n = \mathbf{1}_n, \; \mathbf{1}_n^T M = \mathbf{1}_n^T, \; M \geq 0 \}.
\end{equation}
This constraint ensures that the spectral norm of the mapping matrix $\|B_l\|_2$ is bounded by 1, so the residual transformation is non-expansive, which increases the numerical stability during both the forward pass and backpropagation.
Besides, the set $\mathcal{M}$ is closed under multiplication, which guarantees stability in the scenarios of deep stacks of \mhc{}. 
In addition, the input transformation $A_l$ and output transformation $C_l$ are also constrained to be non-negative and bounded via a Sigmoid function to avoid the risk of signal cancellation.

\paragraph{Dynamic Parameterization.}
The parameters of three linear mappings are dynamically generated, which are decomposed into a dynamic (input-dependent) component and a static (input-independent) component. 
Given the input $X_l \in \mathbb{R}^{n_{\text{hc}} \times d}$, it is first flattened and normalized: $\hat{X}_l = \operatorname{RMSNorm}(\operatorname{vec}(X_l)) \in \mathbb{R}^{1 \times n_{\text{hc}}d}$. 
Then, we follow the conventional HC to generate the unconstrained raw parameters $\tilde{A}_l \in \mathbb{R}^{1 \times n_{\text{hc}}}$, $\tilde{B}_l \in \mathbb{R}^{n_{\text{hc}} \times n_{\text{hc}}}$, and $\tilde{C}_l \in \mathbb{R}^{n_{\text{hc}} \times 1}$:
\begin{align}
    \tilde{A}_l &= \alpha_l^\mathrm{pre} \cdot (\hat{X}_l W^\mathrm{pre}_l) + S_l^\mathrm{pre}, \\
    \tilde{B}_l &= \alpha_l^\mathrm{res} \cdot \operatorname{Mat}(\hat{X}_l W^\mathrm{res}_l) + S_l^\mathrm{res}, \\
    \tilde{C}_l &= \alpha_l^\mathrm{post} \cdot (\hat{X}_l W^\mathrm{post}_l)^T + S_l^\mathrm{post},
\end{align}
where $W^\mathrm{pre}_l, W^\mathrm{post}_l \in \mathbb{R}^{n_{\text{hc}}d \times n_{\text{hc}}}$ and $W^\mathrm{res}_l \in \mathbb{R}^{n_{\text{hc}}d \times n_{\text{hc}}^2}$ are learnable parameters for generating the dynamic components;
$\operatorname{Mat}(\cdot)$ reshapes a vector of size $1 \times n_{\text{hc}}^2$ into a matrix of size $n_{\text{hc}} \times n_{\text{hc}}$; 
$S_l^\mathrm{pre} \in \mathbb{R}^{1 \times n_{\text{hc}}}$, $S_l^\mathrm{post} \in \mathbb{R}^{n_{\text{hc}}\times 1}$, and $S_l^\mathrm{res} \in \mathbb{R}^{n_{\text{hc}} \times n_{\text{hc}}}$ are learnable static biases; 
and $\alpha_l^\mathrm{pre}$, $\alpha_l^\mathrm{res}$, $\alpha_l^\mathrm{post} \in \mathbb{R}$ are learnable gating factors initialized to small values. 

\paragraph{Applying Parameter Constraints.}
After obtaining the unconstrained raw parameters $\tilde{A}_l, \tilde{B}_l, \tilde{C}_l$, we then apply constraints described earlier to them to enhance the numerical stability. 
To be specific, for the input and output mappings, we employ a Sigmoid function $\sigma(\cdot)$ to ensure their non-negativity and boundedness:
\begin{align}
    A_l &= \sigma(\tilde{A}_l), \\
    C_l &= 2\sigma(\tilde{C}_l).
\end{align}
As for the residual mapping $\tilde{B}_l$, we project it onto the manifold of doubly stochastic matrices $\mathcal{M}$. 
This is achieved by the Sinkhorn-Knopp algorithm, which first applies an exponential function to $\tilde{B}_l$ to ensure positivity, getting $M^{(0)} = \exp(\tilde{B}_l)$, and then iteratively performs column and row normalization:
\begin{equation}
    M^{(t)} = \mathcal{T}_r(\mathcal{T}_c(M^{(t-1)})),
\end{equation}
where $\mathcal{T}_r$ and $\mathcal{T}_c$ denote row and column normalization, respectively. 
This iteration converges to a constrained doubly stochastic matrix $B_l = M^{(t_{\text{max}})}$. 
We choose $t_{\text{max}} = 20$ as a practical value.

\subsection{Hybrid Attention with \csa{} and \hca{}}

As the context length reaches extreme scales, the attention mechanism emerges as the dominant computational bottleneck in a model.
For \dsviv{}, we design two efficient attention architectures --- \csafull{}~(\csa{}) and \hcafull{}~(\hca{}) --- and employ their interleaved hybrid configuration, which substantially reduces the computational cost of attention in long-text scenarios. 
\csa{} integrates both compression and sparse attention strategies: it first compresses the Key-Value~(KV) cache of every $m$ tokens into one entry, and then applies \dsafull{} (\dsa{})~\citep{dsv32} where each query token attends to only $k$ compressed KV entries.
\hca{} aims for extreme compression by consolidating the KV cache of every $m^{\prime}$ ($\gg m$) tokens into a single entry. 
The hybrid architecture of \csa{} and \hca{} remarkably improves the long-context efficiency of \dsviv{} series, making one-million-token context feasible in practice. 
This subsection describes the core techniques of our hybrid attention architecture, and we also provide an open-source implementation\footnote{\url{https://huggingface.co/deepseek-ai/DeepSeek-V4-Pro/tree/main/inference}} to specify more details unambiguously. 

\begin{figure}[t]
\centering
\includegraphics[width=0.99\linewidth]{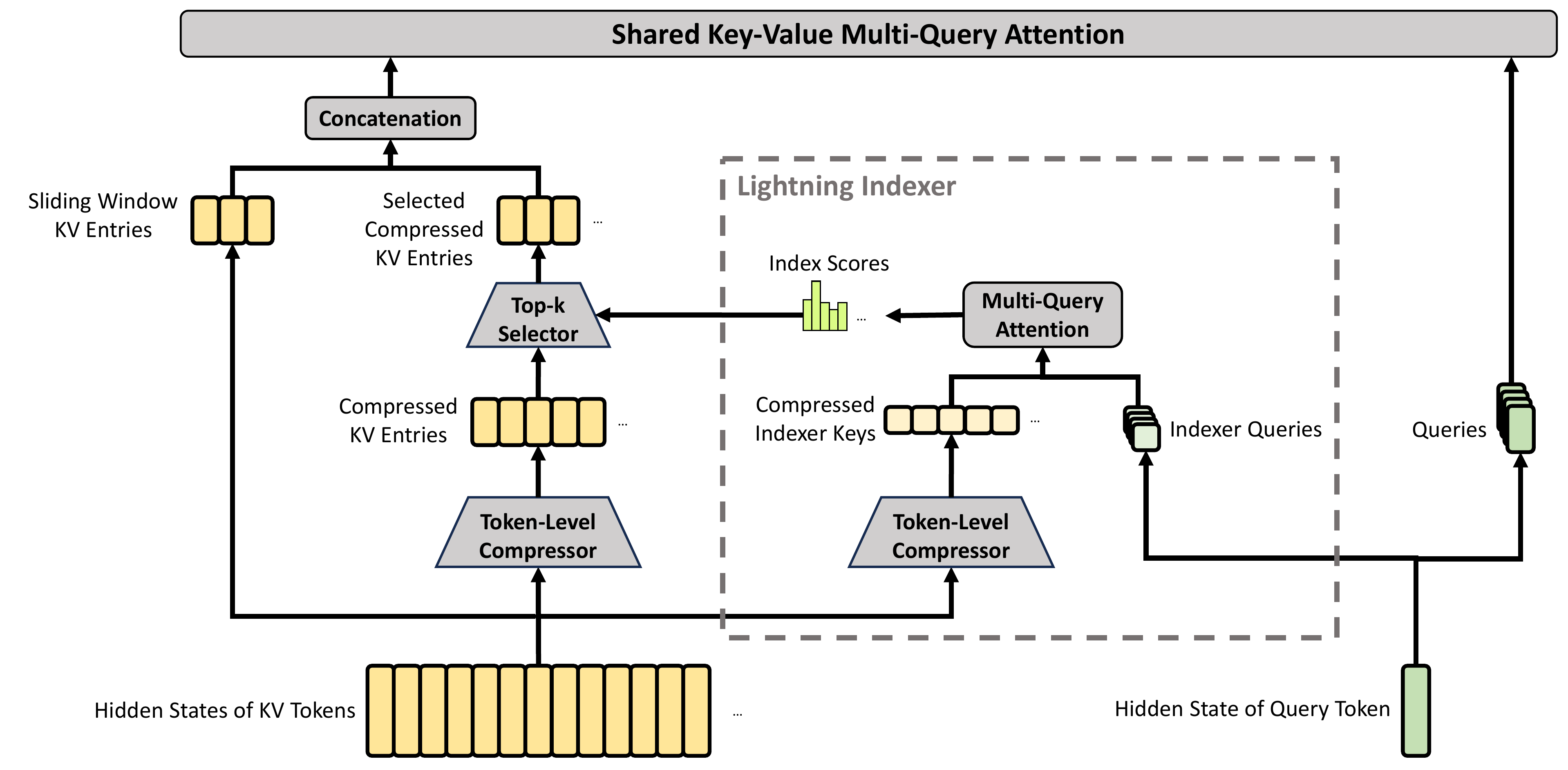}
\caption{
    Core architectures of \csa{}. 
    It compresses the number of KV entries to $\frac{1}{m}$ times, and then applies \dsafull{} for further acceleration. 
    Additionally, a small set of sliding window KV entries is combined with the selected compressed KV entries to enhance local fine-grained dependencies. 
}
\label{fig:csa}
\end{figure}

\subsubsection{\csafull{}}

The core architecture of \csa{} is illustrated in Figure~\ref{fig:csa}, which first compresses the KV cache of each $m$ tokens into one entry, and then applies \dsafull{} for further acceleration. 

\paragraph{Compressed Key-Value Entries.}
Let $H \in \mathbb{R}^{n \times d}$ be a sequence of input hidden states, where $n$ is the sequence length and $d$ is the hidden size. 
\csa{} first computes two series of KV entries $C^{a}, C^{b} \in \mathbb{R}^{n \times c}$ and their corresponding compression weights $Z^{a}, Z^{b} \in \mathbb{R}^{n \times c}$, where $c$ is the head dimension:
\begin{align}
    C^{a} &= H \cdot W^{aKV}, \quad C^{b} = H \cdot W^{bKV}, \\
    Z^{a} &= H \cdot W^{aZ}, \quad~~ Z^{b} = H \cdot W^{bZ},
\end{align}
where $W^{aKV}, W^{bKV}, W^{aZ}, W^{bZ} \in \mathbb{R}^{d \times c}$ are trainable parameters.
Next, each $m$ KV entries in $C^{a}$ and $C^{b}$ will be compressed into one entry according to their compression weights and learnable positional biases $B^a,B^b \in \mathbb{R}^{m \times c}$, producing $C^{\text{Comp}} \in \mathbb{R}^{\frac{n}{m} \times c}$. 
Each compressed entry $C^{\text{Comp}}_{i} \in \mathbb{R}^{c}$ is computed by
\begin{align}
     [S^a_{mi:m(i+1)-1};S^b_{m(i-1):mi-1}] &=  \operatorname{Softmax}_{\text{row}}([Z^{a}_{mi:m(i+1)-1} + B^a;Z^{b}_{m(i-1):mi-1} + B^b]), \\
    C^{\text{Comp}}_{i} &= \sum_{j=mi}^{m(i+1)-1} S^a_j \odot C^{a}_{j} + \sum_{j=m(i-1)}^{mi-1} S^b_j \odot C^{b}_{j},
\end{align}
where $\odot$ denotes the Hadamard product; 
$\operatorname{Softmax}_{\text{row}}(\cdot)$ denotes the softmax operation along the row dimension, which performs normalization across the total of $2m$ elements from both $Z^a$ and $Z^b$. 
When $i=0$, $Z^b_{m(i-1):mi-1}$ is padded with negative infinity and $C^b_{m(i-1):mi-1}$ is padded with zeros.
Note that each $C^{\text{Comp}}_{i}$ is derived from $2m$ KV entries, but the indexes of $C^b$ used for $C^{\text{Comp}}_{i}$ and the indexes of $C^a$ used for $C^{\text{Comp}}_{i-1}$ are overlapped.
Therefore, \csa{} in fact compresses the sequence length to $\frac{1}{m}$ times. 

\paragraph{Lightning Indexer for Sparse Selection.}
After obtaining the compressed KV entries $C^{\text{Comp}}$, \csa{} applies the \dsa{} strategy to select top-k compressed KV entries for core attention. 
First, \csa{} performs the same compression operation used for $C^{\text{Comp}}$ to get compressed indexer keys $K^{\text{IComp}} \in \mathbb{R}^{\frac{n}{m} \times c^I}$, where $c^I$ is the indexer head dimension. 
Then, for a query token $t$, we produce the indexer queries $\{\mathbf{q}_{t, 1}^{I};\mathbf{q}_{t, 2}^{I};...;\mathbf{q}_{t, n_{h}^{I}}^{I}\}$ in a low-rank manner:
\begin{align}
    \mathbf{c}_{t}^{Q} &= \mathbf{h}_{t} \cdot W^{DQ}, \\
    [\mathbf{q}_{t, 1}^{I};\mathbf{q}_{t, 2}^{I};...;\mathbf{q}_{t, n_{h}^{I}}^{I}] = \mathbf{q}_{t}^{I} &= \mathbf{c}_{t}^{Q} \cdot W^{IUQ},
\end{align}
where $\mathbf{h}_{t} \in \mathbb{R}^{d}$ is the input hidden state of the query token $t$; $\mathbf{c}_{t}^{Q} \in \mathbb{R}^{d_c}$ is the compressed latent vector for queries; $d_c$ denotes the query compression dimension; $n_{h}^{I}$ denotes the number of indexer query heads; $W^{DQ} \in \mathbb{R}^{d \times d_c}$ and $W^{IUQ} \in \mathbb{R}^{d_c \times c^I n_h^I}$ are the down-projection and up-projection matrices for indexer queries, respectively.
Next, the index score $I_{t, s} \in \mathbb{R}$ between the query token $t$ and a preceding compressed block $s$ ($s$ < $\operatorname{Floor}(\frac{t}{m})$) is computed by
\begin{align}
    [w_{t, 1}^I; w_{t, 2}^I; ...; w_{t, n_{h}^{I}}^I] = \mathbf{w}_t^I & = \mathbf{h}_{t} \cdot W^w, \\
    I_{t, s} & = \sum_{h=1}^{n_h^I} w_{t, h}^I \cdot \text{ReLU}\left(\mathbf{q}^{I}_{t, h} \cdot K^{\text{IComp}}_{s}\right),
\end{align}
where $W^w \in \mathbb{R}^{d \times n_{h}^{I}}$ is a learnable matrix; $w_{t, h}^I \in \mathbb{R}$ is the weight of the $h$-th indexer head.
For a query token $t$, given its index scores $I_{t, :}$, we employ a top-k selector to selectively retain a subset of compressed KV entries $\mathcal{C}^{\text{SprsComp}}_t$ for subsequent core attention:
\begin{equation}
    \mathcal{C}^{\text{SprsComp}}_t = \left\{ C^{\text{Comp}}_{s} ~\Big|~ I_{t, s} \in \operatorname{Top-k} (I_{t, :}) \right\}.
\end{equation}

\paragraph{Shared Key-Value MQA.}
After selecting the sparse KV entries, \csa{} then performs core attention in a Multi-Query Attention~(MQA)~\citep{mqa} manner, where each compressed KV entry in $\mathcal{C}^{\text{SprsComp}}_t$ serves as both attention key and value. 
To be specific, for a query token $t$, we first produce attention queries $\{\mathbf{q}_{t, 1};\mathbf{q}_{t, 2};...;\mathbf{q}_{t, n_{h}}\}$ from the compressed latent vector $\mathbf{c}_{t}^{Q}$:
\begin{equation}
    [\mathbf{q}_{t, 1};\mathbf{q}_{t, 2};...;\mathbf{q}_{t, n_{h}}] = \mathbf{q}_{t} = \mathbf{c}_{t}^{Q} \cdot W^{UQ},
\end{equation}
where $n_{h}$ denotes the number of query heads; $W^{UQ} \in \mathbb{R}^{d_c \times c n_h}$ is the up-projection matrices for queries.
Note that the latent query vector $\mathbf{c}_{t}^{Q}$ is shared with that used for the indexer queries. 
Next, we perform MQA on $\{\mathbf{q}_{t, i}\}$ and $\mathcal{C}^{\text{SprsComp}}_t$: 
\begin{equation}
    \mathbf{o}_{t,i} = \operatorname{CoreAttn}\left( \texttt{query=}\mathbf{q}_{t,i}, \texttt{key=}\mathcal{C}^{\text{SprsComp}}_t, \texttt{value=}\mathcal{C}^{\text{SprsComp}}_t \right),
\end{equation}
where $\mathbf{o}_{t,i} \in \mathbb{R}^{c}$ is the core attention output of the $i$-th head at the $t$-th token; $\operatorname{CoreAttn}(\cdot)$ denotes the core attention operation. 

\paragraph{Grouped Output Projection.}
In the configuration of \dsviv{}, $c n_h$ is quite large.
Therefore, directly projecting the outputs of the core attention operation $[\mathbf{o}_{t,1}; \mathbf{o}_{t,2}; ...; \mathbf{o}_{t,n_h}]=\mathbf{o}_{t} \in \mathbb{R}^{c n_h}$ to a $d$-dimensional hidden state will impose a substantial computational burden. 
To mitigate this cost, we design a grouped output projection strategy.
To be specific, we first split $n_h$ outputs into $g$ groups, and then for each group of output $\mathbf{o}^G_{t, i} \in \mathbb{R}^{c \frac{n_h}{g}}$, we project it to a $d_g$-dimensional intermediate output $\mathbf{o}^{G'}_{t, i} \in \mathbb{R}^{d_g}$, where $d_g < c \frac{n_h}{g}$. 
Finally, we project the intermediate output $[\mathbf{o}^{G'}_{t, 1}; \mathbf{o}^{G'}_{t, 2}; ...; \mathbf{o}^{G'}_{t, g}] \in \mathbb{R}^{d_gg}$ to the final attention output $\mathbf{\hat{o}}_{t} \in \mathbb{R}^{d}$. 

\begin{figure}[t]
\centering
\includegraphics[width=0.6\linewidth]{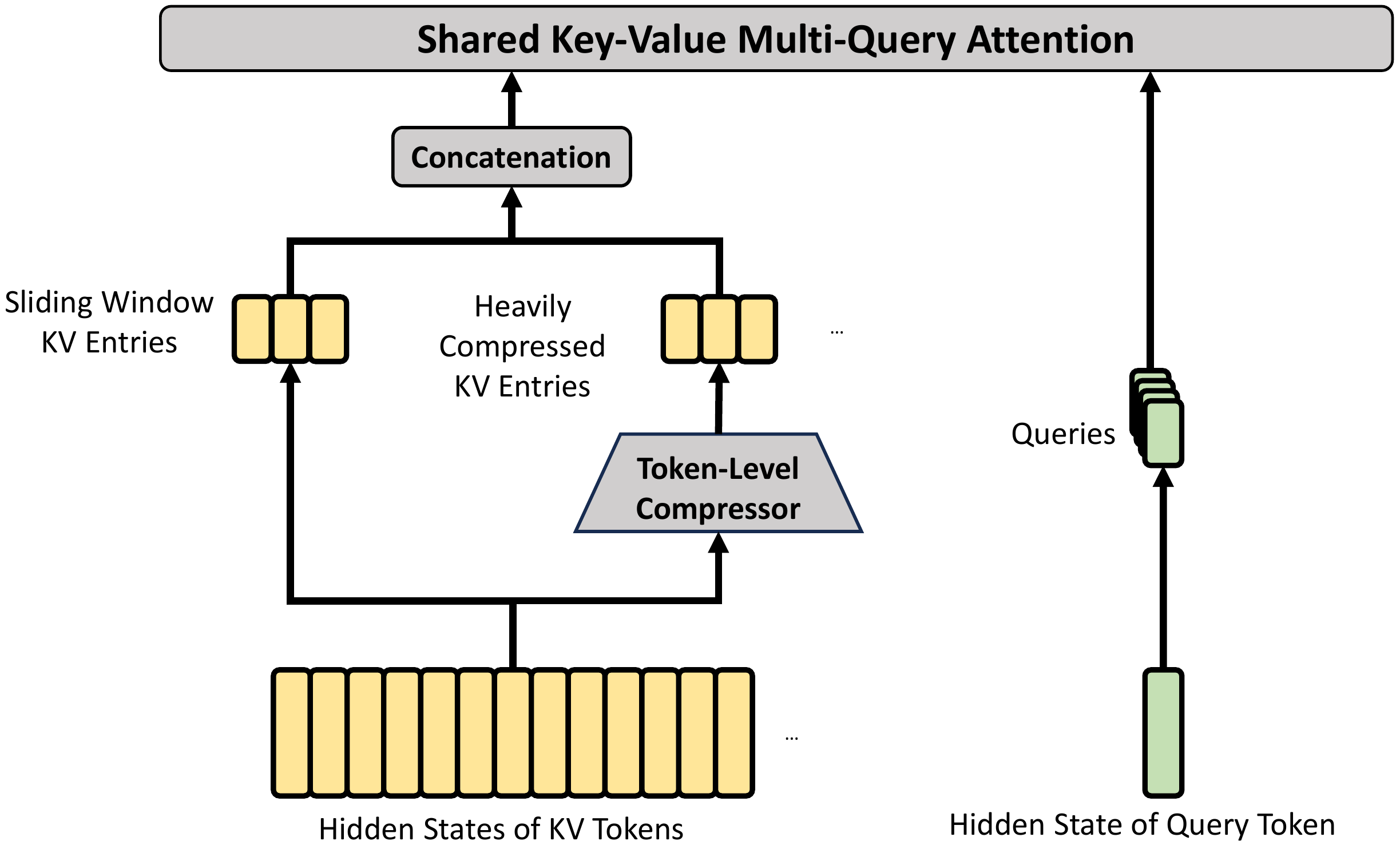}
\caption{
    Core architectures of \hca{}.
    It performs heavier compression, where the KV entries of $m^{\prime}$ ($\gg m$) tokens will be consolidated into one. 
    Also, we additionally introduce a small set of sliding window KV entries to enhance local fine-grained dependencies. 
}
\label{fig:hca}
\end{figure}

\subsubsection{\hcafull{}}

The core architecture of \hca{} is illustrated in Figure~\ref{fig:hca}, which compresses the KV cache in a heavier manner, but does not employ sparse attention. 

\paragraph{Compressed Key-Value Entries.}
By and large, the compression strategy of \hca{} is similar to that of \csa{}, but employs a larger compression rate $m^{\prime} $ ($\gg m$) and does not perform overlapped compression. 
Let $H \in \mathbb{R}^{n \times d}$ be a sequence of input hidden states, \hca{} first computes the original KV entries $C \in \mathbb{R}^{n \times c}$ and their corresponding compression weights $Z \in \mathbb{R}^{n \times c}$:
\begin{align}
    C &= H \cdot W^{KV}, \\
    Z &= H \cdot W^{Z},
\end{align}
where $W^{KV}, W^{Z} \in \mathbb{R}^{d \times c}$ are trainable parameters.
Next, each $m^{\prime}$ KV entries in $C$ will be compressed into one according to the compression weights and learnable positional biases $B \in \mathbb{R}^{m^{\prime} \times c}$, producing $C^{\text{Comp}} \in \mathbb{R}^{\frac{n}{m^{\prime}} \times c}$. 
Each compressed entry $C^{\text{Comp}}_{i} \in \mathbb{R}^{c}$ is computed by
\begin{align}
     S_{m^{\prime}i:m^{\prime}(i+1)-1} &=  \operatorname{Softmax}_{\text{row}}(Z_{m^{\prime}i:m^{\prime}(i+1)-1} + B), \\
    C^{\text{Comp}}_{i} &= \sum_{j=m^{\prime}i}^{m^{\prime}(i+1)-1} S_j \odot C_{j}.
\end{align}
Through this compression operation, \hca{} compresses the sequence length to $\frac{1}{m^{\prime}}$ times. 

\paragraph{Shared Key-Value MQA and Grouped Output Projection.}
\hca{} also employs the shared KV MQA and grouped output projection strategies as \csa{} does. 
After the KV compression, for a query token $t$, \hca{} first produces attention queries $\{\mathbf{q}_{t, 1};\mathbf{q}_{t, 2};...;\mathbf{q}_{t, n_{h}}\}$ in a low-rank manner: 
\begin{align}
    \mathbf{c}_{t}^{Q} &= \mathbf{h}_{t} \cdot W^{DQ}, \\
    [\mathbf{q}_{t, 1};\mathbf{q}_{t, 2};...;\mathbf{q}_{t, n_{h}}] = \mathbf{q}_{t} &= \mathbf{c}_{t}^{Q} \cdot W^{UQ},
\end{align}
where $\mathbf{h}_{t} \in \mathbb{R}^{d}$ is the input hidden state of the query token $t$; $n_{h}$ denotes the number of query heads; $W^{DQ} \in \mathbb{R}^{d \times d_c}$ and $W^{UQ} \in \mathbb{R}^{d_c \times c n_h}$ are the down-projection and up-projection matrices for queries, respectively. 
Next, we perform MQA on $\{\mathbf{q}_{t, i}\}$ and $C^{\text{Comp}}$: 
\begin{equation}
    \mathbf{o}_{t,i} = \operatorname{CoreAttn}\left( \texttt{query=}\mathbf{q}_{t,i}, \texttt{key=}C^{\text{Comp}}, \texttt{value=}C^{\text{Comp}} \right),
\end{equation}
where $\mathbf{o}_{t,i} \in \mathbb{R}^{c}$ is the core attention output of the $i$-th head at the $t$-th token. 
Next, as \csa{} does, \hca{} splits $n_h$ outputs into $g$ groups, and for each group of output $\mathbf{o}^G_{t, i} \in \mathbb{R}^{c \frac{n_h}{g}}$, \hca{} projects it to a $d_g$-dimensional intermediate output $\mathbf{o}^{G'}_{t, i} \in \mathbb{R}^{d_g}$, where $d_g < c \frac{n_h}{g}$. 
Finally, \hca{} projects the intermediate output $[\mathbf{o}^{G'}_{t, 1}; \mathbf{o}^{G'}_{t, 2}; ...; \mathbf{o}^{G'}_{t, g}] \in \mathbb{R}^{d_gg}$ to the final attention output $\mathbf{\hat{o}}_{t} \in \mathbb{R}^{d}$. 

\subsubsection{Other Details}

In addition to the core architectures of \csa{} and \hca{} described above, our hybrid attention incorporates several other techniques. 
For writing clarity, we omit these additional techniques from the above introduction and will briefly describe them in this subsection.
Also, this subsection focuses only on the core ideas of them and may omit some tiny details for simplicity. 
We encourage the readers to refer to our open-source implementation for unambiguous details.

\paragraph{Query and Key-Value Entry Normalization.}
For both \csa{} and \hca{}, we perform an additional RMSNorm operation on each head of the queries and the only head of the compressed KV entries, just before the core attention operation. 
This normalization avoids exploding attention logits and may improve training stability. 

\paragraph{Partial Rotary Positional Embedding.}
For both \csa{} and \hca{}, we partially employ the Rotary Positional Embedding (RoPE)~\citep{su2024roformer} to the attention queries, KV entries, and the core attention outputs. 
To be specific, for each query vector and KV entry vector used in \csa{} and \hca{}, we apply RoPE to its last 64 dimensions. 
Since the KV entries serve as both attention keys and values, the naive core attention outputs $\{\mathbf{o}_{t,i}\}$ will carry absolute position embeddings, derived from the weighted sum of KV entries. 
As a countermeasure, we also apply RoPE with position $-i$ on the last 64 dimensions of each $\mathbf{o}_{t,i}$. 
In this way, the output of the core attention will also carry relative position embeddings --- the contribution of each KV entry to the core attention outputs will also be related to the distance between the query and the KV entry.

\paragraph{Additional Branch of Sliding Window Attention.}
In order to strictly preserve causality in \csa{} and \hca{}, each query attends to only preceding compressed KV blocks. 
Consequently, a query cannot access information from other tokens within its own compressed block.
Meanwhile, recent tokens usually possess greater relevance to the query token in language modeling.
For these reasons, we introduce a supplementary attention branch to both \csa{} and \hca{} in a sliding window manner, for better modeling of local dependencies.
To be specific, for each query token, we additionally produce $n_{\text{win}}$ uncompressed KV entries corresponding to the recent $n_{\text{win}}$ tokens. 
In the core attention of \csa{} and \hca{}, these KV entries in the sliding window will be used along with the compressed KV entries. 

\paragraph{Attention Sink.}
In the core attention of \csa{} and \hca{}, we employ the trick of attention sink~\citep{attn_sink,gpt_oss}. 
To be specific, we set a series of learnable sink logits $\{z^{\prime}_1, z^{\prime}_2, ..., z^{\prime}_{n_h}\}$. 
For the $h$-th attention head, $\operatorname{Exp}(z^{\prime}_h)$ will be added to the denominator of the attention score:
\begin{equation}
    s_{h, i, j} = \frac{\operatorname{Exp}(z_{h, i, j})}{\sum_k \operatorname{Exp}(z_{h, i, k}) + \operatorname{Exp}(z^{\prime}_h)},
\end{equation}
where $s_{h, i, j}, z_{h, i, j} \in \mathbb{R}$ denote the attention score and attention logit of the $h$-th attention head between the $i$-th query token and the $j$-th preceding token or compressed block.
This technique allows each query head to adjust its total attention scores to be not equal to 1, and even to be near 0.

\subsubsection{Efficiency Discussion}

Due to the employment of hybrid \csa{} and \hca{}, together with low-precision computation and storage, the attention module of \dsviv{} series achieves remarkable efficiency in both attention FLOPs and KV cache size, especially in long-context scenarios. 
First, we adopt a mixed storage format for KV entries: BF16 precision is used for the rotary positional embedding (RoPE) dimensions, while FP8 precision is applied to the remaining dimensions. 
This hybrid representation reduces the KV cache size by nearly half compared with pure BF16 storage. 
Second, attention computation within the lightning indexer is performed in FP4 precision, which accelerates the attention operation under extremely long contexts. 
Third, relative to \dsviiiII{}, a smaller attention top-k is chosen in \dsviv{} series, thereby improving model efficiency on short- and medium-length texts. 
Finally, and most importantly, compressed attention and hybrid attention techniques substantially reduce both the KV cache size and the computational FLOPs. 

Taking BF16 GQA8~\citep{ainslie2023gqa} with a head dimension of 128 as the baseline --- one of the common configurations of LLM attention --- the KV cache size of \dsviv{} series can be dramatically reduced to approximately $2\%$ times of that baseline in the 1M-context setting.
Moreover, even when compared with \dsviiiII{}~\citep{dsv32} --- already an efficient baseline --- \dsviv{} series still exhibits substantial advantages in efficiency. 
A comparison of their inference FLOPs and KV cache size is provided in the right part of Figure~\ref{fig:dsv4_performance_and_cost}.

\begin{algorithm}[t]
\caption{Muon Optimizer for \dsviv{}}
\label{alg:muon}
\begin{algorithmic}[1]
\Require{Learning rate $\eta$, momentum $\mu$, weight decay $\lambda$, update rescaling factor $\gamma$}
\For{each training step $t$}
    \For{each logically independent weight $W \in \mathbb{R}^{n \times m}$}
        \State $G_t = \nabla_{W} \mathcal{L}_t(W_{t-1})$ 
        \Comment{Compute gradients}
        \State $M_t = \mu M_{t-1} + G_t$ 
        \Comment{Accumulate momentum buffer}
        \State $O^{\prime}_t = \operatorname{HybridNewtonSchulz}(\mu M_t + G_t)$ 
        \Comment{Nesterov trick and hybrid Newton-Schulz}
        \State $O_t = O^{\prime}_t \cdot \sqrt{\max(n, m)} \cdot \gamma$ 
        \Comment{Rescale the update RMS}
        \State $W_t = W_{t-1} \cdot \left( 1 - \eta \lambda \right) - \eta O_t$ 
        \Comment{Perform weight decay and update}
    \EndFor
\EndFor
\end{algorithmic}
\end{algorithm}

\subsection{Muon Optimizer}

We employ the Muon~\citep{muon,muon_kimi} optimizer for the majority of modules in \dsviv{} series due to its faster convergence and improved training stability.
The full algorithm of our Muon optimization is summarized in Algorithm~\ref{alg:muon}.

\paragraph{Basic Configurations.}
We maintain the AdamW~\citep{adamW} optimizer for the embedding module, the prediction head module, the static biases and gating factors of \mhc{} modules, and the weights of all RMSNorm modules. 
All other modules are updated with Muon.
Following \citet{muon_kimi}, we also apply weight decay to Muon parameters, use the Nesterov~\citep{nesterov,muon} trick, and rescale the Root Mean Square (RMS) of the update matrix for reutilization of our AdamW hyper-parameters.
Different from them, we use hybrid Newton-Schulz iterations for orthogonalization.

\paragraph{Hybrid Newton-Schulz Iterations.}
For a given matrix $M$, let its Singular Value Decomposition (SVD) be $M = U \Sigma V^T$. 
The Newton-Schulz iterations aim to approximately orthogonalize $M$ to be $UV^T$.
Usually, $M$ will be first normalized as $M_0=M / ||M||_F$ to ensure its maximum singular value does not exceed 1. 
Then, each Newton-Schulz iteration performs the following operation:
\begin{equation}
    M_k = a M_{k-1} + b (M_{k-1} M_{k-1}^T) M_{k-1} + c (M_{k-1} M_{k-1}^T)^2 M_{k-1}.
\end{equation}
Our hybrid Newton-Schulz performs 10 iterations over two distinct stages. 
During the first 8 steps, we use coefficients $(a, b, c) = (3.4445, -4.7750, 2.0315)$ to drive rapid convergence, bringing the singular values close to 1. 
In the final 2 steps, we switch to coefficients $(a, b, c) = (2, -1.5, 0.5)$, which stabilize the singular values precisely at 1.

\paragraph{Avoiding Exploding Attention Logits.}
The attention architecture of \dsviv{} series allows us to directly apply RMSNorm on the attention queries and KV entries, which effectively prevents attention logits from exploding. 
Consequently, we do not employ the QK-Clip technique~\citep{muon_kimi} in our Muon optimizer. 

\section{General Infrastructures}
\label{sec:infra}

\subsection{Fine-Grained Communication-Computation Overlap in Expert Parallelism}
Mixture-of-Experts (MoE) can be accelerated via Expert Parallelism (EP).
However, EP requires complex inter-node communication and imposes substantial demands on interconnect bandwidth and latency.
To alleviate the communication bottleneck in EP and achieve higher end-to-end performance under lower interconnection bandwidth requirements, we propose a fine-grained EP scheme that fuses communication and computation into a single pipelined kernel for communication-computation overlapping.

\begin{figure}[h]
    \centering
    \includegraphics[width=\linewidth]{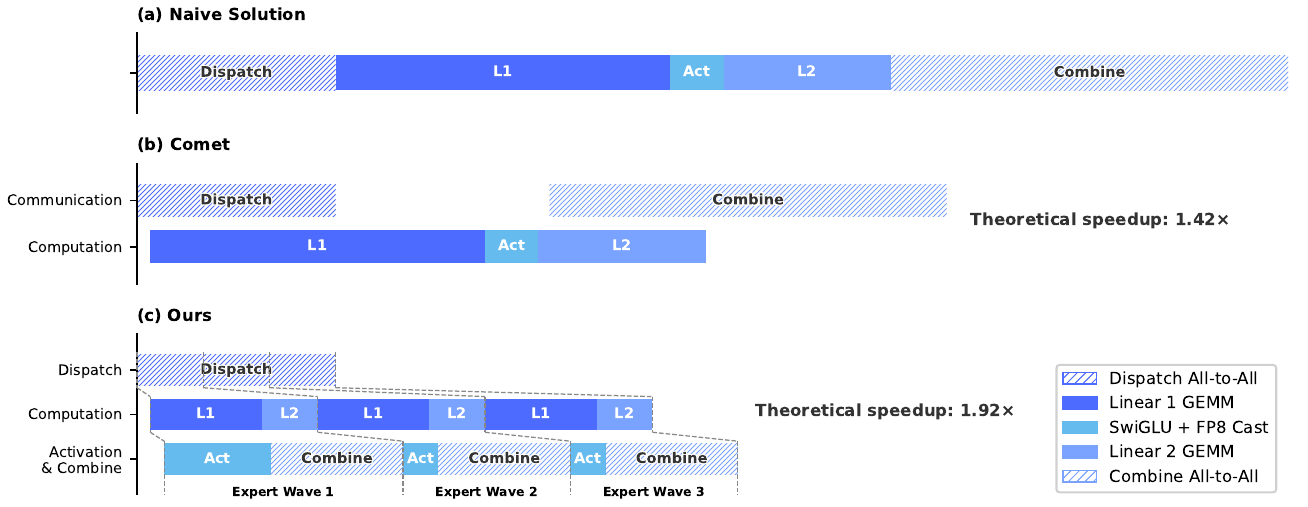}
    \caption{
        Illustration of our EP scheme with related works. 
        Comet~\citep{comet} overlaps Dispatch with Linear-1, and Linear-2 with Combine, separately. 
        Our EP scheme achieves a finer-grained overlapping by splitting and scheduling experts into waves. 
        The theoretical speedup is evaluated in the configuration of the \dsvivf{} architecture. 
    }
    \label{fig:moe-kernel-compare}
\end{figure}

\paragraph{Communication Latency Can Be Hidden.}
The key insight of our EP scheme is that the communication latency can be effectively hidden beneath computation in MoE layers.
As shown in \autoref{fig:moe-kernel-compare}, in \dsviv{} series, each MoE layer can be decomposed mainly into four stages: two communication-bound stages, \emph{Dispatch} and \emph{Combine}, and two computation-bound stages, \emph{Linear-1} and \emph{Linear-2}.
Our profiling reveals that within a single MoE layer, the total time of communication is less than that of the computation.
Therefore, after fusing communication and computation into a unified pipeline, computation remains the dominant bottleneck, implying that the system can tolerate lower interconnect bandwidth without degrading end-to-end performance.

\paragraph{Fine-Grained EP Scheme.}
To further lower the interconnect bandwidth requirement and amplify the benefits of overlapping, we introduce a finer-grained expert partitioning scheme.
Inspired by many related works~\citep{aimuyo2025FlashMoE,comet}, we split and schedule the experts into \emph{waves}.
Each wave consists of a small portion of experts.
As soon as all experts within the wave have completed their communication, computation can commence immediately without waiting for other experts.
In steady state, computation of current wave, token transfer for the next wave, and result sending of completed experts all proceed concurrently, as demonstrated in \autoref{fig:moe-kernel-compare}.
This forms a fine-grained pipeline among experts, keeping both computation and communication continuous throughout the wave.
The wave-based scheduling speeds up the performance on extreme cases such as Reinforcement Learning (RL) rollout, which usually encounters long-tail small batches.

\paragraph{Performance and Open-Sourced Mega-Kernel.}
We validated the fine-grained EP scheme on both NVIDIA GPUs and HUAWEI Ascend NPUs platforms. 
Compared against strong non-fused baselines, it achieves $1.50\sim1.73\times$ speedup for general inference workloads, and up to $1.96\times$ for latency-sensitive scenarios such as RL rollouts and high-speed agent serving. 
We have open-sourced the CUDA-based mega-kernel implementation named  \textbf{MegaMoE}\footnote{\url{https://github.com/deepseek-ai/DeepGEMM/pull/304}} as a component of DeepGEMM.

\paragraph{Observations and Proposals.}
We share observations and lessons from kernel development and offer some proposals to hardware vendors, in the hope of aiding efficient hardware design and achieving better software-hardware co-design:

\begin{itemize}[topsep=0pt]
    \item \textbf{Computation-Communication Ratio.}
    Full communication-computation overlap hinges on the computation-communication ratio, rather than the bandwidth solely.
    Denoting peak compute throughput as $C$ and interconnect bandwidth as $B$, communication can be fully hidden when $C/B \leq V_{\text{comp}} / V_{\text{comm}}$, where $V_{\text{comp}}$ denotes the computation volume and $V_{\text{comm}}$ refers to the communication volume.
    For \dsvivp{}, where each token-expert pair requires $6hd$ FLOPs (SwiGLU gate, up, and down projections) but only $3h$ bytes of communication (FP8 Dispatch + BF16 Combine), this simplifies to:
    $$\frac{C}{B} \leq 2d = 6144 \; \mathrm{FLOPs/Byte}.$$
    That is, each GBps of interconnect bandwidth suffices to hide the communication for 6.1 TFLOP/s of compute.
    Once bandwidth meets this threshold, it ceases to be the bottleneck, and devoting additional silicon area to further bandwidth brings diminishing returns.
    We encourage future hardware designs to target such balance points rather than scale bandwidth unconditionally.

    \item \textbf{Power Budget.}
    Extreme kernel fusion drives compute, memory, and network to high load simultaneously, making power throttling a key performance limiter. 
    We suggest that future hardware designs provide sufficient power headroom for such fully concurrent workloads.

    \item \textbf{Communication Primitives.}
    In the dispatch stage, we adopt a pull-based approach where each GPU actively reads activations from remote GPUs, avoiding the high notification latency that fine-grained push entails. 
    Future hardware with lower-latency cross-GPU signaling would make push viable and enable more natural communication patterns.

    \item \textbf{Activation Function.}
    We propose replacing SwiGLU with a low-cost element-wise activation that involves no exponential or division operations. 
    This directly reduces the overhead of post-GEMM processing, preventing the GEMM pipeline from being stalled by activation function computation, thereby enhancing overall computational throughput and resource utilization.
\end{itemize}

\subsection{Flexible and Efficient Kernel Development with TileLang}

In practice, our elaborate model architecture would have resulted in hundreds of fine-grained Torch ATen operators.
We adopt TileLang~\citep{wangtilelang} to develop a set of fused kernels to replace the vast majority of them, delivering optimal performance with minimal effort.
It also allows us to quickly prototype operators like attention variants during validation.
These kernels play critical roles in model architecture development,  large‑scale training, and ultimately production deployment of inference services.
As a Domain-Specific Language (DSL), TileLang balances development productivity with runtime efficiency, enabling rapid development while supporting deep, iterative optimizations within the same codebase.
Additionally, we collaborate closely with the TileLang community to foster a more agile, efficient, and stable kernel development workflow. 

\paragraph{Reducing Invocation Overhead with Host Codegen.}
As accelerators continue to grow in performance, CPU-side orchestration overhead becomes increasingly prominent. For small, highly optimized kernels, such fixed host overhead can easily cap utilization and throughput. A common source of this overhead is that host-side logic, such as runtime contract checks, is typically written in Python for flexibility and thus incurs a fixed per-invocation cost.

We mitigate this overhead with \textit{Host Codegen}, which moves most host-side logic into generated host code. Specifically, we first co-generate the device kernel and a lightweight host launcher at the IR (Intermediate Representation) level, embedding the necessary metadata—such as data types, rank/shape constraints, and stride/layout assumptions—parsed from the language frontend. The launcher is then lowered to the host source code built on top of the TVM-FFI~\citep{tvm} framework, whose compact calling convention and zero-copy tensor interop together minimize host-side overhead. At runtime, this generated host code performs validation and argument marshaling, shifting all per‑invocation checks out of the Python execution path. Our measurements show that CPU-side validation overhead drops from tens or hundreds of microseconds to less than one microsecond per invocation.

\paragraph{SMT-Solver-Assisted Formal Integer Analysis.}
TileLang kernels involve complex tensor index arithmetic that requires strong formal integer analysis.
During compilation passes such as layout inference, memory hazard detection, and bound analysis, the compiler must verify whether integer expressions satisfy specific properties to enable the corresponding optimizations.
Therefore, stronger formal analysis capabilities can unlock more advanced and complex optimization opportunities.

To this end, we integrate the Z3 SMT solver~\citep{demoura2008z3} into TileLang's algebraic system, providing formal analysis capability for most integer expressions in tensor programs. 
We strike a balance between computational overhead and formal expressiveness by translating TileLang's integer expressions into Z3's quantifier-free non-linear integer arithmetic (QF\_NIA).
Based on Integer Linear Programming (ILP) solvers, QF\_NIA seamlessly resolves standard linear integer expressions common in kernels.
Furthermore, its inherent non-linear reasoning capacity effectively addresses advanced challenges like vectorization over variable tensor shapes.
Under reasonable resource limits, Z3 elevates overall optimization performance while restricting compilation time overhead to just a few seconds.
The impact is substantial across multiple passes, including vectorization, barrier insertion, and code simplification.

\paragraph{Numerical Precision and Bitwise Reproducibility.}
In production settings, numerical correctness and reproducibility are as critical as raw throughput. 
We therefore prioritize accuracy by default: fast‑math optimizations are disabled at the compiler level, and precision‑affecting approximations are provided only as explicit, opt-in frontend operators (e.g., \texttt{T.\_\_exp}, \texttt{T.\_\_log}, and \texttt{T.\_\_sin}). 
Conversely, when strict IEEE-754 semantics are required, TileLang provides IEEE-compliant intrinsics with explicit rounding modes (e.g., \texttt{T.ieee\_fsqrt}, \texttt{T.ieee\_fdiv}, and \texttt{T.ieee\_add}), enabling developers to precisely specify numerical behavior.

We also target bitwise reproducibility for validating kernels against hand‑written CUDA baselines. 
We align TileLang's algebraic simplification and lowering rules with mainstream CUDA toolchains (e.g., NVCC) to avoid transformations that introduce unintended bit‑level differences. 
Layout annotations  (e.g., \texttt{T.annotate\_layout}) further allow users to pin down layout-dependent lowering decisions, keeping evaluation and accumulation order consistent with the reference CUDA implementation and thus enabling bit-identical outputs when desired.

Our evaluation shows that these accuracy- and reproducibility-oriented design choices do not sacrifice performance: under conservative defaults, TileLang kernels remain competitive, while exposing knobs to selectively relax numerical constraints for higher speed.   

\subsection{High-Performance Batch-Invariant and Deterministic Kernel Libraries}
To enable efficient training and inference, we develop a comprehensive set of high-performance computational kernels.
Beyond basic functionalities and maximizing hardware utilization, another pivotal design goal is to ensure training reproducibility and bitwise alignment among pre-training, post-training, and inference pipelines. 
Therefore, we implement end-to-end, bitwise batch-invariant, and deterministic kernels with minimal performance overhead.
These kernels are helpful for debugging, stability analysis, and consistent post-training behavior. 

\paragraph{Batch Invariance.}
Batch invariance ensures that the output of any given token remains bitwise identical, regardless of its position within a batch.
To implement batch invariance, the primary challenges are listed as follows:
\begin{itemize}[topsep=0pt]
    \item \textbf{Attention.} 
    To achieve batch invariance, we cannot use the split-KV method~\citep{flash_decoding}, which distributes the attention computation for a single sequence across multiple Stream Multiprocessors (SMs) to balance the load of SMs. 
    However, abandoning this technique will lead to severe wave-quantization problems\footnote{\url{https://docs.nvidia.com/deeplearning/performance/dl-performance-matrix-multiplication/index.html\#wave-quant}}, which can adversely affect GPU utilization. 
    To address this, we develop a dual-kernel strategy for batch-invariant decoding. 
    The first kernel computes the attention output for an entire sequence within a single SM, ensuring high throughput for fully occupied waves. 
    The second kernel, to minimize the latency of the final partially-filled wave and thus alleviate wave-quantization, uses multiple SMs for a single sequence. 
    For the bitwise identity of these two kernels, we carefully design the calculation path of the second kernel to ensure its accumulation order is the same as that of the first kernel. 
    Additionally, the second kernel utilizes distributed shared memory\footnote{\url{https://docs.nvidia.com/cuda/cuda-programming-guide/02-basics/writing-cuda-kernels.html\#distributed-shared-memory}} within thread-block clusters, enabling high-speed data exchange across SMs.
    This dual-kernel method effectively confines the overhead of batch-invariant decoding to be negligible.
    
    \item \textbf{Matrix Multiplication.} 
    Traditional cuBLAS library~\citep{cublas} cannot achieve batch invariance. 
    Therefore, we replace it end-to-end with DeepGEMM~\citep{deepgemm2025}. 
    Furthermore, for very small batch sizes, conventional implementation usually employs split-k~\citep{stream_k} techniques to improve performance.
    Unfortunately, split-k techniques cannot guarantee batch invariance, a pivotal feature in \dsviv{}. 
    Therefore, we abandon split-k in most scenarios, which, however, may cause performance degradation.
    To address this, we introduce a set of optimizations that enable our implementation of matrix multiplication to match or even surpass the performance of standard split-k in most major scenarios.
\end{itemize}

\paragraph{Determinism.}
Deterministic training is highly beneficial for debugging hardware or software issues.
Moreover, when training exhibits anomalies such as loss spikes, determinism enables researchers to more easily pinpoint numerical causes and further refine the model design.
Non-determinism in training typically stems from non-deterministic accumulation order, often due to the use of atomic addition instructions. 
This issue primarily occurs during the backward pass, notably at the following parts:
\begin{itemize}[topsep=0pt]
    \item \textbf{Attention Backward.} 
    In conventional implementations of backward propagation for sparse attention, we use \texttt{atomicAdd} to accumulate gradients for the KV tokens. 
    This introduces non-determinism due to the non-associativity of floating-point addition. 
    To address this problem, we allocate separate accumulation buffers for each SM, followed by a global deterministic summation across all buffers.
    
    \item \textbf{MoE Backward.} 
    When multiple SMs from different ranks concurrently write data to the same buffer on a receiving rank, negotiating writing positions also introduces non-determinism. 
    To resolve this, we design a token order pre-processing mechanism within each single rank, combined with buffer isolation across multiple ranks. 
    This strategy ensures determinism of both the send results of expert parallelism and the accumulation order in the MoE backward pass.

    \item \textbf{Matrix Multiplication in \mhc{}.} 
    \mhc{} involves a matrix multiplication with an output dimension of only 24. 
    For very small batch sizes, we are compelled to use the split-k~\citep{stream_k} algorithm, whose naive implementation will cause non-determinism. 
    To overcome this, we output each split part separately and perform a deterministic reduction in a subsequent kernel, thereby preserving both performance and determinism.
\end{itemize}

\subsection{Training Framework}
\label{subsec:training-framework}

Our training framework is built upon the scalable and efficient infrastructure developed for \dsviii{}~\citep{dsv3}.
In training \dsviv{}, we inherit this robust foundation while introducing several key innovations to accommodate its novel architectural components --- specifically the Muon optimizer, \mhc{}, and the hybrid attention mechanism --- while maintaining high training efficiency and stability.

\subsubsection{Efficient Implementation of Muon}
The Muon optimizer requires the full gradient matrix to compute parameter updates, which presents a challenge when combined with the Zero Redundancy Optimizer (ZeRO)~\citep{zero}. 
Traditional ZeRO is designed for element-wise optimizers like AdamW, where a single parameter matrix can be partitioned and updated across multiple ranks. 
To address this conflict, we design a hybrid strategy of ZeRO bucket assignment for Muon. 

For dense parameters, we limit the maximum size of ZeRO parallelism and employ a knapsack algorithm to assign parameter matrices to these ranks, ensuring each rank manages a roughly balanced load.
The bucket on each rank is padded to match the size of the largest bucket across ranks, facilitating efficient reduce-scatter operations.
This padding typically incurs less than 10\% memory overhead in our setup, where each rank manages no more than five parameter matrices. 
When the overall size of data parallelism exceeds the limit for ZeRO, we compute the Muon update redundantly across the extra data-parallel groups, trading computation for reduced total bucket memory.

For MoE parameters, we optimize each expert independently. 
We first flatten all down projection matrices in SwiGLU~\citep{shazeer2020glu} of all experts across all layers, followed by flattened up projection matrices and gate matrices. 
Then, we pad the flattened vector to ensure we can evenly distribute this vector across all ranks without splitting any logically independent matrix. 
Given the large number of experts, we do not impose a limit of ZeRO parallelism for MoE parameters, and the padding overhead is also negligible.

Additionally, on each rank, consecutive parameters of identical shape will be automatically merged, enabling batched execution of the Newton-Schulz iterations for better hardware utilization.
Furthermore, we observe that the Newton-Schulz iterations in Muon remain stable when computed with BF16 matrix multiplications. 
Leveraging this, we further quantize, in a stochastic rounding manner, the MoE gradients to be synchronized across data-parallel ranks to the BF16 precision, halving the communication volume. 
To avoid accumulation errors introduced by low-precision adders, we replace conventional tree- or ring-based reduce-scatter collectives with a two-phase approach. 
First, an all-to-all operation exchanges local gradients across ranks, and then each rank performs a local sum in FP32. This design maintains numerical robustness.

\subsubsection{Cost-Effective and Memory-Efficient Implementation of \mhc{}}
The introduction of \mhc{} increases both activation memory consumption and communication volume between pipeline stages, compared with conventional residual connections.
To mitigate these costs, we implement several optimization strategies.

Firstly, we carefully design and implement fused kernels of \mhc{} for both training and inference.
Secondly, we introduce a recomputation strategy that selectively checkpoints intermediate tensors. 
Specifically, we recompute most hidden states between layers and all normalized layer inputs, while avoiding recomputation of compute-intensive operations. 
This achieves a balance between memory saving and computational overhead. 
Thirdly, we adjust the DualPipe 1F1B overlapping scheme to accommodate the increased pipeline communication and enable concurrent execution of some operations in \mhc{}.

Collectively, these optimizations constrain the wall-time overhead of \mhc{} to only 6.7\% of the overlapped 1F1B pipeline stage. 
More details of the engineering optimization can be found in the dedicated \mhc{} paper~\citep{mhc}.

\subsubsection{Contextual Parallelism for Long-Context Attention}
Conventional Context Parallelism (CP) partitions the sequence dimension, with each rank maintaining contiguous $s$ tokens. 
This introduces two challenges to our compressed attention mechanisms (i.e., \csa{} and \hca{}). 
On the one hand, training samples are packed from multiple sequences, and each sequence is compressed independently by a factor of $m$ (or $m^{\prime}$), with any trailing tokens fewer than $m$ being discarded. 
Consequently, the compressed KV lengths are typically less than $\frac{s}{m}$ and vary across ranks. 
On the other hand, the compression requires $m$ consecutive KV entries, which may straddle the boundary between two neighboring CP ranks.

To address these challenges, we design a two-stage communication approach. 
In the first stage, each rank $i$ sends its last $m$ uncompressed KV entries to rank $i+1$. 
Then, rank $i+1$ compresses some of these received entries together with its local $s$ uncompressed KV entries, producing a fixed length of $\frac{s}{m}+1$ compressed entries, in which exist some padding entries. 
In the second stage, an all-gather operation across all CP ranks collects the locally compressed KV entries. 
Then, a fused select-and-pad operator reorganizes them into the full set of compressed KV entries with a total length of $\texttt{cp\_size} \cdot \frac{s}{m}$. 
Any padding entries are placed at the tail.  
For \hca{} and the indexer in \csa{}, the visible range of compressed KV entries for each query token can be precomputed by rules. 
For the sparse attention in \csa{}, the top-$k$ selector explicitly specifies the indices of visible compressed KV entries for each query.

\subsubsection{Extended Automatic Differentiation for Flexible Activation Checkpointing}
Conventional activation checkpointing implementations operate at the granularity of an entire module, deciding whether to retain or recompute its output activations during the backward pass.
This coarse granularity often leads to suboptimal trade-offs between recomputation cost and activation memory footprint. 
An alternative approach is to manually implement the forward and backward logic of an entire layer, explicitly managing tensor checkpointing states. 
While enabling fine-grained control, this method loses the convenience of the automatic differentiation framework, substantially increasing development complexity.

To achieve fine-grained control without sacrificing programming efficiency, we implement a tensor-level activation checkpointing mechanism with automatic differentiation support. 
With this mechanism, developers only need to implement the forward pass and selectively annotate individual tensors for automatic checkpointing and recomputation. 
Our framework leverages TorchFX~\citep{torchfx} to trace the full computation graph.
For each annotated tensor, it performs a backward traversal to identify the minimal subgraph required for its recomputation. 
We define these minimal subgraphs as recomputation graphs and insert them into the backward logic just before the corresponding gradient computation.

Compared with the manual implementation, this design introduces no additional overhead during training. 
Recomputation in this framework is implemented by directly freeing the GPU memory of the annotated tensor and reusing the storage pointer from the recomputed tensor, without any GPU memory copy. 
Furthermore, since graph tracing executes the model concretely, we can track the underlying storage pointer of each tensor, which enables automatic deduplication of recomputation for tensors that share storage (e.g., the input and output of a reshape operation). 
This relieves developers from reasoning about low-level memory details when annotating recomputation.

\subsection{Inference Framework}

Our inference framework largely inherits from that of \dsviii{}, with some differences in KV Cache management.

\subsubsection{KV Cache Structure and Management}
\begin{figure}[t]
\centering
\includegraphics[width=0.98\linewidth]{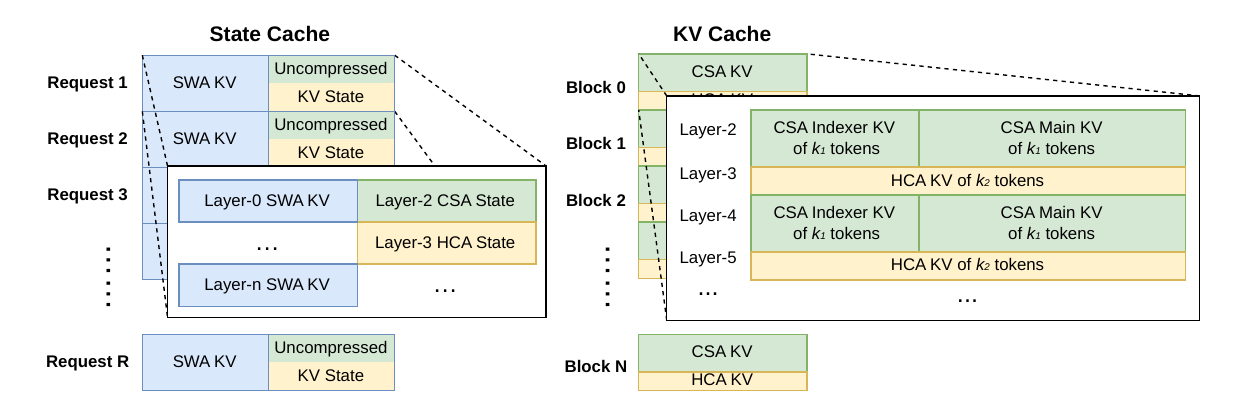}
\caption{
    Illustration of the KV cache Layout for \dsviv{}. 
    The KV cache is organized into two primary components: a classical KV cache for \csa{}/\hca{}, and a state cache for SWA and unready-for-compression tokens in \csa{}/\hca{}. 
    In the state cache, each request is assigned a fixed-size cache block.
    Within this block, the SWA segment stores the KV entries corresponding to the most recent $n_{\text{win}}$ tokens, while the \csa{}/\hca{} segment stores uncompressed tail states that are not yet ready for compression.
    In the classical KV cache, we allocate multiple blocks per request.
    Each cache block covers $\mathrm{lcm}(m, m^{\prime})$ original tokens, producing $k_1=\frac{\mathrm{lcm}(m,m^{\prime})}{m}$ \csa{} compressed tokens and $k_2=\frac{\mathrm{lcm}(m,m^{\prime})}{m^{\prime}}$ \hca{} compressed tokens.
}
\label{fig:kvcache}
\end{figure}

To efficiently manage the heterogeneous KV caches arising from the hybrid attention mechanism in \dsviv{}, we design a customized KV cache layout. 
The layout is illustrated in Figure~\ref{fig:kvcache}, and we will elaborate on it in detail as follows.

\paragraph{Heterogeneous KV Entries in \dsviv{}.}
The hybrid attention mechanism in \dsviv{} series introduces multiple types of KV entries with different Key-Value (KV) cache sizes and update rules.
The lightning indexer for sparse selection introduces additional dimensions into the KV cache that possess embedding sizes distinct from those in the primary attention. 
The compression techniques employed in \csa{} and \hca{} reduce the sequence length by factors of $\frac{1}{m}$ and $\frac{1}{m^\prime}$, respectively, thereby decreasing the overall KV cache size.
As a result, KV cache sizes vary across different layers. 
Furthermore, Sliding Window Attention (SWA) layers also operate with distinct KV cache sizes, as well as separate cache hit and eviction policies. 
In the compression branch, one KV entry is generated for every $m$ tokens. 
When the number of remaining tokens is insufficient for compression, all pending tokens and their associated hidden states must be retained in a buffer until the compression operation can be executed. 
These buffered tokens represent a sequence state determined by positional context and are also managed within the KV cache framework.

\paragraph{Challenges in Managing Hybrid Attention KV Cache.}
The hybrid attention mechanism violates fundamental assumptions behind PagedAttention and its variants.
Although recent hybrid KV cache managing algorithms (e.g., Jenga~\citep{jenga}, Hymba~\citep{dong2025hymba}) target general hybrid attention models or specific structures, two principal obstacles prevent consolidating KV caches across all layers under the PagedAttention framework:
\begin{itemize}[topsep=0pt]
    \item Diverse cache policies, such as those used in Sliding Window Attention.
    \item Constraints imposed by high-performance attention kernels, including alignment requirements.
\end{itemize}
For efficient KV cache management of \dsviv{}, we design corresponding strategies to overcome these two challenges.

\paragraph{State Cache for SWA and Uncompressed Tail Tokens.}
To address the first obstacle, we adopt an alternative cache management mechanism.
Since SWA is designed to enhance performance under a limited KV cache size, it is reasonable to treat it, along with the uncompressed tail tokens from the compression branch, as a state-space model. 
The corresponding KV cache can thus be regarded as a sequence-specific state that depends solely on the current position. 
Accordingly, we pre-allocate a fixed- and limited-size pool of state caches, and dynamically assign it to each sequence.

\paragraph{Sparse Attention Kernel Co-Design.}
Regarding the second obstacle, conventional high-performance attention kernels typically assume a fixed number $B$ of tokens per block to optimize performance, corresponding to $B \cdot m$ original tokens in \csa{} and $B \cdot m^\prime$ in \hca{}. 
Through employing a high-performance sparse-attention kernel, different layers can accommodate variable tokens per block without performance degradation. 
Achieving this requires co-designing the KV cache layout and the sparse attention kernel. 
For instance, padding blocks to align with cache lines can improve performance. 
Thus, for \csa{} with compression ratio $m$ and \hca{} with ratio $m^{\prime}$, the number of original tokens per block can be any multiple of $\operatorname{lcm}(m,m^{\prime})$, the least common multiple of these two compression ratios.

\subsubsection{On-Disk KV Cache Storage}
When serving \dsviv{}, we leverage an on-disk KV cache storage mechanism to eliminate repeated prefilling for shared-prefix requests. 
For the compressed KV entries in \csa{}/\hca{} and the uncompressed KV entries in Sliding Window Attention (SWA), we design separate solutions for storage management.

For \csa{} and \hca{}, we simply store all of the compressed KV entries to the disk.
When a request hits a stored prefix, we read and reuse the compressed KV entries corresponding to the prefix, until the last complete compression block.
Specially, for prefix tokens in the tail incomplete block, we still need to recompute them to restore the uncompressed KV entries, as uncompressed KV entries in \csa{} and \hca{} are not stored.

For the SWA KV entries, since they are not compressed and exist in every layer, their volume is approximately 8 times larger than the compressed \csa{} and \hca{} KV entries. 
To handle these large SWA KV entries efficiently, we propose and implement three distinct strategies for managing on-disk SWA KV entries, each offering a different trade-off between storage overhead and computational redundancy:
\begin{itemize}[topsep=0pt]
    \item \textbf{Full SWA Caching.}
    This strategy stores the complete SWA KV entries for all tokens, ensuring computational zero-redundancy.
    Under this strategy, the SWA KV entries of the hitting prefix can be reconstructed by just reading the on-disk cache of the last $n_{\text{win}}$ tokens within that prefix.
    Despite computational zero-redundancy, this strategy is inefficient for modern SSD-based storage systems --- only a small subset of the stored SWA KV cache will be accessed for each hitting request, which leads to an unbalanced write-intensive access pattern. 
    \item \textbf{Periodic Checkpointing.} 
    This strategy checkpoints SWA KV entries of the last $n_{\text{win}}$ tokens within every $p$ tokens, where $p$ is a tunable parameter. 
    For a hitting prefix, we load the most recent checkpointed state, and then recompute the remaining tail tokens. 
    Through tuning $p$, this strategy enables an on-demand trade-off between storage and computation. 
    \item \textbf{Zero SWA Caching.} 
    This strategy does not store any SWA KV entries. 
    For a hitting prefix, we need to perform more recomputation to restore the SWA KV entries. 
    To be specific, in each attention layer, the SWA KV entry of each token depends on the SWA KV entries of only the most recent $n_{\text{win}}$ tokens from the previous layer.
    Therefore, leveraging cached \csa{} and \hca{} KV entries, recomputing the last $n_{\text{win}} \cdot L$ tokens is enough to restore the last $n_{\text{win}}$ SWA KV entries for an $L$-layer model. 
\end{itemize}
Depending on specific deployment scenarios, we select the most suitable strategy to achieve the desired trade-off between storage and computation.

\section{Pre-Training}
\label{sec:pre-training}

\subsection{Data Construction}

On top of the pre-training data of \dsviii{}, we endeavor to construct a more diverse and higher-quality training corpus with longer effective contexts. 
We continually refine our data construction pipelines. 
For web-sourced data, we implement filtering strategies to remove batched auto-generated and templated content, thereby mitigating the risk of model collapse~\citep{zhu2024synthesize}. 
Mathematical and programming corpora still remain core components of our training data, and we further enhance the coding capabilities of \dsviv{} series by incorporating agentic data during the mid-training phase. 
For multilingual data, we build a larger corpus for \dsviv{}, improving its capture of long-tail knowledge across different cultures. 
For \dsviv{}, we place a particular emphasis on long-document data curation, prioritizing scientific papers, technical reports, and other materials that reflect unique academic values. 
Combining all the above, our pre-training corpus comprises more than 32T tokens, containing mathematical contents, codes, web pages, long documents, and other high-quality categories.

For pre-training data, we largely follow the same pre-processing strategies of \dsviii{}. 
For tokenization, on top of the \dsviii{} tokenizer, we introduce a few special tokens for context construction, and still remain the vocabulary size to be 128K.  
We also inherit the token-splitting~\citep{dsv3} and Fill-in-Middle~(FIM)~\citep{dscodervii} strategies from \dsviii{}.
Inspired by \citet{Ding2024FewerTI}, we pack documents from different sources into appropriate sequences to minimize sample truncation.
Different from \dsviii{}, we employ sample-level attention masking during pre-training. 

\subsection{Pre-Training Setups}

\subsubsection{Model Setups}

\paragraph{\dsvivf{}.}
We set the number of Transformer layers to 43 and the hidden dimension $d$ to 4096. 
For the first two layers, we use pure sliding window attention. 
For the subsequent layers, \csa{} and \hca{} are used in an interleaved manner.
For \csa{}, we set the compression rate $m$ to 4, the number of indexer query heads $n_h^I$ to 64, the indexer head dimension $c^I$ to 128, and the number of KV entries selected for sparse attention (i.e., attention top-k) to 512.  
For \hca{}, we set the compression rate $m^{\prime}$ to 128. 
For both \csa{} and \hca{}, we set the number of query heads $n_h$ to 64, the head dimension $c$ to 512, and the query compression dimension $d_c$ to 1024. 
The number of output projection groups $g$ is set to 8, and the dimension of each intermediate attention output $d_g$ is set to 1024.
For the additional branch of sliding window attention, the window size $n_{\text{win}}$ is set to 128. 
We employ MoE layers in all Transformer blocks, but use the Hash routing strategy for the first 3 MoE layers. 
Each MoE layer consists of 1 shared expert and 256 routed experts, where the intermediate hidden dimension of each expert is 2048. 
Among the routed experts, 6 experts will be activated for each token. 
The multi-token prediction depth is set to 1. 
As for \mhc{}, the expansion factor $n_{\text{hc}}$ is set to 4, and the number of Sinkhorn-Knopp iterations $t_{\text{max}}$ is set to 20. 
Under this configuration, \dsvivf{} comprises \dsvivftp{} total parameters, of which \dsvivfap{} are activated for each token. 

\paragraph{\dsvivp{}.}
We set the number of Transformer layers to 61 and the hidden dimension $d$ to 7168. 
For the first two layers, we use \hca{}. 
For the subsequent layers, \csa{} and \hca{} are used in an interleaved manner.
For \csa{}, we set the compression rate $m$ to 4, the number of indexer query heads $n_h^I$ to 64, the indexer head dimension $c^I$ to 128, and the number of KV entries selected for sparse attention (i.e., attention top-k) to 1024.  
For \hca{}, we set the compression rate $m^{\prime}$ to 128. 
For both \csa{} and \hca{}, we set the number of query heads $n_h$ to 128, the head dimension $c$ to 512, and the query compression dimension $d_c$ to 1536. 
The number of output projection groups $g$ is set to 16, and the dimension of each intermediate attention output $d_g$ is set to 1024.
For the additional branch of sliding window attention, the window size $n_{\text{win}}$ is set to 128. 
We employ MoE layers in all Transformer blocks, but use the Hash routing strategy for the first 3 MoE layers. 
Each MoE layer consists of 1 shared expert and 384 routed experts, where the intermediate hidden dimension of each expert is 3072. 
Among the routed experts, 6 experts will be activated for each token. 
The multi-token prediction depth is set to 1. 
As for \mhc{}, the expansion factor $n_{\text{hc}}$ is set to 4, and the number of Sinkhorn-Knopp iterations $t_{\text{max}}$ is set to 20. 
Under this configuration, \dsvivp{} comprises \dsvivptp{} total parameters, of which \dsvivpap{} are activated for each token. 

\subsubsection{Training Setups}

\paragraph{\dsvivf{}.}
We employ the Muon optimizer~\citep{muon,muon_kimi} for the majority of parameters, but use the AdamW optimizer~\citep{adamW} for the embedding module, the prediction head module, and the weights of all RMSNorm modules. 
For AdamW, we set its hyper-parameters to $\beta_1=0.9$, $\beta_2=0.95$, $\epsilon=10^{-20}$, and $\mathrm{weight\_decay}=0.1$. 
For Muon, we set the momentum to 0.95 and the weight decay to 0.1, and rescale the RMS of each update matrix to 0.18 for reutilization of the AdamW learning rate. 
We train \dsvivf{} on \dsvivftoken{} tokens, and as in \dsviii{}, we also employ a batch size scheduling strategy that increases the batch size (in tokens) from a small size to 75.5M and then keeps it at 75.5M during most of the training. 
The learning rate is linearly warmed up in the first 2000 steps, maintained at $2.7 \times 10^{-4}$ for most of the training. 
Near the end of the training, we finally decay the learning rate to $2.7 \times 10^{-5}$ following a cosine schedule.
The training starts with a sequence length of 4K, and we gradually extend the training sequence length to 16K, 64K, and 1M. 
As for the setups of sparse attention, we first warmup the model with dense attention for the first 1T tokens, and introduce sparse attention at the sequence length of 64K and keep sparse attention during the rest of the training. 
When introducing attention sparsity, we first set a short stage to warm up the lightning indexer in \csa{}, and then train the model with sparse attention for most of the training. 
For auxiliary-loss-free load balancing, we set the bias update speed to 0.001. 
For the balance loss, we set its loss weight to 0.0001 to avoid extreme imbalance within single sequences. 
The MTP loss weight is set to 0.3 for most of the training, and to 0.1 upon the start of learning rate decay. 

\paragraph{\dsvivp{}.}
Except for specific values of hyper-parameters, the training setup of \dsvivp{} is largely consistent with that of \dsvivf{}.
We employ the Muon optimizer for the majority of parameters, but use the AdamW optimizer for the embedding module, the prediction head module, and the weights of all RMSNorm modules. 
The hyper-parameters of AdamW and Muon are the same as those of \dsvivf{}. 
We train \dsvivp{} on \dsvivptoken{} tokens, and also employ a batch size scheduling strategy, with the maximum batch size being 94.4M tokens.
The learning rate scheduling strategy is largely the same as that of \dsvivf{}, but the peak learning rate is set to $2.0 \times 10^{-4}$ and the end learning rate is set to $2.0 \times 10^{-5}$. 
The training also starts with a sequence length of 4K, and the length is gradually extended to 16K, 64K, and 1M. 
Compared with \dsvivf{}, \dsvivp{} starts with a longer stage of dense attention, and the strategy of introducing sparse attention is the same as \dsvivf{}, following a two-stage training method. 
For auxiliary-loss-free load balancing, we set the bias update speed to 0.001. 
For the balance loss, we set its loss weight to 0.0001 to avoid extreme imbalance within single sequences. 
The MTP loss weight is set to 0.3 for most of the training, and to 0.1 upon the start of learning rate decay. 

\subsubsection{Mitigating Training Instability}

Training trillion-parameter MoE models presents significant stability challenges, and \dsviv{} series are no exception. 
We encountered notable instability challenges during training. 
While simple rollbacks could temporarily restore the training state, they proved inadequate as a long-term solution because they do not prevent the recurrence of loss spikes.
Empirically, we identified that the occurrence of spikes is consistently tied to outliers in the MoE layers, and the routing mechanism itself appears to exacerbate the emergence of these outliers. 
Therefore, we sought to tackle this issue from two dimensions: breaking the vicious cycle induced by routing, and directly suppressing anomalous values. 
Fortunately, we discovered two practical techniques that effectively maintain training stability. 
Although a comprehensive theoretical understanding of their underlying mechanisms remains an open question for now, we are sharing them openly to foster further exploration by the community.

\paragraph{Anticipatory Routing.}
We found that decoupling the synchronous updates of the backbone network and the routing network significantly improves training stability. 
Consequently, at step $t$, we use the current network parameters $\theta_t$ for feature computation, but the routing indices are computed and applied using the historical network parameters $\theta_{t-\Delta t}$. 
In practice, to circumvent the overhead of loading model parameters twice, we fetch the data for step $t$ in advance at step $t-\Delta t$. 
We "anticipatorily" compute and cache the routing indices to be used later at step $t$, which is why we name this approach Anticipatory Routing. 
We also heavily optimized this at the infrastructure level. 
First, given that pre-computing the routing indices only requires a single forward pass over the data, we carefully orchestrated the pipeline execution and the overlapping of computation with Expert Parallelism (EP) communication, successfully bounding the additional wall-clock time overhead of Anticipatory Routing to approximately 20\%. 
Second, we introduced an automatic detection mechanism that triggers a short rollback and activates Anticipatory Routing exclusively when a loss spike occurs; after operating in this mode for a certain period, the system reverts to standard training. 
Ultimately, this dynamic application allows us to avert loss spikes with negligible overall additional training overhead, all without compromising model performance.

\paragraph{SwiGLU Clamping.}
In previous literature~\citep{bello*2017neural, gemma2}, clamping has been explicitly utilized to constrain numerical ranges, thereby enhancing training stability. 
In our actual training runs, we empirically found that applying SwiGLU clamping~\citep{gpt_oss} effectively eliminates outliers and substantially aids in stabilizing the training process, without compromising performance. 
Throughout the training of both \dsvivf{} and \dsvivp{}, we clamped the linear component of SwiGLU to the range of $[-10, 10]$, while capping the upper bound of the gate component at $10$.

\subsection{Evaluations}

\subsubsection{Evaluation Benchmarks}

For the evaluation of the base models, we consider benchmarks spanning four key dimensions: world knowledge, language understanding and reasoning, coding and mathematics, and long-context processing.

\textbf{World knowledge} benchmarks include 
AGIEval~\citep{agieval}, C-Eval~\citep{ceval}, CMMLU~\citep{cmmlu}
MMLU~\citep{mmlu}, MMLU-Redux~\citep{mmlu_redux}, MMLU-Pro~\citep{mmlu_pro}, MMMLU~\citep{mmmlu}, MultiLoKo~\citep{multiloko},
Simple-QA verified~\citep{haas2025simpleqa}, SuperGPQA~\citep{du2025supergpqa}, FACTS Parametric~\citep{cheng2025facts},
and TriviaQA~\citep{joshi-etal-2017-triviaqa}.

\textbf{Language understanding and reasoning} benchmarks include 
BigBench Hard (BBH)~\citep{bbh}, DROP~\citep{drop}, 
HellaSwag~\citep{hellaswag}, CLUEWSC~\citep{clue}, and WinoGrande~\citep{sakaguchi2019winogrande}.

\textbf{Coding and mathematical} benchmarks include
BigCodeBench~\citep{big_code_bench}, HumanEval~\citep{codex}, GSM8K~\citep{gsm8k}, MATH~\citep{hendrycks2021measuring}, MGSM \citep{mgsm}, and CMath~\citep{wei2023cmath}.

\textbf{Long context} benchmarks include LongBench-V2~\citep{bai2025longbench}. 

\begin{table}[!h]
    \centering
    \footnotesize
    \caption{
        Comparison among \dsviiiII{}-Base, \dsvivf{}-Base, and \dsvivp{}-Base.
        All models are evaluated in our internal framework and share the same evaluation setting.
        Scores with a gap not exceeding 0.3 are considered to be at the same level.
        The highest score in each row is in \textbf{bold font}, and the second is \underline{underlined}. 
    }
    \setlength{\tabcolsep}{4pt}
    \begin{tabular}{@{}c l c | c c c@{}}
    \toprule
    & \multirow{2}{*}{\centering \textbf{Benchmark {\tiny (Metric)}}} & \multirow{2}{*}{\textbf{\# Shots}} & \textbf{\dsviiiII{}} & \textbf{\dsvivf{}} & \textbf{\dsvivp{}} \\
    & & & \textbf{Base} & \textbf{Base} & \textbf{Base} \\
    \midrule
    & Architecture & - & MoE & MoE & MoE \\
    & \# Activated Params & - & 37B & \dsvivfap{} & \dsvivpap{} \\
    & \# Total Params & - & 671B & \dsvivftp{} & \dsvivptp{} \\
    \midrule
    \multirow{12}{*}{World Knowl.}
    & AGIEval {\tiny (EM)} & 0-shot & 80.1  & \underline{82.6} & \textbf{83.1} \\
    & MMLU {\tiny (EM)} & 5-shot & 87.8 & \underline{88.7} & \textbf{90.1} \\
    & MMLU-Redux {\tiny (EM)} & 5-shot & 87.5 & \underline{89.4} & \textbf{90.8} \\
    & MMLU-Pro {\tiny (EM)} & 5-shot & 65.5 & \underline{68.3} & \textbf{73.5} \\
    & MMMLU {\tiny (EM)} & 5-shot & 87.9 & \underline{88.8} & \textbf{90.3} \\
    & C-Eval {\tiny (EM)} & 5-shot & 90.4 & \underline{92.1} & \textbf{93.1} \\
    & CMMLU {\tiny (EM)} & 5-shot & 88.9 & \underline{90.4} & \textbf{90.8} \\
    & MultiLoKo {\tiny (EM)} & 5-shot & 38.7 & \underline{42.2} & \textbf{51.1} \\
    & Simple-QA verified {\tiny (EM)} & 25-shot & 28.3 & \underline{30.1} & \textbf{55.2} \\
    & SuperGPQA {\tiny (EM)} & 5-shot & 45.0 & \underline{46.5} & \textbf{53.9} \\
    & FACTS Parametric {\tiny (EM)} & 25-shot & 27.1 & \underline{33.9} & \textbf{62.6} \\
    & TriviaQA {\tiny (EM)} & 5-shot & \underline{83.3} & 82.8 & \textbf{85.6} \\
    \midrule
    \multirow{5}{*}{Lang. \& Reas.}
    & BBH {\tiny (EM)} & 3-shot & \textbf{87.6} & \underline{86.9} & \textbf{87.5} \\
    & DROP {\tiny (F1)} & 1-shot & \underline{88.2} & \textbf{88.6} & \textbf{88.7} \\
    & HellaSwag {\tiny (EM)} & 0-shot & \underline{86.4} & 85.7 & \textbf{88.0} \\
    & WinoGrande {\tiny (EM)} & 0-shot & 78.9 & \underline{79.5} & \textbf{81.5} \\
    & CLUEWSC {\tiny (EM)} & 5-shot & \underline{83.5} & 82.2 & \textbf{85.2} \\
    \midrule
    \multirow{6}{*}{Code \& Math}
    & BigCodeBench {\tiny (Pass@1)} & 3-shot & \textbf{63.9} & 56.8 & \underline{59.2} \\
    & HumanEval {\tiny (Pass@1)} & 0-shot & 62.8 & \underline{69.5} & \textbf{76.8} \\
    & GSM8K {\tiny (EM)} & 8-shot & \underline{91.1} & 90.8 & \textbf{92.6} \\
    & MATH {\tiny (EM)} & 4-shot & \underline{60.5} & 57.4 & \textbf{64.5} \\
    & MGSM {\tiny (EM)} & 8-shot & 81.3 & \textbf{85.7} & \underline{84.4} \\
    & CMath {\tiny (EM)} & 3-shot & \underline{92.6} & \textbf{93.6} & 90.9 \\
    \midrule
    \multirow{1}{*}{Long Context}
    & LongBench-V2 {\tiny (EM)} & 1-shot & 40.2 & \underline{44.7} & \textbf{51.5} \\
    \bottomrule
    \end{tabular}
    \label{tab:base}
\end{table}

\subsubsection{Evaluation Results}

In Table~\ref{tab:base}, we provide a detailed comparison of the base models for \dsviiiII{}, \dsvivf{}, and \dsvivp{}, all evaluated under a unified internal framework with strictly consistent settings.

Comparing \dsvivf{}-Base with \dsviiiII{}-Base reveals a compelling efficiency story. 
Despite utilizing a substantially smaller number of both activated and total parameters, \dsvivf{}-Base outperforms \dsviiiII{}-Base across a wide array of benchmarks. 
This advantage is especially evident in world knowledge tasks and challenging long-context scenarios. 
These results underscore that architectural improvements, refined data quality, and training optimizations in \dsvivf{}-Base yield superior performance even with a more compact parameter budget, effectively surpassing the larger \dsviiiII{}-Base on the majority of evaluations.

Furthermore, \dsvivp{}-Base demonstrates a further, decisive leap in capability, establishing near-universal dominance over both \dsviiiII{}-Base and \dsvivf{}-Base. 
With improvements across almost all categories, \dsvivp{}-Base reaches new performance highs among DeepSeek base models on the most demanding benchmarks.
On knowledge-intensive evaluations, it delivers dramatic gains, while also substantially advancing long-context understanding. 
On most reasoning and code benchmarks, \dsvivp{}-Base also exceeds both previous models. 
This comprehensive uplift confirms \dsvivp{}-Base as the strongest foundation model in the DeepSeek series, outperforming its predecessors across the spectrum of knowledge, reasoning, coding, and long-context capabilities.

\section{Post-Training}
\label{sec:post-training}

\subsection{Post-Training Pipeline}

Following pre-training, we conducted a post-training phase to yield the final models of \dsviv{} series. 
Although the training pipeline largely mirrored that of \dsviiiII{}, a critical methodological substitution was made: the mixed Reinforcement Learning (RL) stage was entirely replaced by On-Policy Distillation (OPD;~\citealp{lu2025onpolicydistillation,minillm}).

\subsubsection{Specialist Training}
The development of domain specialists was conducted by adapting the \dsviiiII{} training pipeline. 
Specifically, each model was sequentially optimized through an initial fine-tuning phase and subsequent Reinforcement Learning (RL) guided by domain-specific prompts and reward signals. 
For the RL stage, we implemented the Group Relative Policy Optimization (GRPO) algorithm, maintaining hyper-parameters closely aligned with our prior research~\citep{dsr1,dsv32}. 

\paragraph{Reasoning Efforts.} 
It is widely recognized that a model's performance on reasoning tasks is fundamentally governed by the computational effort expended. 
Consequently, we trained distinct specialist models under divergent RL configurations to facilitate the development of models optimized for varying reasoning capacities. 
As detailed in Table~\ref{tab:reasoning_modes}, \dsvivp{} and \dsvivf{} both support three specific reasoning effort modes. 
For each mode, we apply distinct length penalties and context windows during RL training, which results in varying output token lengths for reasoning.
To integrate these distinct reasoning modes, we utilize specialized response formats demarcated by the \texttt{<think>} and \texttt{</think>} tokens. 
Furthermore, for the "Think Max" mode, we prepend a specific instruction to the beginning of the system prompt to guide the model's reasoning process, as shown in Table~\ref{tab:reasoning_instruction}. 

\begin{table}[ht]
    \centering
    \caption{Comparison of three reasoning modes}
    \label{tab:reasoning_modes}
    \begin{tabular}{>{\raggedright\arraybackslash}p{2.5cm} p{3.5cm} p{3.5cm} p{3.5cm}}
    \toprule
    \textbf{Reasoning Mode} & \textbf{Characteristics} & \textbf{Typical Use Cases} & \textbf{Response Format} \\
    \midrule
    \textbf{Non-think} & 
    Fast, intuitive responses based on habits or simple rules. &
    Routine daily tasks, emergency reactions, low-risk decisions. &
     \texttt{</think>} summary \\
    \addlinespace
    \textbf{Think High} & 
    Conscious logical analysis, slower but more accurate. &
    Complex problem-solving, planning, medium-risk decisions. &
    \texttt{<think>} thinking tokens \texttt{</think>} summary\\
    \addlinespace
    \textbf{Think Max} & 
    Push reasoning to its fullest extent. Slow but powerful. &
    Exploring the boundary of model reasoning capability. &
    1. A special system prompt at the beginning. \newline 
     2. \texttt{<think>} thinking tokens \texttt{</think>} summary\\
    \bottomrule
    \end{tabular}
\end{table}

\begin{table}[ht]
    \centering
    \caption{Instruction injected into the system prompt for the "Think Max" mode. }
    \label{tab:reasoning_instruction}
    \begin{tcolorbox}[
      floatplacement=ht,
      title={Injected Instruction},
    ]
        Reasoning Effort: Absolute maximum with no shortcuts permitted. \\
        You MUST be very thorough in your thinking and comprehensively decompose the problem to resolve the root cause, rigorously stress-testing your logic against all potential paths, edge cases, and adversarial scenarios. \\
        Explicitly write out your entire deliberation process, documenting every intermediate step, considered alternative, and rejected hypothesis to ensure absolutely no assumption is left unchecked. 
    \end{tcolorbox}
\end{table}

\paragraph{Generative Reward Model.} 
Typically, easy-to-verify tasks can be effectively optimized using simple rule-based verifiers or test cases. 
In contrast, hard-to-verify tasks traditionally rely on Reinforcement Learning from Human Feedback (RLHF), which necessitates extensive human annotation to train a scalar reward model. 
In the post-training phase of \dsviv{} series, however, we dispense with these conventional scalar-based reward models. 
Instead, to address hard-to-verify tasks, we curate rubric-guided RL data and employ a Generative Reward Model (GRM) to evaluate policy trajectories. 
Crucially, we apply RL optimization directly to the GRM itself. 
In this paradigm, the actor network natively functions as the GRM, enabling the joint optimization of the model's evaluative (judging) proficiency alongside its standard generative capabilities. 
By unifying these roles, the model’s internal reasoning capabilities are inherently fused into its evaluative process, resulting in highly robust scoring. 
Furthermore, this approach achieves superior performance with only a minimal set of diverse human annotations, as the model leverages its own logic to generalize across complex tasks.

\begin{table}[ht]
    \centering
    \footnotesize
    \caption{Tool-call schema for \dsviv{} series. }
    \label{tab:tool_call_schema}
    \begin{tcolorbox}[
      floatplacement=ht,
      title={Tool Call Schema},
    ]
    \begin{lstlisting}
    ## Tools
    
    You have access to a set of tools to help answer the user's question. You can invoke tools by writing a "<|DSML|tool_calls>" block like the following:
    
    <|DSML|tool_calls>
    <|DSML|invoke name="$TOOL_NAME">
    <|DSML|parameter name="$PARAMETER_NAME" string="true|false">$PARAMETER_VALUE</|DSML|parameter>
    ...
    </|DSML|invoke>
    <|DSML|invoke name="$TOOL_NAME2">
    ...
    </|DSML|invoke>
    </|DSML|tool_calls>
    
    String parameters should be specified as is and set `string="true"`. For all other types (numbers, booleans, arrays, objects), pass the value in JSON format and set `string="false"`.
    
    If thinking_mode is enabled (triggered by <think>), you MUST output your complete reasoning inside <think>...</think> BEFORE any tool calls or final response.
    
    Otherwise, output directly after </think> with tool calls or final response.
    
    ### Available Tool Schemas
    
    {Tool Definition...}
    
    You MUST strictly follow the above definedtool name and parameter schemas to invoke tool calls.
    \end{lstlisting}
    \end{tcolorbox}
\end{table}

\paragraph{Tool-Call Schema and Special Token.}
Consistent with our previous version, we utilize a dedicated <think></think> tag to delineate the reasoning path. 
In \dsviv{} series, we introduce a new tool-call schema that employs a special "|DSML|" token and utilizes an XML-based format for tool invocations, as demonstrated in Table~\ref{tab:tool_call_schema}. 
Our experiments demonstrate that the XML format effectively mitigates escaping failures and reduces tool-call errors, providing a more robust interface for model-tool interactions.

\begin{figure}[h]
\centering
\includegraphics[width=0.6\textwidth]{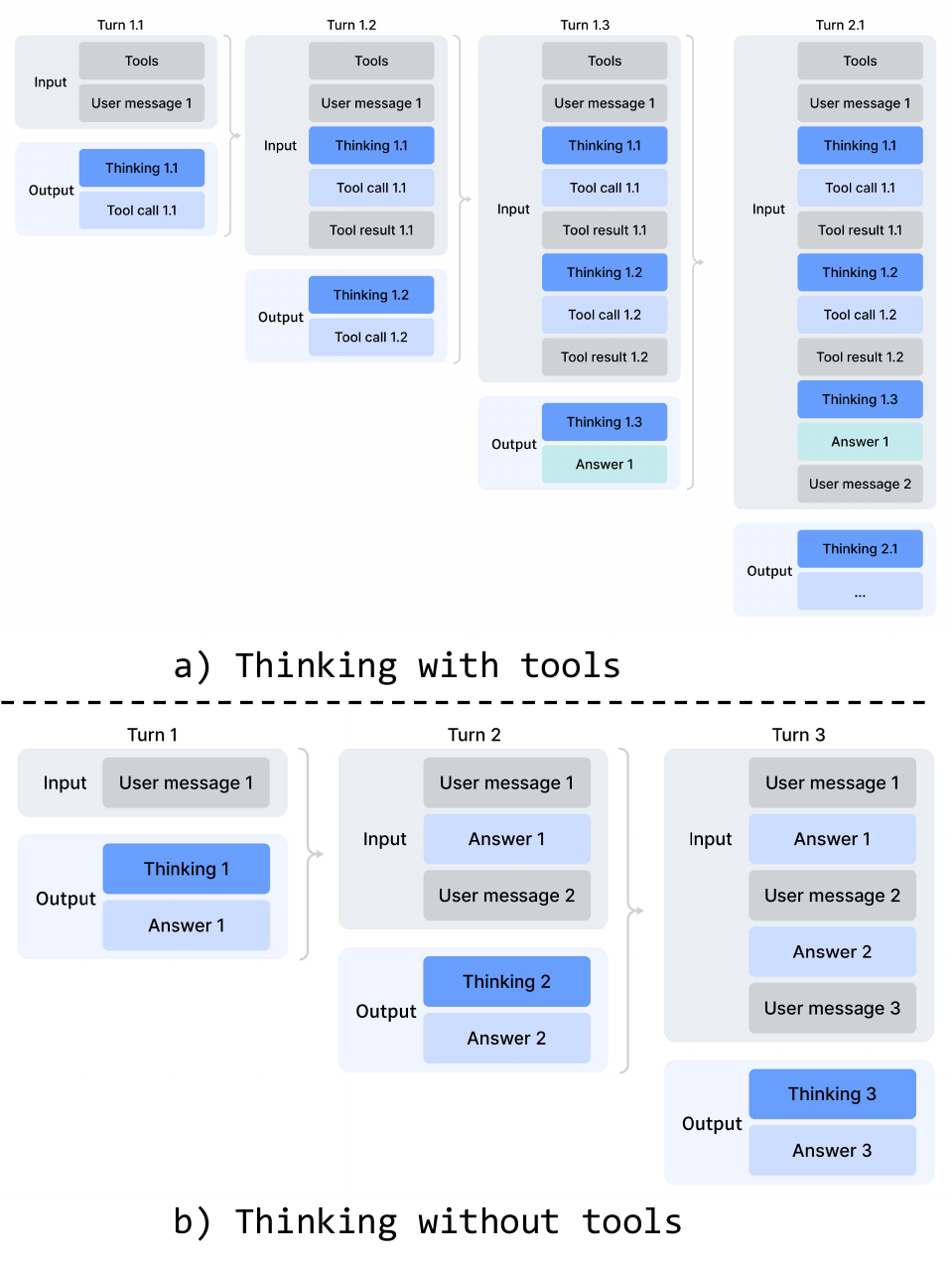}
\caption{
    \centering
    Thinking management of \dsviv{} series.
}
\label{fig:interleave_thinking}
\end{figure}

\paragraph{Interleaved Thinking.}
\dsviiiII{} introduced a context management strategy that retains reasoning traces across tool-result rounds but discards them upon the arrival of new user messages. 
While effective, this still caused unnecessary token waste in complex agentic workflows --- each new user turn would flush all accumulated reasoning content, forcing the model to reconstruct its problem-solving state from scratch.
Leveraging the expanded 1M-token context window of \dsviv{} series, we further refine this mechanism to maximize the effectiveness of interleaved thinking in agentic environments:
\begin{itemize}
    \item \textbf{Tool-Calling Scenarios.} 
    As illustrated in Figure~\ref{fig:interleave_thinking}(a), all reasoning content is fully preserved throughout the entire conversation. 
    Unlike \dsviiiII{}, which discarded thinking traces upon each new user turn, \dsviv{} series retain the complete reasoning history across all rounds, including across user message boundaries. 
    This allows the model to maintain a coherent, cumulative chain of thought over long-horizon agent tasks.
    \item \textbf{General Conversational Scenarios.} 
    As illustrated in Figure~\ref{fig:interleave_thinking}(b), the original strategy is preserved: reasoning content from previous turns is discarded when a new user message arrives, keeping the context concise for settings where persistent reasoning traces provide limited benefit.
\end{itemize}
As with \dsviiiII{}, agent frameworks that simulate tool interactions via user messages (e.g., Terminus) may not trigger the tool-calling context path and thus may not benefit from enhanced reasoning persistence. 
We continue to recommend non-think models for such architectures.

\begin{table}[h]
\centering
\caption{Quick Instruction special tokens for auxiliary tasks.}
\label{tab:quick_instruction}
\small
\begin{tabular}{p{3cm} p{4.5cm} p{6.5cm}}
\toprule
\textbf{Special Token} & \textbf{Description} & \textbf{Format} \\
\midrule
\texttt{<|action|>} & Determines whether the user prompt requires a web search or can be answered directly. & \texttt{...<|User|>\{prompt\}\allowbreak<|Assistant|>\allowbreak<think>\allowbreak<|action|>} \\
\addlinespace
\texttt{<|title|>} & Generates a concise conversation title after the first assistant response. & \texttt{...<|Assistant|>\{response\}\allowbreak<|end\_of\_sentence|>\allowbreak<|title|>} \\
\addlinespace
\texttt{<|query|>} & Generates search queries for the user prompt. & \texttt{...<|User|>\{prompt\}\allowbreak<|query|>} \\
\addlinespace
\texttt{<|authority|>} & Classifies the user prompt's demand for source authoritativeness. & \texttt{...<|User|>\{prompt\}\allowbreak<|authority|>} \\
\addlinespace
\texttt{<|domain|>} & Identifies the domain of the user prompt. & \texttt{...<|User|>\{prompt\}\allowbreak<|domain|>} \\
\addlinespace
\texttt{<|extracted\_url|>} \texttt{<|read\_url|>} & Determines whether each URL in the user prompt should be fetched and read. & \texttt{...<|User|>\{prompt\}\allowbreak<|extracted\_url|>\allowbreak\{url\}\allowbreak<|read\_url|>} \\
\bottomrule
\end{tabular}
\end{table}

\paragraph{Quick Instruction.}
In chatbot scenarios, a number of auxiliary tasks (e.g., determining whether to trigger a web search, intent recognition, etc.) must be executed before generating the response. Conventionally, these tasks are handled by a separate small model, requiring redundant prefilling since it cannot reuse the existing KV cache. To overcome this limitation, we introduce Quick Instruction. We append a set of dedicated special tokens directly to the input sequence, where each token corresponds to a specific auxiliary task. By directly reusing the already-computed KV cache, this mechanism completely avoids redundant prefilling and allows certain tasks, such as generating search queries and determining authority and domain, to be executed in parallel. Consequently, this approach significantly reduces the user-perceived time-to-first-token (TTFT) and eliminates the engineering overhead of maintaining and iterating an extra small model. The supported Quick Instruction tokens are summarized in Table~\ref{tab:quick_instruction}.

\subsubsection{On-Policy Distillation}

After training multiple domain-specific experts via specialized fine-tuning and reinforcement learning, we employ multi-teacher On-Policy Distillation (OPD;~\citealt{lu2025onpolicydistillation,minillm}) as the primary technique for merging expert capabilities into the final model. 
OPD has emerged as an effective post-training paradigm for efficiently transferring the knowledge and capabilities of domain experts to a single, unified model. 
This is achieved by having the student learn from the output distributions of teacher models on its own generated trajectories. 
Formally, given a set of $N$ expert models $\{\pi_{E_1}, \pi_{E_2}, \dots, \pi_{E_N}\}$, the OPD objective function is defined as:
\begin{equation}
    \mathcal{L}_{\text{OPD}}(\theta) = \sum_{i=1}^{N} w_i \cdot \text{D}_{\text{KL}} \left( \pi_{\theta}   \parallel \pi_{E_i}  \right).
\end{equation}
In this formulation, $w_i$ represents the assigned weight for each expert, typically determined by the relative importance of the expert.
Computing the reverse KL loss $\text{D}_{\text{KL}} \left( \pi_{\theta}   \parallel \pi_{E_i}  \right)$ requires sampling training trajectories from the student $\pi_\theta$ to maintain on-policy learning. 
The underlying logic ensures that the unified policy $\pi_{\theta}$ selectively learns from the specialized expert relevant to the current task context (e.g., aligning with the mathematics expert for math reasoning tasks and the coding expert for programming tasks). 
Through this mechanism, the knowledge from physically distinct expert weights is consolidated into a unified parameter space via logits-level alignment, practically circumventing the performance degradation often encountered in traditional weight-merging or mixed RL techniques. 
In this stage, more than ten teacher models covering various domains are employed to distill a single student model.

In handling the above OPD objective, prior works usually simplify the full-vocabulary KL loss into a token-level KL estimate at each token position, and reuse RL framework by replacing $\texttt{sg}\big[\log \frac{\pi_{E_i}(y_{t}|x,y_{<t})}{\pi_{\theta}(y_{t}|x,y_{<t})}\big]$ ($\texttt{sg}$ represents the stop gradient operation) as the per-token advantage estimate in the policy loss calculation. 
Although this approach is resource-efficient, it leads to high variance in gradient estimation and often causes training instability. 
Therefore, we adopt full-vocabulary logit distillation in our OPD. 
Preserving the complete logit distribution in calculating reverse KL loss yields more stable gradient estimates and ensures faithful distillation of the teachers' knowledge. 
In the following subsection, we describe the engineering efforts that make full-vocabulary OPD feasible at scale.

\subsection{Post-Training Infrastructures} 

Our post-training infrastructure is built upon the scalable framework developed for \dsviiiII{}. 
Specifically, we integrate the same distributed training stack described in Section~\ref{subsec:training-framework} and the rollout engine introduced earlier for efficient auto-regressive sampling. 
Building on this foundation, we introduce the following principal enhancements in the present work.
These designs enable efficient execution of ultra-long-context RL and OPD merging tasks involving over ten distinct teacher models, thereby substantially accelerating the iteration cycle for model releases.

\subsubsection{FP4 Quantization-Aware Training}
\label{subsec:fp4}
To achieve inference acceleration and reducing memory traffic at deployment, we introduce Quantization-Aware Training (QAT)~\citep{QAT} during the post-training stage, enabling the model, including those of teacher and reference models, to adapt to the precision degradation introduced by quantization. 
We apply FP4 (MXFP4) quantization~\citep{OCP_MXFormat} to two components: (1) MoE expert weights, which are a major source of GPU memory occupancy~\citep{gpt_oss}, and (2) the Query-Key (QK) path in the indexer of \csa{}, where QK activations are cached, loaded, and multiplied entirely in FP4, accelerating attention score computation in long-context scenarios.
In addition, we further quantize the index scores $I_{:, :}$ from FP32 to BF16 during this QAT process. 
This optimization achieves a 2$\times$ speedup for the top-k selector, while preserving a 99.7\% recall rate of KV entries.

For MoE expert weights, following the common practice of QAT, the FP32 master weights maintained by the optimizer are first quantized to FP4, then dequantized back to FP8 for computation. 
Notably, our FP4-to-FP8 dequantization is lossless. 
This is because FP8 (E4M3) has 2 additional exponent bits compared with FP4 (E2M1), offering a larger dynamic range. 
Consequently, as long as the ratio between the maximum and minimum scale factors of the FP4 sub-blocks ($1\times 32$ tiles) within each FP8 quantization block ($128\times 128$ tiles) does not exceed a certain threshold, the fine-grained scale information can be fully absorbed by the extended dynamic range of FP8.
We empirically verify that current weights satisfy this condition. 
This allows the entire QAT pipeline to fully reuse the existing FP8 training framework without any modification. 
In the backward pass, gradients are computed with respect to the same FP8 weights in the forward pass and directly propagated back to the FP32 master weights, equivalent to applying the Straight-Through Estimator (STE) through the quantization operation. 
This also avoids the need to re-quantize transposed weights.

During the inference and rollout phases of RL training, which do not involve backward passes, we directly use native FP4 quantized weights instead of simulated quantization. 
This ensures that model behavior during sampling is fully consistent with online deployment, while also reducing kernel memory loading for actual speedup and significantly lowering memory consumption.
We process the QK path in the indexer of \csa{} similarly.

\subsubsection{Efficient Teacher Scheduling for Full-Vocabulary OPD} 
Our framework supports full-vocabulary On-Policy Distillation (OPD) with an effectively unbounded number of teachers, each potentially comprising trillions of parameters. 
To enable this, all teacher weights are offloaded to a centralized distributed storage and are loaded on demand during the teacher forward pass with ZeRO-like parameter sharding to alleviate both I/O and DRAM pressure.
Furthermore, naively materializing logits for a vocabulary size $|V| > 100\text{k}$ across all teachers is prohibitive, even when spooled to disk.
We address this by caching only the last-layer teacher hidden states in a centralized buffer during the forward pass. 
At training time, these cached states are retrieved and passed through the corresponding prediction head module to reconstruct the full logits on the fly. 
This design incurs negligible recomputation overhead while completely circumventing the memory burden associated with explicit logits materialization. 
To mitigate the GPU memory footprint of the teacher prediction head, we order training samples by teacher index during data dispatching. 
This arrangement ensures that each distinct teacher head is loaded only once per mini-batch and that at most one teacher head resides in device memory at any given time. 
All parameters and hidden state loading/offloading operations proceed asynchronously in the background, without blocking computation on the critical path. 
Finally, the exact KL divergences between teacher and student logits are computed using a specialized TileLang kernel, which accelerates the computation and curtails dynamic memory allocation.

\subsubsection{Preemptible and Fault-Tolerant Rollout Service}
\label{subsec:inference-stream-caching}

To maximize GPU resource utilization while enabling rapid hardware provisioning for high-priority tasks, our GPU cluster employs a cluster-wide preemptive task scheduler, where any running task may be preempted at any time. 
Also, hardware failures are prevalent in large-scale GPU clusters. 
To this end, we implement a preemptible and fault-tolerant LLM generation service for RL/OPD rollout.

Specifically, we implement a token-granular Write-Ahead Log (WAL) for each generation request. 
Whenever a new token is generated for a request, we immediately append it to that request's WAL. 
During preemption, we pause the inference engine and save the KV cache of unfinished requests. 
Upon resumption, we use the persisted WALs and saved KV cache to continue decoding. 
Even when a fatal hardware error occurs, we can re-run the prefill phase using the persisted tokens in WAL to reconstruct the KV cache.

Importantly, it is mathematically incorrect to regenerate unfinished requests from scratch, as this introduces length bias. 
Because shorter responses are more likely to survive interruption, regenerating from scratch makes the model more prone to producing shorter sequences whenever an interruption occurs. 
If the inference stack is batch-invariant and deterministic, this correctness issue could also be addressed by regenerating with a consistent seed for the pseudorandom number generator used in the sampler. 
However, this approach still incurs the extra cost of re-running the decoding phase, making it far less efficient than our token-granular WAL method.

\subsubsection{Scaling RL Framework for Million-Token Context} 
We introduce targeted optimizations for efficient RL and OPD on million-token sequences. 
During the rollout phase, we adopt a preemptible and fault-tolerant rollout service, detailed in Section~\ref{subsec:inference-stream-caching}. 
For the inference and training phase, we decompose the rollout data format into lightweight metadata and heavy per-token fields. 
During data dispatching, the metadata for the entire rollout data can be loaded to perform global shuffling and packing layout computation. 
Heavy per-token fields are loaded via a shared-memory data loader to eliminate intra-node data redundancy and are released immediately upon consumption at the mini-batch granularity, substantially reducing both CPU and GPU memory pressure. 
The number of on-device mini-batches is dynamically determined based on workload, allowing an efficient trade-off between computational throughput and I/O overlap.

\subsubsection{Sandbox Infrastructure for Agentic AI} 
\label{sec:dsec} 
 
To meet the diverse execution demands of agentic AI during post-training and evaluation, we build a production-grade sandbox platform, \textbf{DeepSeek Elastic Compute (DSec)}. 
DSec comprises three Rust components --- the API gateway (\texttt{Apiserver}), per-host agent (\texttt{Edge}), and the cluster monitor (\texttt{Watcher}) --- that are interconnected by a custom RPC protocol and scale horizontally atop the 3FS distributed filesystem~\citep{ds3fs}. 
In production, a single DSec cluster manages hundreds of thousands of concurrent sandbox instances. 
 
The design of DSec is motivated by four observations:
(1) agentic workloads are highly heterogeneous, spanning lightweight function calls to full software-engineering pipelines with diverse OS and security requirements; 
(2) environment images are numerous and large, yet must load quickly and support iterative customization; 
(3) high-density deployment demands efficient CPU and memory utilization; 
(4) sandbox lifecycles must coordinate with GPU training schedules, including preemption and checkpoint-based resumption. 
Based on these observations, we elaborate on the four core designs of DSec individually in the following.

\paragraph{Four Execution Substrates Behind One Unified Interface.}
DSec exposes a single Python SDK (\texttt{libdsec}) that abstracts four execution substrates. 
\textbf{Function Call} dispatches stateless invocations to a pre-warmed container pool, eliminating cold-start overhead.
\textbf{Container} is fully Docker-compatible and leverages EROFS \citep{erofs} on-demand loading for efficient image assembly.
\textbf{microVM}, built on Firecracker~\citep{firecracker}, adds VM-level isolation for security-sensitive, high-density deployments.
\textbf{fullVM}, built on QEMU~\citep{qemu}, supports arbitrary guest operating systems. 
All four share a common API surface --- command execution, file transfer, and TTY access --- and switching between them requires only a parameter change.

\paragraph{Fast Image Loading via Layered Storage.} 
DSec reconciles fast startup with a large and growing corpus of environment images through layered, on-demand loading. 
For containers, base images and filesystem commits are stored as 3FS-backed readonly EROFS layers mounted directly into overlay \texttt{lowerdir}s. 
We keep file metadata readily available on the local disk at mount time; meanwhile, data blocks are fetched from 3FS upon request.
For microVMs, DSec uses the \texttt{overlaybd}~\citep{overlaybd} disk format: the read-only base layer resides on 3FS for cross-instance sharing, while writes go to a local copy-on-write layer. 
Such snapshots are chainable, facilitating efficient versioning and millisecond-scale resumption.

\paragraph{Density Optimizations Under Massive Concurrency.} 
To accommodate hundreds of thousands of sandboxes per cluster, DSec tackles two resource bottlenecks.
First, it mitigates duplicate page-cache footprints in virtualized environments and applies memory reclamation to enable safe overcommitment. 
Second, it alleviates spinlock contention in the container runtime and therefore, reduces per-sandbox CPU overhead, significantly increasing per-host packing density. 

\paragraph{Trajectory Logging and Preemption-Safe Resumption.} 
DSec maintains a globally ordered trajectory log for each sandbox, persistently recording every command invocation and its results. 
The trajectory serves three purposes: 
(1) \textbf{client fast-forwarding} --- when a training task is preempted, sandbox resources are retained nonetheless; upon resumption, DSec replays cached results for previously completed commands, accelerating task recovery whilst also preventing errors from re-execution of non-idempotent operations;
(2) \textbf{fine-grained provenance} --- the origin and corresponding outcomes of each state change are traceable;
(3) \textbf{deterministic replay} --- any historical session can be faithfully reproduced from its trajectory.

\subsection{Standard Benchmark Evaluation}

\subsubsection{Evaluation Setup}

\paragraph{Knowledge and Reasoning.} 
Knowledge and reasoning datasets include MMLU-Pro~\citep{mmlu_pro}, GPQA~\citep{gpqa}, Human Last Exam~\citep{hle}, Simple-QA Verified~\citep{haas2025simpleqa}, Chinese-SimpleQA~\citep{csimpleqa}, LiveCodeBench-v6~\citep{jain2024livecodebench}, CodeForces~(Internal Benchmark), HMMT 2026 Feb, Apex~\citep{balunovic2025matharena}, Apex Shortlist~\citep{balunovic2025matharena}, IMOAnswerBench~\citep{luong-etal-2025-towards}, and PutnamBench~\citep{tsoukalas2024putnambenchevaluatingneuraltheoremprovers}.

For code, we evaluate \dsviv{} series on LiveCodeBench-v6 and an internal Codeforces benchmark. 
For Codeforces, we collect 14 Codeforces Division 1 contests comprising 114 problems (May 2025 - November 2025). 
The Elo rating is computed as follows. 
For each contest, we generate 32 candidate solutions per problem. 
For each problem independently, we sample 10 of these solutions without replacement and arrange them in a random order to form the submission sequence. 
Each submission is judged against a test suite constructed by domain experts. 
The score for a solved problem follows the penalty scheme of OpenAI (2025): the model receives the median score of human participants who solved the same problem with the same number of prior failed attempts. 
This yields a total contest score for each sampled submission sequence, which is then converted into a contest rank and subsequently into an estimated rating via the standard Codeforces rating system. 
The contest-level expected rating is defined as the expectation of this estimated rating over all possible random selections and orderings of the 10 submissions per problem. 
The model's overall rating is the average of these contest-level expected ratings across all 14 contests.

For reasoning and knowledge tasks, we set the temperature to 1.0 and the context window to 8K, 128K, and 384K tokens for the Non-think, High, and Max modes, respectively. 
For math tasks (e.g., HMMT, IMOAnswerBench, Apex, and HLE), we evaluate using the following template: \texttt{"\{question\}\textbackslash nPlease reason step by step, and put your final answer within \textbackslash boxed\{\}."}
For \dsvivp{}-Max on math tasks, we use the following template to elicit deeper reasoning: \texttt{"Solve the following problem. The problem may ask you to prove a statement, or ask for an answer. If finding an answer is required, you should come up with the answer, and your final solution should also be a rigorous proof of that answer being valid.\textbackslash n\textbackslash n\{question\}"}.

For formal math tasks, we evaluate in an agentic setting on Lean v4.28.0-rc1~\citep{moura2021lean}, with access to the Lean compiler and a semantic tactic search engine, running up to 500 tool calls with max reasoning effort. In addition, we evaluate a more compute-intensive pipeline in which candidate natural-language solutions are first generated and filtered by self-verification\citep{shao2025deepseekmathv2selfverifiablemathematicalreasoning}, and the retained solutions are then provided as guidance to a formal agent for proving the corresponding Lean statement. This design uses informal reasoning to improve exploration while preserving strict correctness through formal verification.
A submission is counted as correct only if the strict verifier Comparator accepts it for both settings. 

We have left some entries blank for K2.6 and GLM-5.1, as their APIs were too busy to return responses to our queries.

\paragraph{1M-Token Context.} 
Since \dsviv{} series supports 1M-token contexts, we evaluate model performance in a long context scenario by selecting OpenAI MRCR~\citep{mrcr} and CorpusQA~\citep{lu2026corpusqa} as the benchmarks. 
We re-evaluate Claude Opus 4.6 and Gemini 3.1 Pro on these tasks with the goal of standardizing the configuration across all models. 
We did not evaluate GPT-5.4 because its API failed to respond to a large portion of our queries.

\paragraph{Agent.} 
Agent datasets include Terminal Bench 2.0~\citep{merrill2026terminal}, SWE-Verified~\citep{swe_verified}, SWE Multilingual~\citep{yang2025swesmith}, SWE-Pro~\citep{deng2025swebenchproaiagents}, BrowseComp~\citep{wei2025browsecomp}, the public evaluation set of MCPAtlas~\citep{bandi2026mcp}, GDPval-AA \citep{patwardhan2025gdpval,gdpaa}, and Tool-Decathlon~\citep{li2025tool}. 

For code agent tasks (SWE-Verified, Terminal-Bench, SWE-Pro, SWE Multilingual), we evaluate \dsviv{} series using an internally developed evaluation framework. 
This framework provides a minimal set of tools  ---  a bash tool and a file-edit tool. 
The maximum number of interaction steps is set to 500, and the maximum context length is set to 512K tokens. 
Regarding Terminal-Bench 2.0, we acknowledge the environment-related issues noted by GLM-5.1. 
Nevertheless, we report our performance on the original Terminal-Bench 2.0 dataset for consistency. 
On the Terminal-Bench 2.0 Verified subset, \dsvivp{} achieves a score of approximately 72.0.

For search agent tasks (BrowseComp, HLE w/ tool), we also use an in-house harness with websearch and Python tool, and set maximum interaction steps to 500 and the maximum context length to 512K tokens. 
For BrowseComp, we use the same discard-all context management strategy as \dsviiiII~\citep{dsv32}.

\subsubsection{Evaluation Results}

\begin{table}[ht] 
    \centering
    \footnotesize
    \setlength{\tabcolsep}{1.2pt}
    \caption{
        Comparison between \dsvivpm{} and closed/open source models.
        "Max", "xHigh", and "High" denote reasoning effort. 
        The best results are highlighted in bold; the second-best results are underlined.
    }
    \label{tab:large_eval} 
    \resizebox{\textwidth}{!}{
    \begin{tabular}{@{}c l | c c c | c c c  c @{}} 
    \toprule

    & \multirow{2}{*}{\centering \textbf{Benchmark {\tiny (Metric)}}} & \textbf{Opus-4.6} & \textbf{GPT-5.4} & \textbf{Gemini-3.1-Pro} & \textbf{K2.6} & \textbf{GLM-5.1} & \textbf{DS-V4-Pro} \\ 
    & & \textbf{Max} & \textbf{xHigh} & \textbf{High} & \textbf{Thinking} & \textbf{Thinking} & \textbf{Max}  \\ 
  
    % ========================

    \midrule

    \multirow{11}{*}{\rotatebox{90}{Knowledge \& Reasoning}} 
    
    & MMLU-Pro {\tiny (EM)}               & \underline{89.1} & 87.5             & \textbf{91.0} & 87.1 & 86.0 & 87.5             \\ 
    & SimpleQA-Verified {\tiny (Pass@1)}  & 46.2             & 45.3             & \textbf{75.6} & 36.9 & 38.1 & \underline{57.9} \\ 
    & Chinese-SimpleQA {\tiny (Pass@1)}   & 76.4             & 76.8             & \textbf{85.9} & 75.9 & 75.0 & \underline{84.4} \\ 
    & GPQA Diamond {\tiny (Pass@1)}       & 91.3             & \underline{93.0} & \textbf{94.3} & 90.5 & 86.2 & 90.1             \\ 
    & HLE {\tiny (Pass@1)}                & \underline{40.0} & 39.8             & \textbf{44.4} & 36.4 & 34.7 & 37.7             \\ 
    & LiveCodeBench {\tiny (Pass@1)}      & 88.8             & -                & \underline{91.7}& 89.6 & -    & \textbf{93.5}    \\ 
    & Codeforces {\tiny (Rating)}         & -                & \underline{3168} & 3052          & -    & -    & \textbf{3206}    \\ 
    & HMMT 2026 Feb {\tiny (Pass@1)}      & \underline{96.2}& \textbf{97.7}    & 94.7          & 92.7 & 89.4 & {95.2} \\ 
    & IMOAnswerBench {\tiny (Pass@1)}     & 75.3             & \textbf{91.4}    & 81.0          & 86.0 & 83.8 & \underline{89.8} \\ 
    & Apex {\tiny (Pass@1)}               & 34.5             & \underline{54.1} & \textbf{60.9} & 24.0 & 11.5 & 38.3 \\
    & Apex Shortlist {\tiny (Pass@1)}     & 85.9 & 78.1   & \underline{89.1} & 75.5 & 72.4 & \textbf{90.2}             \\ 
    \midrule

     \multirow{2}{*}{\rotatebox{90}{Long}} 
    & MRCR 1M {\tiny (MMR)}  & \textbf{92.9} & - & 76.3 & - & - & \underline{83.5}  \\ 
    & CorpusQA 1M {\tiny (ACC)}  & \textbf{71.7} & - & 53.8 & - & - & \underline{62.0} & \\ 
    \midrule
    
    \multirow{10}{*}{\rotatebox{90}{Agentic}} 
    & Terminal Bench 2.0 {\tiny (Acc)}  & 65.4 & \textbf{75.1} & \underline{68.5} & 66.7 & 63.5 & 67.9 & \\ 
    & SWE Verified {\tiny (Resolved)}  & \textbf{80.8} & - & \underline{80.6} & 80.2 & -  & \underline{80.6} & \\ 
    & SWE Pro {\tiny (Resolved)}  & 57.3 & 57.7 & 54.2 & \textbf{58.6} & \underline{58.4} & 55.4 & \\ 
    & SWE Multilingual {\tiny (Resolved)}  & \textbf{77.5} & - & - & \underline{76.7} & 73.3 & \underline{76.2} & \\  
    & BrowseComp {\tiny (Pass@1)}  & \underline{83.7} & 82.7 & \textbf{85.9} & 83.2 & 79.3 & 83.4 & \\ 
    & HLE w/ tools {\tiny (Pass@1)}  & \underline{53.1} & 52.0 & 51.6  & \textbf{54.0} & 50.4 & 48.2 & \\ 
    & GDPval-AA {\tiny (Elo)} &\underline{1619}& \textbf{1674}&1314&1482&1535&1554 \\ 
    & MCPAtlas Public{\tiny (Pass@1)} & \textbf{73.8} & 67.2 & 69.2 & 66.6  & 71.8 & \underline{73.6} \\ 
    & Toolathlon {\tiny (Pass@1)}  & 47.2 & \textbf{54.6} & 48.8 & 50.0 & 40.7 & \underline{51.8} & \\
      
    \bottomrule
    \end{tabular}} 
\end{table}

\begin{table}[ht]
    \centering
    \small
    \setlength{\tabcolsep}{8pt}
    \caption{
        Comparison among different sizes and modes of \dsviv{} series.
        "Non-Think", "High", and "Max" denote reasoning effort. 
    }
    \label{tab:small_eval}
    \begin{tabular}{@{} c l | c c c | c c c @{}}
        \toprule
        & \multirow{2}{*}{\centering \textbf{Benchmark {\tiny (Metric)}}}
        & \multicolumn{3}{c|}{\textbf{\dsvivf{}}}
        & \multicolumn{3}{c}{\textbf{\dsvivp{}}} \\
        & & \textbf{Non-Think} & \textbf{High} & \textbf{Max}
          & \textbf{Non-Think} & \textbf{High} & \textbf{Max} \\
        \midrule
        \multirow{11}{*}{\rotatebox{90}{Knowledge \& Reasoning}}
        & MMLU-Pro {\tiny (EM)}               & 83.0 & 86.4 & 86.2 & 82.9 & 87.1 & 87.5 \\
        & SimpleQA-Verified {\tiny (Pass@1)}  & 23.1 & 28.9 & 34.1 & 45.0 & 46.2 & 57.9 \\
        & Chinese-SimpleQA {\tiny (Pass@1)}   & 71.5 & 73.2 & 78.9 & 75.8 & 77.7 & 84.4 \\
        & GPQA Diamond {\tiny (Pass@1)}       & 71.2 & 87.4 & 88.1 & 72.9 & 89.1 & 90.1 \\
        & HLE {\tiny (Pass@1)}                & 8.1  & 29.4 & 34.8 & 7.7  & 34.5 & 37.7 \\
        & LiveCodeBench {\tiny (Pass@1-COT)}  & 55.2 & 88.4 & 91.6 & 56.8 & 89.8 & 93.5 \\
        & Codeforces {\tiny (Rating)}         & -    & 2816 & 3052 & -    & 2919 & 3206 \\
        & HMMT 2026 Feb {\tiny (Pass@1)}      & 40.8 & 91.9 & 94.8 & 31.7 & 94.0 & 95.2 \\
        & IMOAnswerBench {\tiny (Pass@1)}     & 41.9 & 85.1 & 88.4 & 35.3 & 88.0 & 89.8 \\
        & Apex {\tiny (Pass@1)}               & 1.0  & 19.1 & 33.0 & 0.4  & 27.4 & 38.3 \\
        & Apex Shortlist {\tiny (Pass@1)}     & 9.3  & 72.1 & 85.7 & 9.2  & 85.5 & 90.2 \\
        \midrule
        \multirow{2}{*}{\rotatebox{90}{Long}}
        & MRCR 1M{\tiny (MMR)}   & 37.5 & 76.9 & 78.7 & 44.7 & 83.3 & 83.5 \\
        & CorpusQA 1M{\tiny (ACC)} & 15.5 & 59.3 & 60.5 & 35.6 & 56.5 & 62.0 \\
        \midrule
        \multirow{10}{*}{\rotatebox{90}{Agentic}}
        & Terminal Bench 2.0 {\tiny (Acc)} & 49.1 & 56.6 & 56.9 & 59.1 & 63.3 & 67.9 \\
        & SWE Verified {\tiny (Resolved)}  & 73.7 & 78.6 & 79.0 & 73.6 & 79.4 & 80.6 \\
        & SWE Pro {\tiny (Resolved)}       & 49.1 & 52.3 & 52.6 & 52.1 & 54.4 & 55.4 \\
        & SWE Multilingual {\tiny (Resolved)} & 69.7 & 70.2 & 73.3 & 69.8 & 74.1 & 76.2 \\
        & BrowseComp {\tiny (Pass@1)}      & -    & 53.5 & 73.2 & -    & 80.4 & 83.4 \\
        & HLE w/ tools {\tiny (Pass@1)}    & -    & 40.3 & 45.1 & -    & 44.7 & 48.2 \\
        & MCPAtlas Public {\tiny (Pass@1)}       & 64.0 & 67.4 & 69.0 & 69.4 & 74.2 & 73.6 \\
        & GDPval-AA {\tiny (Elo)}          &   -   &   -   & 1395 &   -   &   -   & 1554 \\
        & Toolathlon {\tiny (Pass@1)}      & 40.7 & 43.5 & 47.8 & 46.3 & 49.0 & 51.8 \\
        \bottomrule
    \end{tabular}
\end{table}

The comparison of \dsvivpm{} and other closed/open source models is presented in Table~\ref{tab:large_eval}.
Also, we evaluate different modes of \dsvivf{} and \dsvivp{} and show the results in Table~\ref{tab:small_eval}.

\begin{figure*}[t]
    \centering
    \begin{minipage}[t]{0.48\textwidth}
    \vspace{0pt}
    \centering
    \textbf{Practical Regime}\par
    \vspace{0.25em}
    \small Putnam-200 Pass@8 with minimal tools\\and bounded sampling.\par
    \vspace{0.8em}
    \includegraphics[width=\linewidth]{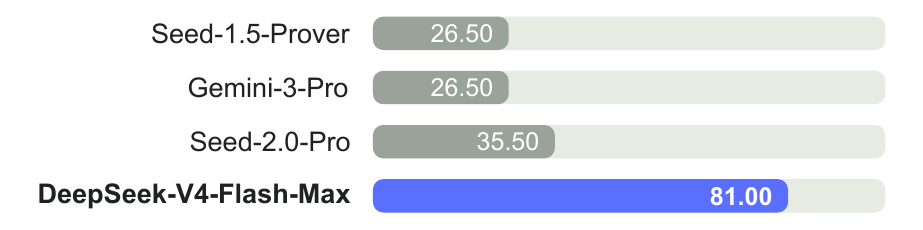}
    \end{minipage}\hfill
    \begin{minipage}[t]{0.48\textwidth}
    \vspace{0pt}
    \centering
    \textbf{Frontier Regime}\par
    \vspace{0.25em}
    \small Putnam-2025 with hybrid formal-informal reasoning and substantial compute scaling.\par
    \vspace{0.8em}
    \includegraphics[width=\linewidth]{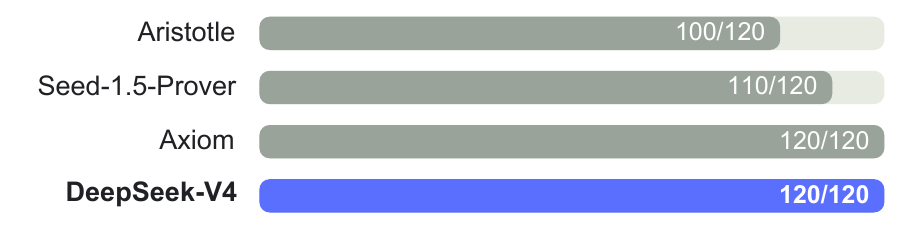}
    \end{minipage}
    \caption{Formal reasoning under practical and frontier regimes. Left: Putnam-200 Pass@8 evaluates a fixed random subset of PutnamBench~\citep{tsoukalas2024putnambenchevaluatingneuraltheoremprovers} following the setup introduced by Seed-Prover; all models are tested on the same problem set. We follow the Seed-Prover protocol but replace proprietary search tools with the open-source LeanExplore~\citep{Asher_LeanExplore_2025}, yielding a lightweight setting with minimal agent tools and bounded sampling. Right: Putnam-2025 probes the frontier of mathematical reasoning in a scaled hybrid formal-informal regime, where informal reasoning is combined with formal verification to expose gaps and improve rigor; DeepSeek-V4 reaches a proof-perfect 120/120.}

    \label{fig:formal-reasoning-overview}
\end{figure*}

\paragraph{Knowledge.}
In the evaluation of general world knowledge, \dsvivpm{}, the maximum reasoning effort mode of \dsvivp{}, establishes a new state-of-the-art among open-source large language models. 
As demonstrated by the SimpleQA-Verified, \dsvivpm{} significantly outperforms all existing open-source baselines by a margin of 20 absolute percentage points. 
Despite these advances, it currently trails the leading proprietary model, Gemini-3.1-Pro. 
In the domain of educational knowledge and reasoning, \dsvivpm{} marginally outperforms Kimi and GLM across the MMLU-Pro, GPQA, and HLE benchmarks, although it lags behind leading proprietary models. 
Broadly, \dsvivpm{} marks a significant milestone in enhancing the world knowledge capabilities of open-source models.

In addition, a significant performance gap exists between \dsvivf{} and \dsvivp{} on knowledge-based tasks; this is anticipated, as larger parameter counts facilitate greater knowledge retention during pre-training. 
Notably, both models demonstrate improved results on knowledge benchmarks when allocated higher reasoning effort. 

\paragraph{Reasoning.}
\dsvivpm{} outperforms all prior open models across reasoning benchmarks, and matches state-of-the-art closed models on many metrics, while the smaller \dsvivfm{} also surpasses the previous best open-source model, K2.6-Thinking, on code and math reasoning tasks.
Meanwhile, \dsvivp{} and \dsvivf{} excel in coding competitions. According to our evaluation, their performance is comparable to GPT-5.4, making this the first time an open model has matched a closed model on this task. On the Codeforces leaderboard, \dsvivpm{} currently ranks 23rd among human candidates.
\dsviv{} also demonstrates strong performance on formal mathematical task under both agentic and compute-intensive settings. Under an agentic setup, it achieves state-of-the-art results, shown in Figure \ref{fig:formal-reasoning-overview}, outperforming prior models such as Seed Prover \citep{chen2025seedprover15masteringundergraduatelevel}. With a more compute-intensive pipeline, performance further improves, surpassing systems including Aristotle\citep{achim2025aristotle} and matching the best known results under this setting.

\paragraph{Agent.}
The \dsviv{} series demonstrates strong agent performance in evaluations. 
For code agent tasks, \dsvivp{} achieves results comparable to K2.6 and GLM-5.1, though all these open models still lag behind their closed-source counterparts. 
\dsvivf{} underperforms \dsvivp{} on coding tasks, particularly on Terminal Bench 2.0. 
A similar trend is observed across other agent evaluations. It is worth noting that \dsvivp{} performs well on MCPAtlas and Toolathlon—two evaluation test sets that include a wide range of tools and MCP services—indicating that our model has excellent generalization capability and does not perform well only on internal frameworks.

\begin{figure}[h]
\centering
\includegraphics[width=0.8\textwidth]{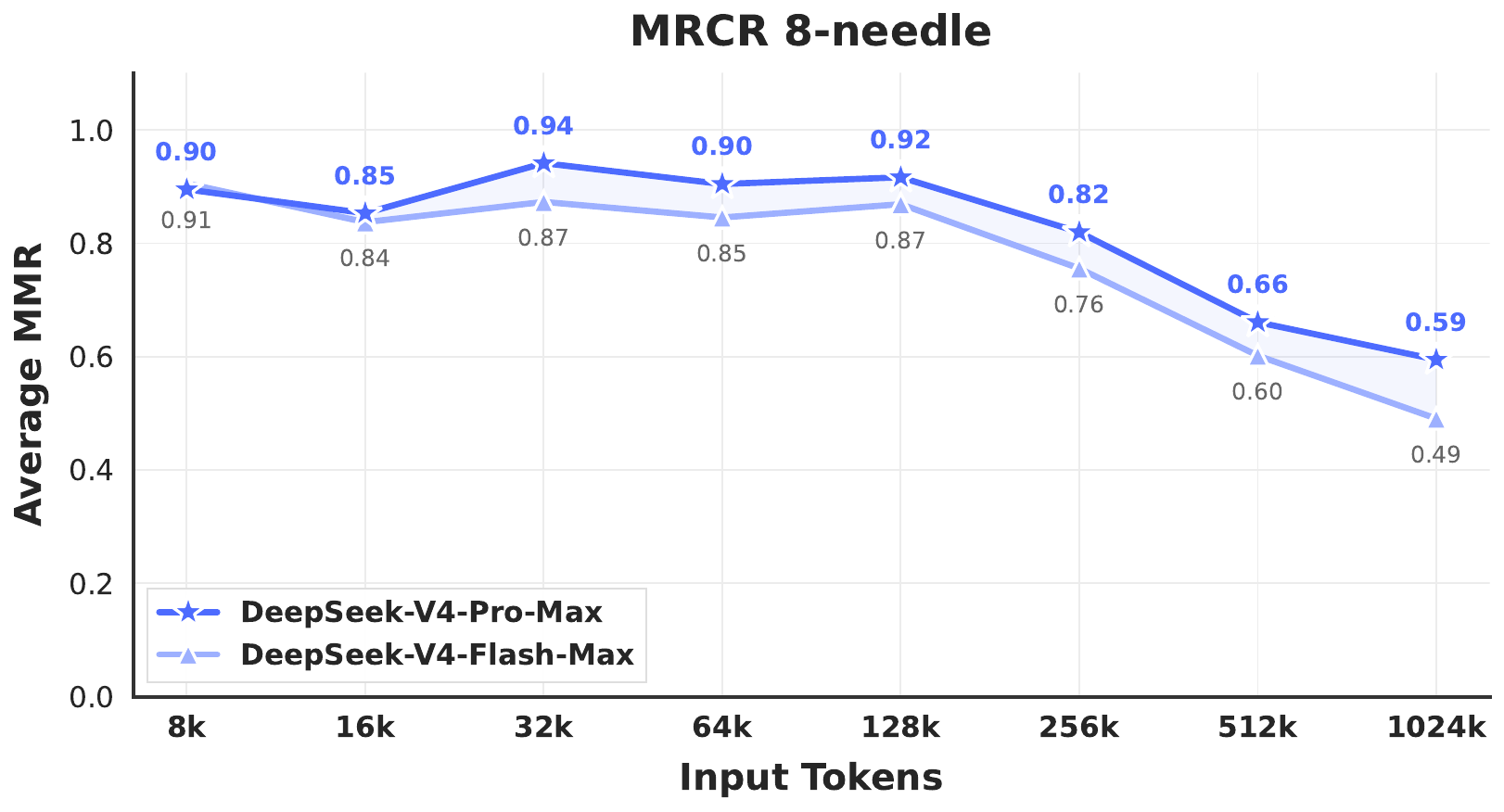}
\caption{
\dsviv{} series performance on the MRCR task.
}
\label{fig:mrcr}
\end{figure}

\paragraph{1M-Token Context.}
\dsvivp{} outperforms Gemini-3.1-Pro on the MRCR task, which measures in-context retrieval, but remains behind Claude Opus 4.6. 
As illustrated in Figure \ref{fig:mrcr}, retrieval performance remains highly stable within a 128K context window. 
While a performance degradation becomes visible beyond the 128K mark, the model’s retrieval capabilities at 1M tokens remain remarkably strong compared to both proprietary and open-source counterparts. 
Unlike MRCR, CorpusQA is similar to real scenarios. 
The evaluation results also indicate that \dsvivp{} is better than Gemini-3.1-Pro.

\begin{figure}[h]
\centering
\includegraphics[width=0.9\textwidth]{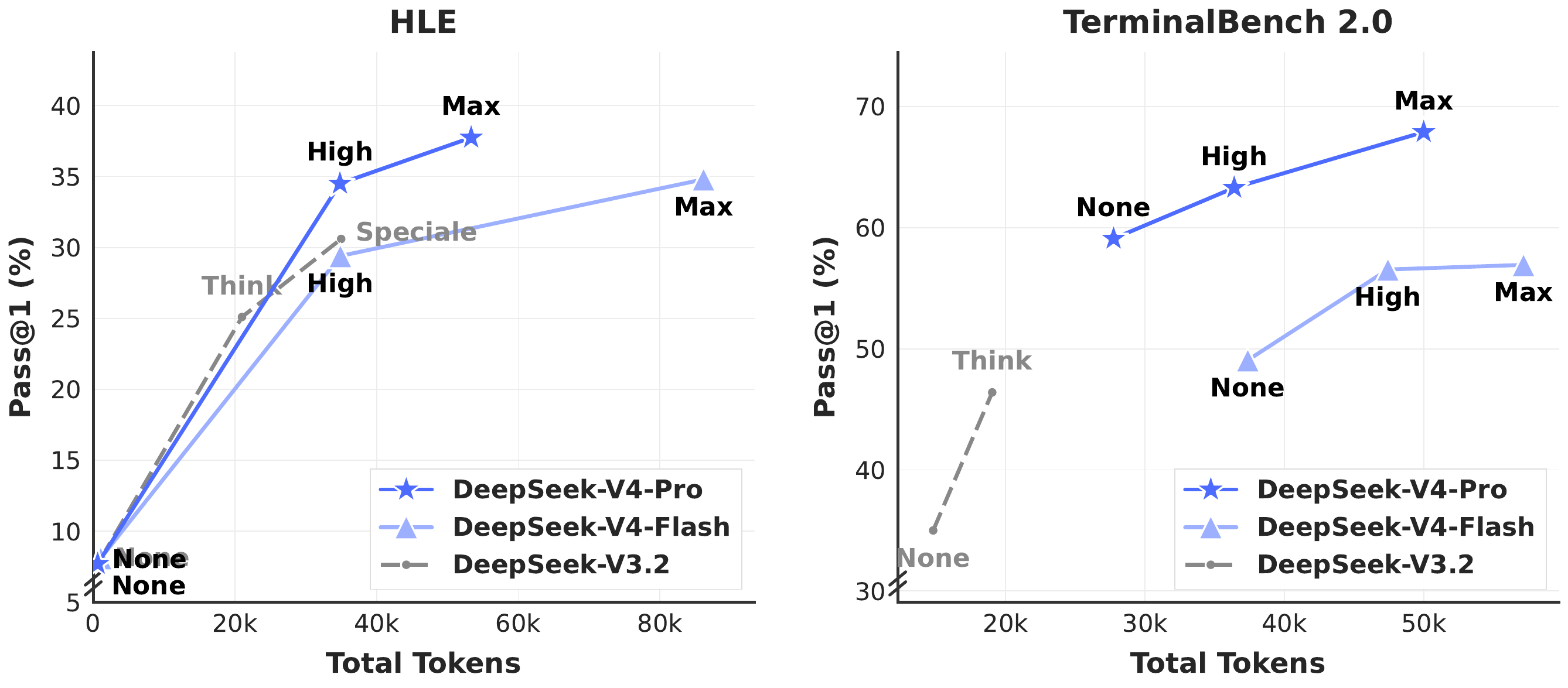}
\caption{
    HLE and Terminal Bench 2.0 performance by reasoning effort. ``None'' indicates Non-think mode, and ``Speciale'' indicates DeepSeek-V3.2-Speciale model.
}
\label{fig:dsv4_effort}
\end{figure}

\paragraph{Reasoning Effort.}
As shown in Table~\ref{tab:small_eval}, the Max mode, which employs longer contexts and reduced length penalties in RL, outperforms the High mode on the most challenging tasks. 
Figure~\ref{fig:dsv4_effort} presents a comparison of performance and cost among \dsvivp{}, \dsvivf{}, and \dsviiiII{} on representative reasoning and agentic tasks. 
By scaling test-time compute, \dsviv{} series achieve substantial improvements over the predecessor. 
Furthermore, on reasoning tasks like HLE, \dsvivp{} demonstrates higher token efficiency than \dsviiiII{}.

\subsection{Performance on Real-World Tasks}

Standardized benchmarks often struggle to capture the complexities of diverse, real-world tasks, creating a gap between test results and actual user experience. 
To bridge this, we have developed proprietary internal metrics that prioritize real-world usage patterns over traditional benchmarks. 
This approach ensures that our optimizations translate into tangible benefits. 
Our evaluation framework specifically targets the primary use cases of the DeepSeek API and Chatbot, aligning model performance with practical demands.

\subsubsection{Chinese Writing}

One of the primary use cases for DeepSeek is Chinese writing. 
We conducted a rigorous evaluation on functional writing and creative writing. 
Table~\ref{tab:funceval_conclusion} presents a pairwise comparison between \dsvivp{} and Gemini-3.1-Pro on functional writing tasks. 
These tasks consist of common daily writing queries, where prompts are typically concise and straightforward. 
Gemini-3.1-Pro was selected as the baseline, as it stands as the top-performing external model for Chinese writing in our evaluations. 
The results indicate that \dsvivp{} outperforms the baseline with an overall win rate of 62.7\% versus 34.1\%; this is primarily because Gemini occasionally allows its inherent stylistic preferences to override the user’s explicit requirements in Chinese writing scenarios.

Table~\ref{tab:creative_writing} presents the creative writing comparison, which is evaluated along two axes: instruction following and writing quality. 
Compared with Gemini-3.1-Pro, \dsvivp{} achieves a 60.0\% win rate in instruction following and 77.5\% in writing quality, demonstrating a marginal improvement in instruction following and a substantial gain in writing quality. 
Although \dsvivp{} yields superior results in aggregate user case analysis, an evaluation restricted to the most challenging prompts --- specifically those involving high-complexity constraints or multi-turn scenarios --- reveals that Claude Opus 4.5 retains a performance advantage over \dsvivp{}. 
As shown in Table~\ref{tab:complex_multiturn}, Claude Opus 4.5 achieves a 52.0\% win rate versus 45.9\%.

\subsubsection{Search}

Search-augmented question answering is a core capability of the DeepSeek chatbot. 
On the DeepSeek web and app, the "non-think" mode employs Retrieval-Augmented Search (RAG), whereas the "thinking" mode utilizes agentic search.  

\paragraph{Retrieval Augmented Search.}
We conducted a pairwise evaluation comparing \dsvivp{} and \dsviiiII{} across both objective and subjective Q\&A categories. 
As presented in Table~\ref{tab:search_qa}, \dsvivp{} outperforms \dsviiiII{} by a substantial margin, demonstrating a consistent advantage across both categories. 
The most pronounced gains are observed in single-value search and planning \& strategy tasks, suggesting that \dsvivp{} excels at locating precise factual answers and synthesizing structured plans from retrieved context. 
However, \dsviiiII{} remains relatively competitive on comparison and recommendation tasks, indicating potential room for improvement for \dsvivp{} in scenarios requiring balanced, multi-perspective reasoning over search results.

\paragraph{Agentic Search.}
Unlike standard RAG, agentic search empowers the model to iteratively invoke search and fetch tools per query, significantly enhancing overall search performance. 
For the thinking mode in DeepSeek-Chat, we optimized the agentic search function to maximize response accuracy within a predefined "thinking budget". 
As shown in Table~\ref{tab:agentic_search}, agentic search consistently outperforms RAG, particularly on complex tasks. 
Furthermore, its cost remains highly efficient, with agentic search being only marginally more expensive than standard RAG (see Table~\ref{tab:cost_comparison}).

\subsubsection{White-Collar Task}

To rigorously evaluate the model's utility in sophisticated enterprise productivity scenarios, we constructed a comprehensive suite of 30 advanced Chinese professional tasks. 
These workflows deliberately encompass high-level cognitive demands, including in-depth information analysis, comprehensive document generation, and nuanced document editing, spanning a diverse spectrum of 13 critical industries (e.g., finance, education, law, and technology). 
The evaluation was conducted within an in-house agent harness equipped with basic tools, including Bash and web search. 

Given the open-ended nature of these tasks, automated metrics usually fall short in capturing the nuances of a high-quality response.
Therefore, we conducted human evaluations to compare the performance of \dsvivpm{} against Opus-4.6-Max. 
Annotators blindly assessed the model outputs across four dimensions:

\begin{itemize}
    \item \textbf{Task Completion:} Whether the core problem was successfully resolved.
    \item \textbf{Instruction Following:} Adherence to specific constraints and directives.
    \item \textbf{Content Quality:} Factual accuracy, logical coherence, and professional tone.
    \item \textbf{Formatting Aesthetics:} Layout readability and visual presentation.
\end{itemize}

\begin{figure}[htbp]
    \centering
    \begin{minipage}{0.48\textwidth}
        \centering
        \includegraphics[width=\linewidth]{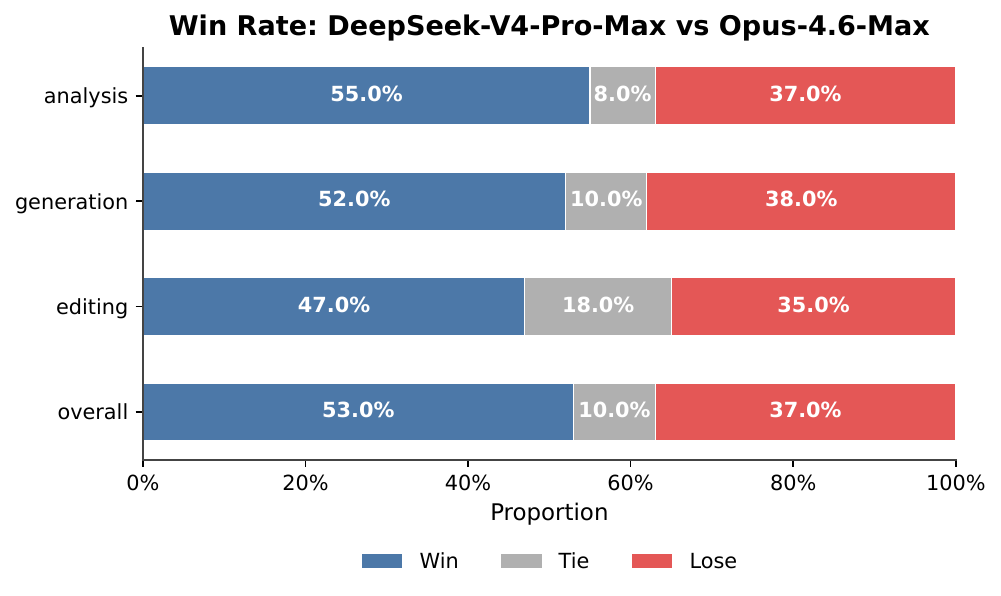}
        \caption{Win-rate comparison across analysis, generation, editing tasks, and the overall performance.}
        \label{fig:winrate}
    \end{minipage}\hfill
    \begin{minipage}{0.48\textwidth}
        \centering
        \includegraphics[width=\linewidth]{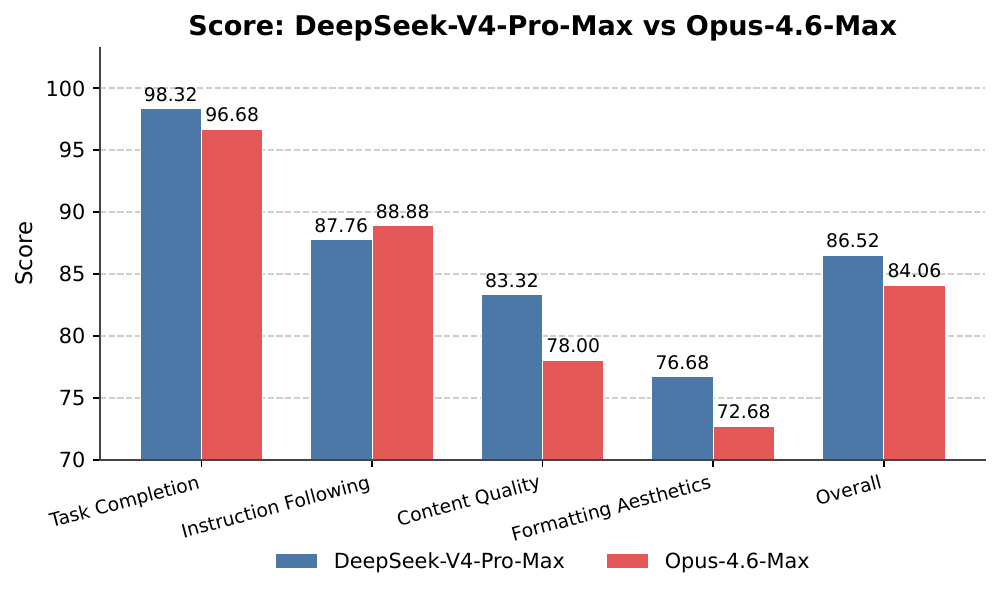}
        \caption{Detailed dimension scores including Task Completion, Content Quality, Formatting Aesthetics, and Instruction Following.}
        \label{fig:scores}
    \end{minipage}
\end{figure}

\begin{figure}[htbp]
    \centering
    \includegraphics[width=\linewidth]{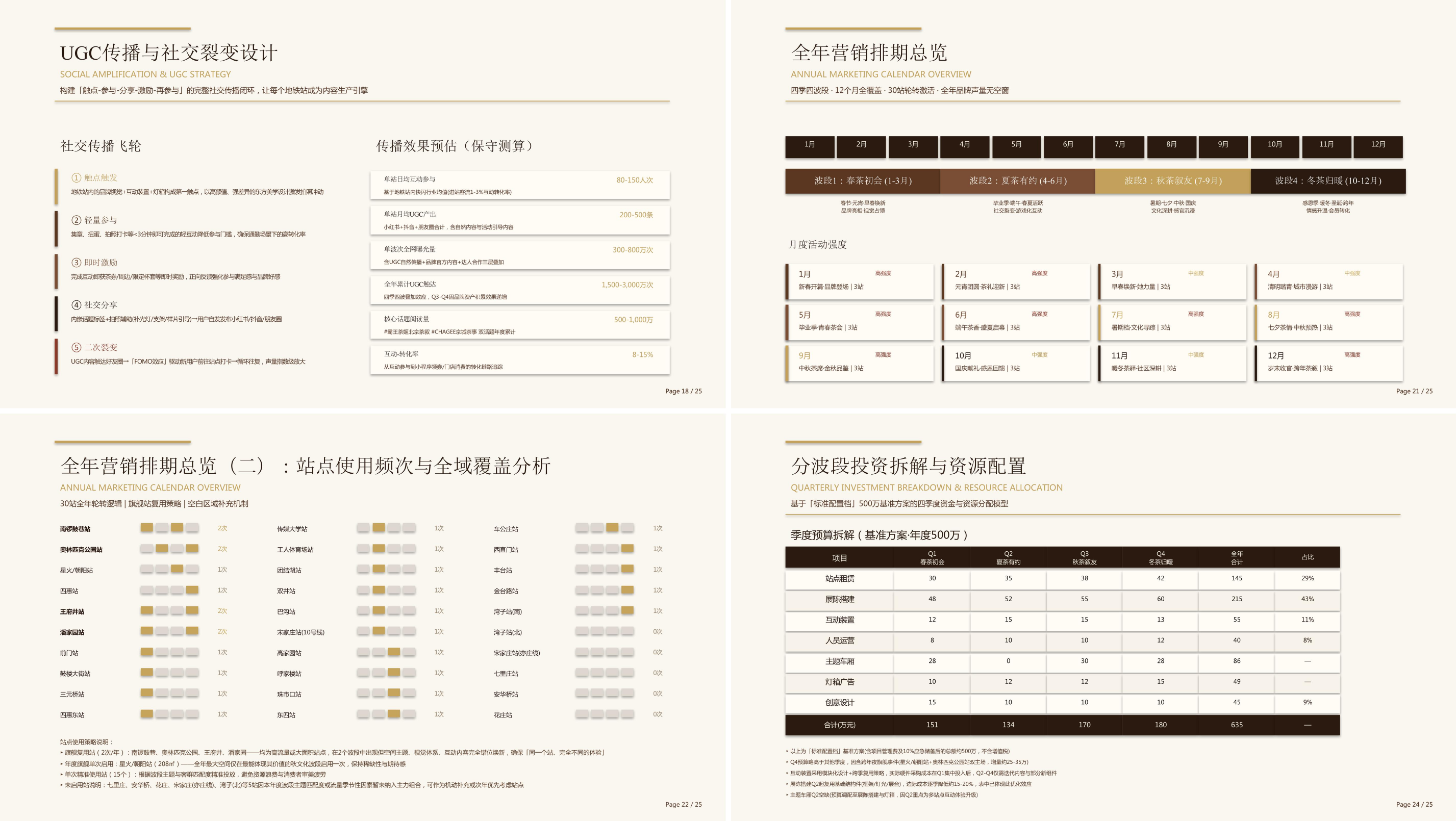} 
    \caption{Example output of a task which requires drafting a joint marketing proposal for a popular bubble tea brand and the Beijing Subway.}
    \label{fig:case-bawangchaji}
\end{figure}

As illustrated in Figure~\ref{fig:winrate}, \dsvivpm{} outperforms Opus-4.6-Max on diverse Chinese white-collar tasks, achieving an impressive non-loss rate of 63\%, and demonstrating consistent advantages across analysis, generation, and editing tasks. 
The detailed dimension scores shown in Figure~\ref{fig:scores} highlight the model's primary strengths in Task Completion and Content Quality. 
Specifically, \dsvivpm{} proactively anticipates implicit user intents by frequently providing supplementary insights and self-verification steps. 
It also excels in long-form generation, delivering in-depth, coherent narratives rather than relying on the overly simplistic bullet points frequently produced by Opus-4.6-Max. 
Additionally, the model strictly conforms to formal professional conventions, such as standardized Chinese hierarchical numbering.
However, in terms of Instruction Following, it occasionally overlooks specific formatting constraints and slightly trails Opus. 
Furthermore, the model is less proficient at condensing extensive text inputs into succinct summaries. 
Finally, its Formatting Aesthetics still have substantial room for improvement regarding the overall visual design of presentation slides.
Figure~\ref{fig:case-bawangchaji}, \ref{fig:case-nasdaq}, and~\ref{fig:case-nobel} present several test cases; due to the extensive length of certain outputs, only partial pages are displayed.

\subsubsection{Code Agent}

To benchmark our coding agent capability, we curate tasks from real internal R\&D workloads
We collect ${\sim}$200 challenging tasks from 50+ internal engineers,
spanning feature development, bug fixing, refactoring, and diagnostics
across diverse technology stacks including PyTorch, CUDA, Rust, and C++.
Each task is accompanied by its original repository, the corresponding execution environment, and human-annotated scoring rubrics; after rigorous quality filtering, 30 tasks are retained as the evaluation set.
As shown in Table~\ref{tab:code-agent}, \dsvivp{} significantly outperforms Claude Sonnet 4.5 and approaches the level of Claude Opus 4.5.

\begin{table}[ht]
\centering
\footnotesize
\caption{Comparison on R\&D Coding Benchmark (external models included strictly for evaluation purposes).}
\label{tab:code-agent}
\begin{tabular}{@{}ccccccc@{}}
\toprule
Model & Haiku 4.5 & Sonnet 4.5 & \dsvivpm{} & Opus 4.5 & \shortstack{Opus 4.5\\Thinking} & \shortstack{Opus 4.6\\Thinking} \\
\midrule
Pass Rate (\%) & 13 & 47 & 67 & 70 & 73 & \textbf{80} \\
\bottomrule
\end{tabular}
\end{table}

In a survey asking DeepSeek developers and researchers ($N=85$) --- all with experience of using \dsvivp{} for agentic coding in their daily work --- whether \dsvivp{} is ready to serve as their default and primary coding model compared to other frontier models, 52\% said yes, 39\% leaned toward yes, and fewer than 9\% said no.
Respondents find \dsvivp{} to deliver satisfactory results across most tasks, but note trivial mistakes, misinterpretation of vague prompts, and occasional over-thinking.

\section{Conclusion, Limitations, and Future Directions}
\label{sec:conclusion}

In this work, we present a preview version of \dsviv{} series, aiming at next-generation large language models that break the efficiency barrier of ultra-long-context processing. 
By combining a hybrid attention architecture that integrates \csa{} and \hca{}, \dsviv{} series achieve a dramatic leap in long-sequence efficiency. 
The architectural innovations, together with extensive infrastructure optimization, enable efficient native support for million-token contexts and establish a necessary foundation for future test-time scaling, long-horizon tasks, and emerging paradigms such as online learning.
Evaluation results demonstrate that \dsvivpm{}, the maximum reasoning effort mode of \dsvivp{}, redefines the state-of-the-art for open models. 
It substantially outperforms prior open-source models on knowledge benchmarks, achieves superior reasoning performance close to the frontier proprietary models, and delivers competitive agent capabilities. 
Meanwhile, \dsvivfm{} attains comparable reasoning performance to leading closed models while maintaining a highly cost-efficient architecture. 
We believe \dsviv{} series usher in a new era of million-length contexts for open models and pave the way toward better efficiency, scale, and intelligence.

In pursuit of extreme long-context efficiency, \dsviv{} series adopted a bold architectural design. 
To minimize risk, we retained many preliminarily validated components and tricks, which, while effective, made the architecture relatively complex. 
In future iterations, we will carry out more comprehensive and principled investigations to distill the architecture down to its most essential designs, making it more elegant without sacrificing performance. 
Meanwhile, although Anticipatory Routing and SwiGLU Clamping have been proven effective in mitigating training instabilities, their underlying principles remain insufficiently understood. 
We will actively study foundational problems on training stability and strengthen internal metric monitoring, aiming for a more principled and predictive approach to stable large-scale training. 
In addition, beyond the MoE and sparse attention architecture, we will also proactively explore model sparsity along new dimensions --- such as more sparse embedding modules~\citep{engram} --- to further improve computational and memory efficiency without compromising capability. 
We will also continuously investigate low-latency architectures and system techniques to make long-context deployment and interaction more responsive.
Furthermore, we recognize the importance and practical value of long-horizon, multi-round agentic tasks, and will continue to iterate and explore in this direction.
We are also working on incorporating multimodal capabilities to our models. 
Finally, we are committed to developing better data curation and synthesis strategies to consistently enhance model intelligence, robustness, and practical usability across an increasingly broad range of scenarios and tasks.

\bibliography{main}

@article{achim2025aristotle,
  title={Aristotle: Imo-level automated theorem proving},
  author={Achim, Tudor and Best, Alex and Bietti, Alberto and Der, Kevin and F{\'e}d{\'e}rico, Math{\"\i}s and Gukov, Sergei and Halpern-Leistner, Daniel and Henningsgard, Kirsten and Kudryashov, Yury and Meiburg, Alexander and others},
  journal={arXiv preprint arXiv:2510.01346},
  year={2025}
}

@misc{chen2025seedprover15masteringundergraduatelevel,
      title={Seed-Prover 1.5: Mastering Undergraduate-Level Theorem Proving via Learning from Experience}, 
      author={Jiangjie Chen and Wenxiang Chen and Jiacheng Du and Jinyi Hu and Zhicheng Jiang and Allan Jie and Xiaoran Jin and Xing Jin and Chenggang Li and Wenlei Shi and Zhihong Wang and Mingxuan Wang and Chenrui Wei and Shufa Wei and Huajian Xin and Fan Yang and Weihao Gao and Zheng Yuan and Tianyang Zhan and Zeyu Zheng and Tianxi Zhou and Thomas Hanwen Zhu},
      year={2025},
      eprint={2512.17260},
      archivePrefix={arXiv},
      primaryClass={cs.CL},
      url={https://arxiv.org/abs/2512.17260}, 
}

@article{patwardhan2025gdpval,
  title={Gdpval: Evaluating ai model performance on real-world economically valuable tasks},
  author={Patwardhan, Tejal and Dias, Rachel and Proehl, Elizabeth and Kim, Grace and Wang, Michele and Watkins, Olivia and Fishman, Sim{\'o}n Posada and Aljubeh, Marwan and Thacker, Phoebe and Fauconnet, Laurance and others},
  journal={arXiv preprint arXiv:2510.04374},
  year={2025}
}

@article{aimuyo2025FlashMoE,
  title={FlashMoE: Fast Distributed MoE in a Single Kernel},
  author={Aimuyo, Osayamen Jonathan and Oh, Byungsoo and Singh, Rachee},
  journal={Advances in Neural Information Processing Systems},
  year={2025},
  url={https://neurips.cc/virtual/2025/poster/119124}
}

@inproceedings{moura2021lean,
  title={The lean 4 theorem prover and programming language},
  author={Moura, Leonardo de and Ullrich, Sebastian},
  booktitle={International Conference on Automated Deduction},
  pages={625--635},
  year={2021},
  organization={Springer}
}

@software{Asher_LeanExplore_2025,
  author = {Asher, Justin},
  title = {{LeanExplore: A search engine for Lean 4 declarations}},
  year = {2025},
  url = {https://arxiv.org/abs/2506.11085}
}

@misc{tsoukalas2024putnambenchevaluatingneuraltheoremprovers,
      title={PutnamBench: Evaluating Neural Theorem-Provers on the Putnam Mathematical Competition}, 
      author={George Tsoukalas and Jasper Lee and John Jennings and Jimmy Xin and Michelle Ding and Michael Jennings and Amitayush Thakur and Swarat Chaudhuri},
      year={2024},
      eprint={2407.11214},
      archivePrefix={arXiv},
      primaryClass={cs.AI},
      url={https://arxiv.org/abs/2407.11214}, 
}

@article{jain2024livecodebench,
  title={LiveCodeBench: Holistic and Contamination Free Evaluation of Large Language Models for Code},
  author={Jain, Naman and Han, King and Gu, Alex and Li, Wen-Ding and Yan, Fanjia and Zhang, Tianjun and Wang, Sida and Solar-Lezama, Armando and Sen, Koushik and Stoica, Ion},
  journal={arXiv preprint arXiv:2403.07974},
  year={2024}
}

@article{su2024roformer,
  title={Roformer: Enhanced transformer with rotary position embedding},
  author={Su, Jianlin and Ahmed, Murtadha and Lu, Yu and Pan, Shengfeng and Bo, Wen and Liu, Yunfeng},
  journal={Neurocomputing},
  volume={568},
  pages={127063},
  year={2024},
  publisher={Elsevier}
}

@article{ainslie2023gqa,
  title={GQA: Training Generalized Multi-Query Transformer Models from Multi-Head Checkpoints},
  author={Ainslie, Joshua and Lee-Thorp, James and de Jong, Michiel and Zemlyanskiy, Yury and Lebr{\'o}n, Federico and Sanghai, Sumit},
  journal={arXiv preprint arXiv:2305.13245},
  year={2023}
}

@article{mqa,
  author       = {Noam Shazeer},
  title        = {Fast Transformer Decoding: One Write-Head is All You Need},
  journal      = {CoRR},
  volume       = {abs/1911.02150},
  year         = {2019},
  url          = {http://arxiv.org/abs/1911.02150},
}

@inproceedings{joshi-etal-2017-triviaqa,
    title = "{T}rivia{QA}: A Large Scale Distantly Supervised Challenge Dataset for Reading Comprehension",
    author = "Joshi, Mandar  and
      Choi, Eunsol  and
      Weld, Daniel  and
      Zettlemoyer, Luke",
    editor = "Barzilay, Regina  and
      Kan, Min-Yen",
    booktitle = "Proceedings of the 55th Annual Meeting of the Association for Computational Linguistics (Volume 1: Long Papers)",
    month = jul,
    year = "2017",
    address = "Vancouver, Canada",
    publisher = "Association for Computational Linguistics",
    url = "https://aclanthology.org/P17-1147",
    doi = "10.18653/v1/P17-1147",
    pages = "1601--1611",
}

@misc{OCP_MXFormat,
      title={Microscaling Data Formats for Deep Learning}, 
      author={Bita Darvish Rouhani and Ritchie Zhao and Ankit More and Mathew Hall and Alireza Khodamoradi and Summer Deng and Dhruv Choudhary and Marius Cornea and Eric Dellinger and Kristof Denolf and Stosic Dusan and Venmugil Elango and Maximilian Golub and Alexander Heinecke and Phil James-Roxby and Dharmesh Jani and Gaurav Kolhe and Martin Langhammer and Ada Li and Levi Melnick and Maral Mesmakhosroshahi and Andres Rodriguez and Michael Schulte and Rasoul Shafipour and Lei Shao and Michael Siu and Pradeep Dubey and Paulius Micikevicius and Maxim Naumov and Colin Verrilli and Ralph Wittig and Doug Burger and Eric Chung},
      year={2023},
      eprint={2310.10537},
      archivePrefix={arXiv},
      primaryClass={cs.LG},
}

@InProceedings{QAT,
author = {Jacob, Benoit and Kligys, Skirmantas and Chen, Bo and Zhu, Menglong and Tang, Matthew and Howard, Andrew and Adam, Hartwig and Kalenichenko, Dmitry},
title = {Quantization and Training of Neural Networks for Efficient Integer-Arithmetic-Only Inference},
booktitle = {Proceedings of the IEEE Conference on Computer Vision and Pattern Recognition (CVPR)},
month = {June},
year = {2018}
}

@article{transformer,
  title={Attention is all you need},
  author={Vaswani, Ashish and Shazeer, Noam and Parmar, Niki and Uszkoreit, Jakob and Jones, Llion and Gomez, Aidan N and Kaiser, {\L}ukasz and Polosukhin, Illia},
  journal={Advances in neural information processing systems},
  volume={30},
  year={2017}
}

@inproceedings{zero,
  title={Zero: Memory optimizations toward training trillion parameter models},
  author={Rajbhandari, Samyam and Rasley, Jeff and Ruwase, Olatunji and He, Yuxiong},
  booktitle={SC20: International Conference for High Performance Computing, Networking, Storage and Analysis},
  pages={1--16},
  year={2020},
  organization={IEEE}
}

@misc{sakaguchi2019winogrande,
      title={WinoGrande: An Adversarial Winograd Schema Challenge at Scale}, 
      author={Keisuke Sakaguchi and Ronan Le Bras and Chandra Bhagavatula and Yejin Choi},
      year={2019},
      eprint={1907.10641},
      archivePrefix={arXiv},
      primaryClass={cs.CL}
}

@misc{wei2023cmath,
      title={CMATH: Can Your Language Model Pass Chinese Elementary School Math Test?}, 
      author={Tianwen Wei and Jian Luan and Wei Liu and Shuang Dong and Bin Wang},
      year={2023},
      eprint={2306.16636},
      archivePrefix={arXiv},
      primaryClass={cs.CL}
}

@article{mmlu,
  title={Measuring massive multitask language understanding},
  author={Hendrycks, Dan and Burns, Collin and Basart, Steven and Zou, Andy and Mazeika, Mantas and Song, Dawn and Steinhardt, Jacob},
  journal={arXiv preprint arXiv:2009.03300},
  year={2020}
}

@article{bbh,
  title={Challenging big-bench tasks and whether chain-of-thought can solve them},
  author={Suzgun, Mirac and Scales, Nathan and Sch{\"a}rli, Nathanael and Gehrmann, Sebastian and Tay, Yi and Chung, Hyung Won and Chowdhery, Aakanksha and Le, Quoc V and Chi, Ed H and Zhou, Denny and others},
  journal={arXiv preprint arXiv:2210.09261},
  year={2022}
}

@inproceedings{clue,
  author       = {Liang Xu and
                  Hai Hu and
                  Xuanwei Zhang and
                  Lu Li and
                  Chenjie Cao and
                  Yudong Li and
                  Yechen Xu and
                  Kai Sun and
                  Dian Yu and
                  Cong Yu and
                  Yin Tian and
                  Qianqian Dong and
                  Weitang Liu and
                  Bo Shi and
                  Yiming Cui and
                  Junyi Li and
                  Jun Zeng and
                  Rongzhao Wang and
                  Weijian Xie and
                  Yanting Li and
                  Yina Patterson and
                  Zuoyu Tian and
                  Yiwen Zhang and
                  He Zhou and
                  Shaoweihua Liu and
                  Zhe Zhao and
                  Qipeng Zhao and
                  Cong Yue and
                  Xinrui Zhang and
                  Zhengliang Yang and
                  Kyle Richardson and
                  Zhenzhong Lan},
  editor       = {Donia Scott and
                  N{\'{u}}ria Bel and
                  Chengqing Zong},
  title        = {{CLUE:} {A} Chinese Language Understanding Evaluation Benchmark},
  booktitle    = {Proceedings of the 28th International Conference on Computational
                  Linguistics, {COLING} 2020, Barcelona, Spain (Online), December 8-13,
                  2020},
  pages        = {4762--4772},
  publisher    = {International Committee on Computational Linguistics},
  year         = {2020},
  url          = {https://doi.org/10.18653/v1/2020.coling-main.419},
  doi          = {10.18653/V1/2020.COLING-MAIN.419},
  bibsource    = {dblp computer science bibliography, https://dblp.org}
}

@article{cmmlu,
  title={{CMMLU}: Measuring massive multitask language understanding in {Chinese}},
  author={Li, Haonan and Zhang, Yixuan and Koto, Fajri and Yang, Yifei and Zhao, Hai and Gong, Yeyun and Duan, Nan and Baldwin, Timothy},
  journal={arXiv preprint arXiv:2306.09212},
  year={2023}
}

@inproceedings{drop,
  author       = {Dheeru Dua and
                  Yizhong Wang and
                  Pradeep Dasigi and
                  Gabriel Stanovsky and
                  Sameer Singh and
                  Matt Gardner},
  editor       = {Jill Burstein and
                  Christy Doran and
                  Thamar Solorio},
  title        = {{DROP:} {A} Reading Comprehension Benchmark Requiring Discrete Reasoning
                  Over Paragraphs},
  booktitle    = {Proceedings of the 2019 Conference of the North American Chapter of
                  the Association for Computational Linguistics: Human Language Technologies,
                  {NAACL-HLT} 2019, Minneapolis, MN, USA, June 2-7, 2019, Volume 1 (Long
                  and Short Papers)},
  pages        = {2368--2378},
  publisher    = {Association for Computational Linguistics},
  year         = {2019},
  url          = {https://doi.org/10.18653/v1/n19-1246},
  doi          = {10.18653/V1/N19-1246},
  bibsource    = {dblp computer science bibliography, https://dblp.org}
}

@article{ceval,
  title={{C-Eval}: A multi-level multi-discipline chinese evaluation suite for foundation models},
  author={Huang, Yuzhen and Bai, Yuzhuo and Zhu, Zhihao and Zhang, Junlei and Zhang, Jinghan and Su, Tangjun and Liu, Junteng and Lv, Chuancheng and Zhang, Yikai and Lei, Jiayi and others},
  journal={arXiv preprint arXiv:2305.08322},
  year={2023}
}

@article{gsm8k,
  title={Training verifiers to solve math word problems},
  author={Cobbe, Karl and Kosaraju, Vineet and Bavarian, Mohammad and Chen, Mark and Jun, Heewoo and Kaiser, Lukasz and Plappert, Matthias and Tworek, Jerry and Hilton, Jacob and Nakano, Reiichiro and others},
  journal={arXiv preprint arXiv:2110.14168},
  year={2021}
}

@article{agieval,
  author       = {Wanjun Zhong and
                  Ruixiang Cui and
                  Yiduo Guo and
                  Yaobo Liang and
                  Shuai Lu and
                  Yanlin Wang and
                  Amin Saied and
                  Weizhu Chen and
                  Nan Duan},
  title        = {{AGIEval}: {A} Human-Centric Benchmark for Evaluating Foundation Models},
  journal      = {CoRR},
  volume       = {abs/2304.06364},
  year         = {2023},
  url          = {https://doi.org/10.48550/arXiv.2304.06364},
  doi          = {10.48550/arXiv.2304.06364},
  eprinttype    = {arXiv},
  eprint       = {2304.06364},
  bibsource    = {dblp computer science bibliography, https://dblp.org}
}

@article{codex,
  author       = {Mark Chen and
                  Jerry Tworek and
                  Heewoo Jun and
                  Qiming Yuan and
                  Henrique Pond{\'{e}} de Oliveira Pinto and
                  Jared Kaplan and
                  Harrison Edwards and
                  Yuri Burda and
                  Nicholas Joseph and
                  Greg Brockman and
                  Alex Ray and
                  Raul Puri and
                  Gretchen Krueger and
                  Michael Petrov and
                  Heidy Khlaaf and
                  Girish Sastry and
                  Pamela Mishkin and
                  Brooke Chan and
                  Scott Gray and
                  Nick Ryder and
                  Mikhail Pavlov and
                  Alethea Power and
                  Lukasz Kaiser and
                  Mohammad Bavarian and
                  Clemens Winter and
                  Philippe Tillet and
                  Felipe Petroski Such and
                  Dave Cummings and
                  Matthias Plappert and
                  Fotios Chantzis and
                  Elizabeth Barnes and
                  Ariel Herbert{-}Voss and
                  William Hebgen Guss and
                  Alex Nichol and
                  Alex Paino and
                  Nikolas Tezak and
                  Jie Tang and
                  Igor Babuschkin and
                  Suchir Balaji and
                  Shantanu Jain and
                  William Saunders and
                  Christopher Hesse and
                  Andrew N. Carr and
                  Jan Leike and
                  Joshua Achiam and
                  Vedant Misra and
                  Evan Morikawa and
                  Alec Radford and
                  Matthew Knight and
                  Miles Brundage and
                  Mira Murati and
                  Katie Mayer and
                  Peter Welinder and
                  Bob McGrew and
                  Dario Amodei and
                  Sam McCandlish and
                  Ilya Sutskever and
                  Wojciech Zaremba},
  title        = {Evaluating Large Language Models Trained on Code},
  journal      = {CoRR},
  volume       = {abs/2107.03374},
  year         = {2021},
  url          = {https://arxiv.org/abs/2107.03374},
  eprinttype    = {arXiv},
  eprint       = {2107.03374},
}

@inproceedings{hellaswag,
  author       = {Rowan Zellers and
                  Ari Holtzman and
                  Yonatan Bisk and
                  Ali Farhadi and
                  Yejin Choi},
  editor       = {Anna Korhonen and
                  David R. Traum and
                  Llu{\'{\i}}s M{\`{a}}rquez},
  title        = {{HellaSwag}: Can a Machine Really Finish Your Sentence?},
  booktitle    = {Proceedings of the 57th Conference of the Association for Computational
                  Linguistics, {ACL} 2019, Florence, Italy, July 28- August 2, 2019,
                  Volume 1: Long Papers},
  pages        = {4791--4800},
  publisher    = {Association for Computational Linguistics},
  year         = {2019},
  url          = {https://doi.org/10.18653/v1/p19-1472},
  doi          = {10.18653/v1/p19-1472},
  bibsource    = {dblp computer science bibliography, https://dblp.org}
}

@article{hendrycks2021measuring,
  title={Measuring mathematical problem solving with the math dataset},
  author={Hendrycks, Dan and Burns, Collin and Kadavath, Saurav and Arora, Akul and Basart, Steven and Tang, Eric and Song, Dawn and Steinhardt, Jacob},
  journal={arXiv preprint arXiv:2103.03874},
  year={2021}
}

@misc{shao2025deepseekmathv2selfverifiablemathematicalreasoning,
      title={DeepSeekMath-V2: Towards Self-Verifiable Mathematical Reasoning}, 
      author={Zhihong Shao and Yuxiang Luo and Chengda Lu and Z. Z. Ren and Jiewen Hu and Tian Ye and Zhibin Gou and Shirong Ma and Xiaokang Zhang},
      year={2025},
      eprint={2511.22570},
      archivePrefix={arXiv},
      primaryClass={cs.AI},
      url={https://arxiv.org/abs/2511.22570}, 
}

@article{adamW,
  title={Decoupled weight decay regularization},
  author={Loshchilov, Ilya and Hutter, Frank},
  journal={arXiv preprint arXiv:1711.05101},
  year={2017}
}

@misc{gdpaa,
  title = {GDPval-AA Leaderboard},
  author = {AA},
  url = {https://artificialanalysis.ai/methodology/intelligence-benchmarking#gdpval-aa},
  year={2025}
}

@misc{mrcr,
  title = {OpenAI MRCR: Long context multiple needle in a haystack benchmark},
  author = {OpenAI},
  url = {https://huggingface.co/datasets/openai/mrcr},
  year={2024}
}

@article{merrill2026terminal,
  title={Terminal-bench: Benchmarking agents on hard, realistic tasks in command line interfaces},
  author={Merrill, Mike A and Shaw, Alexander G and Carlini, Nicholas and Li, Boxuan and Raj, Harsh and Bercovich, Ivan and Shi, Lin and Shin, Jeong Yeon and Walshe, Thomas and Buchanan, E Kelly and others},
  journal={arXiv preprint arXiv:2601.11868},
  year={2026}
}

@article{lu2025onpolicydistillation,
  author = {Kevin Lu and Thinking Machines Lab},
  title = {On-Policy Distillation},
  journal = {Thinking Machines Lab: Connectionism},
  year = {2025},
  note = {https://thinkingmachines.ai/blog/on-policy-distillation},
  doi = {10.64434/tml.20251026},
}

@article{bandi2026mcp,
  title={MCP-Atlas: A Large-Scale Benchmark for Tool-Use Competency with Real MCP Servers},
  author={Bandi, Chaithanya and Hertzberg, Ben and Boo, Geobio and Polakam, Tejas and Da, Jeff and Hassaan, Sami and Sharma, Manasi and Park, Andrew and Hernandez, Ernesto and Rambado, Dan and others},
  journal={arXiv preprint arXiv:2602.00933},
  year={2026}
}

@article{shazeer2020glu,
  title={Glu variants improve transformer},
  author={Shazeer, Noam},
  journal={arXiv preprint arXiv:2002.05202},
  year={2020}
}

@article{deepseekmoe,
  author       = {Damai Dai and
                  Chengqi Deng and
                  Chenggang Zhao and
                  R. X. Xu and
                  Huazuo Gao and
                  Deli Chen and
                  Jiashi Li and
                  Wangding Zeng and
                  Xingkai Yu and
                  Y. Wu and
                  Zhenda Xie and
                  Y. K. Li and
                  Panpan Huang and
                  Fuli Luo and
                  Chong Ruan and
                  Zhifang Sui and
                  Wenfeng Liang},
  title        = {DeepSeekMoE: Towards Ultimate Expert Specialization in Mixture-of-Experts
                  Language Models},
  journal      = {CoRR},
  volume       = {abs/2401.06066},
  year         = {2024},
  url          = {https://doi.org/10.48550/arXiv.2401.06066},
}

@misc{swe_verified,
  title={Introducing {SWE}-bench Verified
We’re releasing a human-validated subset of SWE-bench that more},
  author={OpenAI},
  year={2024},
  url = {https://openai.com/index/introducing-swe-bench-verified/}
}

@article{hle,
  title={Humanity's last exam},
  author={Phan, Long and Gatti, Alice and Han, Ziwen and Li, Nathaniel and Hu, Josephina and Zhang, Hugh and Zhang, Chen Bo Calvin and Shaaban, Mohamed and Ling, John and Shi, Sean and others},
  journal={arXiv preprint arXiv:2501.14249},
  year={2025}
}

@article{gpqa,
  title={{GPQA}: A graduate-level google-proof q\&a benchmark},
  author={Rein, David and Hou, Betty Li and Stickland, Asa Cooper and Petty, Jackson and Pang, Richard Yuanzhe and Dirani, Julien and Michael, Julian and Bowman, Samuel R},
  journal={arXiv preprint arXiv:2311.12022},
  year={2023}
}

@misc{simpleqa,
  title={Introducing {SimpleQA}},
  author={OpenAI},
  year={2024},
  url = {https://openai.com/index/introducing-simpleqa/}
}

@article{csimpleqa,
  title={Chinese simpleqa: A chinese factuality evaluation for large language models},
  author={He, Yancheng and Li, Shilong and Liu, Jiaheng and Tan, Yingshui and Wang, Weixun and Huang, Hui and Bu, Xingyuan and Guo, Hangyu and Hu, Chengwei and Zheng, Boren and others},
  journal={arXiv preprint arXiv:2411.07140},
  year={2024}
}

@article{dsvii,
  author       = {DeepSeek-AI},
  title        = {DeepSeek-V2: {A} Strong, Economical, and Efficient Mixture-of-Experts
                  Language Model},
  journal      = {CoRR},
  volume       = {abs/2405.04434},
  year         = {2024},
  url          = {https://doi.org/10.48550/arXiv.2405.04434},
}

@article{noaux_tc,
  author       = {Lean Wang and
                  Huazuo Gao and
                  Chenggang Zhao and
                  Xu Sun and
                  Damai Dai},
  title        = {Auxiliary-Loss-Free Load Balancing Strategy for Mixture-of-Experts},
  journal      = {CoRR},
  volume       = {abs/2408.15664},
  year         = {2024},
  url          = {https://doi.org/10.48550/arXiv.2408.15664},
}

@article{dscodervii,
  author       = {DeepSeek-AI},
  title        = {DeepSeek-Coder-V2: Breaking the Barrier of Closed-Source Models in
                  Code Intelligence},
  journal      = {CoRR},
  volume       = {abs/2406.11931},
  year         = {2024},
  url          = {https://doi.org/10.48550/arXiv.2406.11931},
}

@article{mmlu_redux,
  author       = {Aryo Pradipta Gema and
                  Joshua Ong Jun Leang and
                  Giwon Hong and
                  Alessio Devoto and
                  Alberto Carlo Maria Mancino and
                  Rohit Saxena and
                  Xuanli He and
                  Yu Zhao and
                  Xiaotang Du and
                  Mohammad Reza Ghasemi Madani and
                  Claire Barale and
                  Robert McHardy and
                  Joshua Harris and
                  Jean Kaddour and
                  Emile van Krieken and
                  Pasquale Minervini},
  title        = {Are We Done with MMLU?},
  journal      = {CoRR},
  volume       = {abs/2406.04127},
  year         = {2024},
  url          = {https://doi.org/10.48550/arXiv.2406.04127},
}

@article{mmlu_pro,
  author       = {Yubo Wang and
                  Xueguang Ma and
                  Ge Zhang and
                  Yuansheng Ni and
                  Abhranil Chandra and
                  Shiguang Guo and
                  Weiming Ren and
                  Aaran Arulraj and
                  Xuan He and
                  Ziyan Jiang and
                  Tianle Li and
                  Max Ku and
                  Kai Wang and
                  Alex Zhuang and
                  Rongqi Fan and
                  Xiang Yue and
                  Wenhu Chen},
  title        = {MMLU-Pro: {A} More Robust and Challenging Multi-Task Language Understanding
                  Benchmark},
  journal      = {CoRR},
  volume       = {abs/2406.01574},
  year         = {2024},
  url          = {https://doi.org/10.48550/arXiv.2406.01574},
}

@inproceedings{mgsm,
  author       = {Freda Shi and
                  Mirac Suzgun and
                  Markus Freitag and
                  Xuezhi Wang and
                  Suraj Srivats and
                  Soroush Vosoughi and
                  Hyung Won Chung and
                  Yi Tay and
                  Sebastian Ruder and
                  Denny Zhou and
                  Dipanjan Das and
                  Jason Wei},
  title        = {Language models are multilingual chain-of-thought reasoners},
  booktitle    = {The Eleventh International Conference on Learning Representations,
                  {ICLR} 2023, Kigali, Rwanda, May 1-5, 2023},
  publisher    = {OpenReview.net},
  year         = {2023},
  url          = {https://openreview.net/forum?id=fR3wGCk-IXp},
}

@misc{mmmlu,
  title={Multilingual Massive Multitask Language Understanding (MMMLU)},
  author={OpenAI},
  year={2024},
  url = {https://huggingface.co/datasets/openai/MMMLU}
}

@inproceedings{meta_mtp,
  author       = {Fabian Gloeckle and
                  Badr Youbi Idrissi and
                  Baptiste Rozi{\`{e}}re and
                  David Lopez{-}Paz and
                  Gabriel Synnaeve},
  title        = {Better {\&} Faster Large Language Models via Multi-token Prediction},
  booktitle    = {Forty-first International Conference on Machine Learning, {ICML} 2024,
                  Vienna, Austria, July 21-27, 2024},
  publisher    = {OpenReview.net},
  year         = {2024},
  url          = {https://openreview.net/forum?id=pEWAcejiU2},
}

@inproceedings{ms_mtp,
  author       = {Weizhen Qi and
                  Yu Yan and
                  Yeyun Gong and
                  Dayiheng Liu and
                  Nan Duan and
                  Jiusheng Chen and
                  Ruofei Zhang and
                  Ming Zhou},
  editor       = {Trevor Cohn and
                  Yulan He and
                  Yang Liu},
  title        = {ProphetNet: Predicting Future N-gram for Sequence-to-Sequence Pre-training},
  booktitle    = {Findings of the Association for Computational Linguistics: {EMNLP}
                  2020, Online Event, 16-20 November 2020},
  series       = {Findings of {ACL}},
  volume       = {{EMNLP} 2020},
  pages        = {2401--2410},
  publisher    = {Association for Computational Linguistics},
  year         = {2020},
  url          = {https://doi.org/10.18653/v1/2020.findings-emnlp.217},
}

@inproceedings{eagle,
  author       = {Yuhui Li and
                  Fangyun Wei and
                  Chao Zhang and
                  Hongyang Zhang},
  title        = {{EAGLE:} Speculative Sampling Requires Rethinking Feature Uncertainty},
  booktitle    = {Forty-first International Conference on Machine Learning, {ICML} 2024,
                  Vienna, Austria, July 21-27, 2024},
  publisher    = {OpenReview.net},
  year         = {2024},
  url          = {https://openreview.net/forum?id=1NdN7eXyb4},
}

@article{Ding2024FewerTI,
  title={Fewer Truncations Improve Language Modeling},
  author={Hantian Ding and Zijian Wang and Giovanni Paolini and Varun Kumar and Anoop Deoras and Dan Roth and Stefano Soatto},
  journal={arXiv preprint arXiv:2404.10830},
  year={2024}
}

@inproceedings{hc,
  author       = {Defa Zhu and
                  Hongzhi Huang and
                  Zihao Huang and
                  Yutao Zeng and
                  Yunyao Mao and
                  Banggu Wu and
                  Qiyang Min and
                  Xun Zhou},
  title        = {Hyper-Connections},
  booktitle    = {The Thirteenth International Conference on Learning Representations,
                  {ICLR} 2025, Singapore, April 24-28, 2025},
  publisher    = {OpenReview.net},
  year         = {2025},
  url          = {https://openreview.net/forum?id=9FqARW7dwB},
}

@article{muon,
  title={Muon: An optimizer for hidden layers in neural networks},
  author={Jordan, Keller and Jin, Yuchen and Boza, Vlado and You, Jiacheng and Cesista, Franz and Newhouse, Laker and Bernstein, Jeremy},
  journal={Cited on},
  pages={10},
  year={2024}
}

@article{muon_kimi,
  author       = {Jingyuan Liu and
                  Jianlin Su and
                  Xingcheng Yao and
                  Zhejun Jiang and
                  Guokun Lai and
                  Yulun Du and
                  Yidao Qin and
                  Weixin Xu and
                  Enzhe Lu and
                  Junjie Yan and
                  Yanru Chen and
                  Huabin Zheng and
                  Yibo Liu and
                  Shaowei Liu and
                  Bohong Yin and
                  Weiran He and
                  Han Zhu and
                  Yuzhi Wang and
                  Jianzhou Wang and
                  Mengnan Dong and
                  Zheng Zhang and
                  Yongsheng Kang and
                  Hao Zhang and
                  Xinran Xu and
                  Yutao Zhang and
                  Yuxin Wu and
                  Xinyu Zhou and
                  Zhilin Yang},
  title        = {Muon is Scalable for {LLM} Training},
  journal      = {CoRR},
  volume       = {abs/2502.16982},
  year         = {2025},
  url          = {https://doi.org/10.48550/arXiv.2502.16982},
}

@article{dsv3,
  author       = {DeepSeek{-}AI},
  title        = {DeepSeek-V3 Technical Report},
  journal      = {CoRR},
  volume       = {abs/2412.19437},
  year         = {2024},
  url          = {https://doi.org/10.48550/arXiv.2412.19437},
}

@article{dsr1,
  author       = {DeepSeek{-}AI},
  title        = {DeepSeek-R1 incentivizes reasoning in LLMs through reinforcement learning},
  journal      = {Nat.},
  volume       = {645},
  number       = {8081},
  pages        = {633--638},
  year         = {2025},
  url          = {https://doi.org/10.1038/s41586-025-09422-z},
}

@article{kimi_k2,
  author       = {Yifan Bai and
                  Yiping Bao and
                  Guanduo Chen and
                  Jiahao Chen and
                  Ningxin Chen and
                  Ruijue Chen and
                  Yanru Chen and
                  Yuankun Chen and
                  Yutian Chen and
                  Zhuofu Chen and
                  Jialei Cui and
                  Hao Ding and
                  Mengnan Dong and
                  Angang Du and
                  Chenzhuang Du and
                  Dikang Du and
                  Yulun Du and
                  Yu Fan and
                  Yichen Feng and
                  Kelin Fu and
                  Bofei Gao and
                  Hongcheng Gao and
                  Peizhong Gao and
                  Tong Gao and
                  Xinran Gu and
                  Longyu Guan and
                  Haiqing Guo and
                  Jianhang Guo and
                  Hao Hu and
                  Xiaoru Hao and
                  Tianhong He and
                  Weiran He and
                  Wenyang He and
                  Chao Hong and
                  Yangyang Hu and
                  Zhenxing Hu and
                  Weixiao Huang and
                  Zhiqi Huang and
                  Zihao Huang and
                  Tao Jiang and
                  Zhejun Jiang and
                  Xinyi Jin and
                  Yongsheng Kang and
                  Guokun Lai and
                  Cheng Li and
                  Fang Li and
                  Haoyang Li and
                  Ming Li and
                  Wentao Li and
                  Yanhao Li and
                  Yiwei Li and
                  Zhaowei Li and
                  Zheming Li and
                  Hongzhan Lin and
                  Xiaohan Lin and
                  Zongyu Lin and
                  Chengyin Liu and
                  Chenyu Liu and
                  Hongzhang Liu and
                  Jingyuan Liu and
                  Junqi Liu and
                  Liang Liu and
                  Shaowei Liu and
                  T. Y. Liu and
                  Tianwei Liu and
                  Weizhou Liu and
                  Yangyang Liu and
                  Yibo Liu and
                  Yiping Liu and
                  Yue Liu and
                  Zhengying Liu and
                  Enzhe Lu and
                  Lijun Lu and
                  Shengling Ma and
                  Xinyu Ma and
                  Yingwei Ma and
                  Shaoguang Mao and
                  Jie Mei and
                  Xin Men and
                  Yibo Miao and
                  Siyuan Pan and
                  Yebo Peng and
                  Ruoyu Qin and
                  Bowen Qu and
                  Zeyu Shang and
                  Lidong Shi and
                  Shengyuan Shi and
                  Feifan Song and
                  Jianlin Su and
                  Zhengyuan Su and
                  Xinjie Sun and
                  Flood Sung and
                  Heyi Tang and
                  Jiawen Tao and
                  Qifeng Teng and
                  Chensi Wang and
                  Dinglu Wang and
                  Feng Wang and
                  Haiming Wang},
  title        = {Kimi {K2:} Open Agentic Intelligence},
  journal      = {CoRR},
  volume       = {abs/2507.20534},
  year         = {2025},
  url          = {https://doi.org/10.48550/arXiv.2507.20534},
}

@article{nesterov,
  title={A method of solving a convex programming problem with convergence rate ${O}(1 / k^2)$},
  author={Nesterov, Yurii},
  journal={Soviet Mathematics Doklady},
  volume={27},
  pages={372–376},
  year={1983},
}

@inproceedings{hash_layer,
  author       = {Stephen Roller and
                  Sainbayar Sukhbaatar and
                  Arthur Szlam and
                  Jason Weston},
  editor       = {Marc'Aurelio Ranzato and
                  Alina Beygelzimer and
                  Yann N. Dauphin and
                  Percy Liang and
                  Jennifer Wortman Vaughan},
  title        = {Hash Layers For Large Sparse Models},
  booktitle    = {Advances in Neural Information Processing Systems 34: Annual Conference
                  on Neural Information Processing Systems 2021, NeurIPS 2021, December
                  6-14, 2021, virtual},
  pages        = {17555--17566},
  year         = {2021},
  url          = {https://proceedings.neurips.cc/paper/2021/hash/92bf5e6240737e0326ea59846a83e076-Abstract.html},
}

@misc{dsv32,
      title={DeepSeek-V3.2: Pushing the Frontier of Open Large Language Models}, 
      author={DeepSeek-AI},
      year={2025},
      eprint={2512.02556},
      archivePrefix={arXiv},
      primaryClass={cs.CL},
      url={https://arxiv.org/abs/2512.02556}, 
}

@misc{deepgemm2025,
      title={DeepGEMM: clean and efficient FP8 GEMM kernels with fine-grained scaling}, 
      author={Chenggang Zhao and Liang Zhao and Jiashi Li and Zhean Xu and Chenhao Xu},
      year={2025},
      publisher = {GitHub},
      howpublished = {\url{https://github.com/deepseek-ai/DeepGEMM}},
}

@misc{cublas,
  author = {{NVIDIA Corporation}},
  title = {cuBLAS Documentation},
  year = {2024},
  url = {https://docs.nvidia.com/cuda/cublas/},
  note = {Version 12.4. Accessed: 2024-09-16}
}

@inproceedings{attn_sink,
  author       = {Guangxuan Xiao and
                  Yuandong Tian and
                  Beidi Chen and
                  Song Han and
                  Mike Lewis},
  title        = {Efficient Streaming Language Models with Attention Sinks},
  booktitle    = {The Twelfth International Conference on Learning Representations,
                  {ICLR} 2024, Vienna, Austria, May 7-11, 2024},
  publisher    = {OpenReview.net},
  year         = {2024},
  url          = {https://openreview.net/forum?id=NG7sS51zVF},
}

@article{gpt_oss,
  author       = {OpenAI},
  title        = {gpt-oss-120b {\&} gpt-oss-20b Model Card},
  journal      = {CoRR},
  volume       = {abs/2508.10925},
  year         = {2025},
  url          = {https://doi.org/10.48550/arXiv.2508.10925},
  doi          = {10.48550/ARXIV.2508.10925},
}

@misc{o1,
  title={Learning to reason with LLMs},
  author={OpenAI},
  year={2024},
  url = {https://openai.com/index/learning-to-reason-with-llms}
}

@misc{minimax_m2,
  title={Meet MiniMax-M2},
  author={MiniMax},
  year={2025},
  url = {https://github.com/MiniMax-AI/MiniMax-M2}
}

@article{qwen3,
  author       = {Qwen},
  title        = {Qwen3 Technical Report},
  journal      = {CoRR},
  volume       = {abs/2505.09388},
  year         = {2025},
  url          = {https://doi.org/10.48550/arXiv.2505.09388},
  doi          = {10.48550/ARXIV.2505.09388},
}

@misc{torchfx,
      title={Torch.fx: Practical Program Capture and Transformation for Deep Learning in Python}, 
      author={James K. Reed and Zachary DeVito and Horace He and Ansley Ussery and Jason Ansel},
      year={2022},
      eprint={2112.08429},
      archivePrefix={arXiv},
      primaryClass={cs.LG},
      url={https://arxiv.org/abs/2112.08429}, 
}

@misc{mhc,
      title={mHC: Manifold-Constrained Hyper-Connections}, 
      author={Zhenda Xie and Yixuan Wei and Huanqi Cao and Chenggang Zhao and Chengqi Deng and Jiashi Li and Damai Dai and Huazuo Gao and Jiang Chang and Kuai Yu and Liang Zhao and Shangyan Zhou and Zhean Xu and Zhengyan Zhang and Wangding Zeng and Shengding Hu and Yuqing Wang and Jingyang Yuan and Lean Wang and Wenfeng Liang},
      year={2026},
      eprint={2512.24880},
      archivePrefix={arXiv},
      primaryClass={cs.CL},
      url={https://arxiv.org/abs/2512.24880}, 
}

@inproceedings{jenga,
author = {Zhang, Chen and Du, Kuntai and Liu, Shu and Kwon, Woosuk and Mo, Xiangxi and Wang, Yufeng and Liu, Xiaoxuan and You, Kaichao and Li, Zhuohan and Long, Mingsheng and Zhai, Jidong and Gonzalez, Joseph and Stoica, Ion},
title = {Jenga: Effective Memory Management for Serving LLM with Heterogeneity},
year = {2025},
isbn = {9798400718700},
publisher = {Association for Computing Machinery},
address = {New York, NY, USA},
url = {https://doi.org/10.1145/3731569.3764823},
doi = {10.1145/3731569.3764823},
booktitle = {Proceedings of the ACM SIGOPS 31st Symposium on Operating Systems Principles},
pages = {446–461},
numpages = {16},
location = {Lotte Hotel World, Seoul, Republic of Korea},
series = {SOSP '25}
}

@inproceedings{
dong2025hymba,
title={Hymba: A Hybrid-head Architecture for Small Language Models},
author={Xin Dong and Yonggan Fu and Shizhe Diao and Wonmin Byeon and ZIJIA CHEN and Ameya Sunil Mahabaleshwarkar and Shih-Yang Liu and Matthijs Van keirsbilck and Min-Hung Chen and Yoshi Suhara and Yingyan Celine Lin and Jan Kautz and Pavlo Molchanov},
booktitle={The Thirteenth International Conference on Learning Representations},
year={2025},
url={https://openreview.net/forum?id=A1ztozypga}
}

@misc{flash_decoding,
    title = {Flash-Decoding for long-context inference} ,
    author = {Tri Dao and Daniel Haziza and Francisco Massa and Grigory Sizov},
    url = {https://pytorch.org/blog/flash-decoding/},
    year = {2023},
}

@inproceedings{stream_k,
  title={Stream-k: Work-centric parallel decomposition for dense matrix-matrix multiplication on the gpu},
  author={Osama, Muhammad and Merrill, Duane and Cecka, Cris and Garland, Michael and Owens, John D},
  booktitle={Proceedings of the 28th ACM SIGPLAN Annual Symposium on Principles and Practice of Parallel Programming},
  pages={429--431},
  year={2023}
}

@article{zhu2024synthesize,
  title={How to synthesize text data without model collapse?},
  author={Zhu, Xuekai and Cheng, Daixuan and Li, Hengli and Zhang, Kaiyan and Hua, Ermo and Lv, Xingtai and Ding, Ning and Lin, Zhouhan and Zheng, Zilong and Zhou, Bowen},
  journal={arXiv preprint arXiv:2412.14689},
  year={2024}
}

@inproceedings{wangtilelang,
  title={TileLang: Bridge Programmability and Performance in Modern Neural Kernels},
  author={Wang, Lei and Cheng, Yu and Shi, Yining and Mo, Zhiwen and Tang, Zhengju and Xie, Wenhao and Wu, Tong and Ma, Lingxiao and Xia, Yuqing and Xue, Jilong and others},
  booktitle={The Fourteenth International Conference on Learning Representations},
  year={2026}
}

@inproceedings{demoura2008z3,
author = {De Moura, Leonardo and Bj\o{}rner, Nikolaj},
title = {Z3: an efficient SMT solver},
year = {2008},
isbn = {3540787992},
publisher = {Springer-Verlag},
address = {Berlin, Heidelberg},
booktitle = {Proceedings of the Theory and Practice of Software, 14th International Conference on Tools and Algorithms for the Construction and Analysis of Systems},
pages = {337–340},
numpages = {4},
location = {Budapest, Hungary},
series = {TACAS'08/ETAPS'08}
}

@article{lu2026corpusqa,
  title={CorpusQA: A 10 Million Token Benchmark for Corpus-Level Analysis and Reasoning},
  author={Lu, Zhiyuan and Li, Chenliang and Shi, Yingcheng and Shen, Weizhou and Yan, Ming and Huang, Fei},
  journal={arXiv preprint arXiv:2601.14952},
  year={2026}
}

@article{haas2025simpleqa,
  title={Simpleqa verified: A reliable factuality benchmark to measure parametric knowledge},
  author={Haas, Lukas and Yona, Gal and D'Antonio, Giovanni and Goldshtein, Sasha and Das, Dipanjan},
  journal={arXiv preprint arXiv:2509.07968},
  year={2025}
}

@inproceedings{luong-etal-2025-towards,
title = "Towards Robust Mathematical Reasoning",
author  = {Thang Luong and Dawsen Hwang and Hoang H. Nguyen and Golnaz Ghiasi and Yuri Chervonyi and Insuk Seo and Junsu Kim and Garrett Bingham and Jonathan Lee and Swaroop Mishra and Alex Zhai and Clara Huiyi Hu and Henryk Michalewski and Jimin Kim and Jeonghyun Ahn and Junhwi Bae and Xingyou Song and Trieu H. Trinh and Quoc V. Le and Junehyuk Jung},
booktitle = "Proceedings of the 2025 Conference on Empirical Methods in Natural Language Processing",
year = "2025",
url = "https://aclanthology.org/2025.emnlp-main.1794/",
}

@article{balunovic2025matharena,
  title = {MathArena: Evaluating LLMs on Uncontaminated Math Competitions},
  author = {Mislav Balunović and Jasper Dekoninck and Ivo Petrov and Nikola Jovanović and Martin Vechev},
  journal = {Proceedings of the Neural Information Processing Systems Track on Datasets and Benchmark},
  year={2025}
}

@misc{yang2025swesmith,
    title={SWE-smith: Scaling Data for Software Engineering Agents},
    author={John Yang and Kilian Lieret and Carlos E. Jimenez and Alexander Wettig and Kabir Khandpur and Yanzhe Zhang and Binyuan Hui and Ofir Press and Ludwig Schmidt and Diyi Yang},
    year={2025},
    eprint={2504.21798},
    archivePrefix={arXiv},
    primaryClass={cs.SE},
    url={https://arxiv.org/abs/2504.21798},
}

@article{wei2025browsecomp,
  title={Browsecomp: A simple yet challenging benchmark for browsing agents},
  author={Wei, Jason and Sun, Zhiqing and Papay, Spencer and McKinney, Scott and Han, Jeffrey and Fulford, Isa and Chung, Hyung Won and Passos, Alex Tachard and Fedus, William and Glaese, Amelia},
  journal={arXiv preprint arXiv:2504.12516},
  year={2025}
}

@article{li2025tool,
  title={The Tool Decathlon: Benchmarking Language Agents for Diverse, Realistic, and Long-Horizon Task Execution},
  author={Li, Junlong and Zhao, Wenshuo and Zhao, Jian and Zeng, Weihao and Wu, Haoze and Wang, Xiaochen and Ge, Rui and Cao, Yuxuan and Huang, Yuzhen and Liu, Wei and others},
  journal={arXiv preprint arXiv:2510.25726},
  year={2025}
}

@article{cheng2025facts,
  title={The FACTS Leaderboard: A Comprehensive Benchmark for Large Language Model Factuality},
  author={Cheng, Aileen and Jacovi, Alon and Globerson, Amir and Golan, Ben and Kwong, Charles and Alberti, Chris and Tao, Connie and Ben-David, Eyal and Tomar, Gaurav Singh and Haas, Lukas and others},
  journal={arXiv preprint arXiv:2512.10791},
  year={2025}
}

@article{du2025supergpqa,
  title={Supergpqa: Scaling llm evaluation across 285 graduate disciplines},
  author={Du, Xinrun and Yao, Yifan and Ma, Kaijing and Wang, Bingli and Zheng, Tianyu and Zhu, King and Liu, Minghao and Liang, Yiming and Jin, Xiaolong and Wei, Zhenlin and others},
  journal={arXiv preprint arXiv:2502.14739},
  year={2025}
}

@inproceedings{bai2025longbench,
  title={Longbench v2: Towards deeper understanding and reasoning on realistic long-context multitasks},
  author={Bai, Yushi and Tu, Shangqing and Zhang, Jiajie and Peng, Hao and Wang, Xiaozhi and Lv, Xin and Cao, Shulin and Xu, Jiazheng and Hou, Lei and Dong, Yuxiao and others},
  booktitle={Proceedings of the 63rd Annual Meeting of the Association for Computational Linguistics (Volume 1: Long Papers)},
  pages={3639--3664},
  year={2025}
}

@article{multiloko,
  author       = {Dieuwke Hupkes and
                  Nikolay Bogoychev},
  title        = {MultiLoKo: a multilingual local knowledge benchmark for LLMs spanning
                  31 languages},
  journal      = {CoRR},
  volume       = {abs/2504.10356},
  year         = {2025},
  url          = {https://doi.org/10.48550/arXiv.2504.10356},
  doi          = {10.48550/ARXIV.2504.10356},
}

@inproceedings{big_code_bench,
  author       = {Terry Yue Zhuo and
                  Minh Chien Vu and
                  Jenny Chim and
                  Han Hu and
                  Wenhao Yu and
                  Ratnadira Widyasari and
                  Imam Nur Bani Yusuf and
                  Haolan Zhan and
                  Junda He and
                  Indraneil Paul and
                  Simon Brunner and
                  Chen Gong and
                  James Hoang and
                  Armel Randy Zebaze and
                  Xiaoheng Hong and
                  Wen{-}Ding Li and
                  Jean Kaddour and
                  Ming Xu and
                  Zhihan Zhang and
                  Prateek Yadav and
                  et al.},
  title        = {BigCodeBench: Benchmarking Code Generation with Diverse Function Calls
                  and Complex Instructions},
  booktitle    = {The Thirteenth International Conference on Learning Representations,
                  {ICLR} 2025, Singapore, April 24-28, 2025},
  publisher    = {OpenReview.net},
  year         = {2025},
  url          = {https://openreview.net/forum?id=YrycTjllL0},
}

@misc{deng2025swebenchproaiagents,
      title={SWE-Bench Pro: Can AI Agents Solve Long-Horizon Software Engineering Tasks?}, 
      author={Xiang Deng and Jeff Da and Edwin Pan and Yannis Yiming He and Charles Ide and Kanak Garg and Niklas Lauffer and Andrew Park and Nitin Pasari and Chetan Rane and Karmini Sampath and Maya Krishnan and Srivatsa Kundurthy and Sean Hendryx and Zifan Wang and Vijay Bharadwaj and Jeff Holm and Raja Aluri and Chen Bo Calvin Zhang and Noah Jacobson and Bing Liu and Brad Kenstler},
      year={2025},
      eprint={2509.16941},
      archivePrefix={arXiv},
      primaryClass={cs.SE},
      url={https://arxiv.org/abs/2509.16941}, 
}

@article{comet,
      title={Comet: Fine-grained Computation-communication Overlapping for Mixture-of-Experts}, 
      author={Shulai Zhang and Ningxin Zheng and Haibin Lin and Ziheng Jiang and Wenlei Bao and Chengquan Jiang and Qi Hou and Weihao Cui and Size Zheng and Li-Wen Chang and Quan Chen and Xin Liu},
      year={2025},
      eprint={2502.19811},
      archivePrefix={arXiv},
      primaryClass={cs.DC},
      url={https://arxiv.org/abs/2502.19811}, 
}

@misc{ds3fs,
  author = {DeepSeek-AI},
  year = {2025},
  url = {https://github.com/deepseek-ai/3FS},
  title = {Fire-Flyer File System}
}

@inproceedings{erofs,
author = {Gao, Xiang and Dong, Mingkai and Miao, Xie and Du, Wei and Yu, Chao and Chen, Haibo},
title = {EROFS: a compression-friendly readonly file system for resource-scarce devices},
year = {2019},
isbn = {9781939133038},
publisher = {USENIX Association},
address = {USA},
booktitle = {Proceedings of the 2019 USENIX Conference on Usenix Annual Technical Conference},
pages = {149–162},
numpages = {14},
location = {Renton, WA, USA},
series = {USENIX ATC '19}
}

@inproceedings{firecracker,
author = {Agache, Alexandru and Brooker, Marc and Florescu, Andreea and Iordache, Alexandra and Liguori, Anthony and Neugebauer, Rolf and Piwonka, Phil and Popa, Diana-Maria},
title = {Firecracker: lightweight virtualization for serverless applications},
year = {2020},
isbn = {9781939133137},
publisher = {USENIX Association},
address = {USA},
booktitle = {Proceedings of the 17th Usenix Conference on Networked Systems Design and Implementation},
pages = {419–434},
numpages = {16},
location = {Santa Clara, CA, USA},
series = {NSDI'20}
}

@inproceedings{qemu,
author = {Bellard, Fabrice},
title = {QEMU, a fast and portable dynamic translator},
year = {2005},
publisher = {USENIX Association},
address = {USA},
booktitle = {Proceedings of the Annual Conference on USENIX Annual Technical Conference},
pages = {41},
numpages = {1},
location = {Anaheim, CA},
series = {ATEC '05}
}

@inproceedings {tvm,
    author = {Tianqi Chen and Thierry Moreau and Ziheng Jiang and Lianmin Zheng and Eddie Yan and Haichen Shen and Meghan Cowan and Leyuan Wang and Yuwei Hu and Luis Ceze and Carlos Guestrin and Arvind Krishnamurthy},
    title = {{TVM}: An Automated {End-to-End} Optimizing Compiler for Deep Learning},
    booktitle = {13th USENIX Symposium on Operating Systems Design and Implementation (OSDI 18)},
    year = {2018},
    isbn = {978-1-939133-08-3},
    address = {Carlsbad, CA},
    pages = {578--594},
    url = {https://www.usenix.org/conference/osdi18/presentation/chen},
    publisher = {USENIX Association},
    month = oct
}

@article{engram,
  author       = {Xin Cheng and
                  Wangding Zeng and
                  Damai Dai and
                  Qinyu Chen and
                  Bingxuan Wang and
                  Zhenda Xie and
                  Kezhao Huang and
                  Xingkai Yu and
                  Zhewen Hao and
                  Yukun Li and
                  Han Zhang and
                  Huishuai Zhang and
                  Dongyan Zhao and
                  Wenfeng Liang},
  title        = {Conditional Memory via Scalable Lookup: {A} New Axis of Sparsity for
                  Large Language Models},
  journal      = {CoRR},
  volume       = {abs/2601.07372},
  year         = {2026},
  url          = {https://doi.org/10.48550/arXiv.2601.07372},
  doi          = {10.48550/ARXIV.2601.07372},
  eprinttype   = {arXiv},
  eprint       = {2601.07372},
  bibsource    = {dblp computer science bibliography, https://dblp.org}
}

@inproceedings {overlaybd,
author = {Huiba Li and Yifan Yuan and Rui Du and Kai Ma and Lanzheng Liu and Windsor Hsu},
title = {{DADI}: {Block-Level} Image Service for Agile and Elastic Application Deployment},
booktitle = {2020 USENIX Annual Technical Conference (USENIX ATC 20)},
year = {2020},
isbn = {978-1-939133-14-4},
pages = {727--740},
url = {https://www.usenix.org/conference/atc20/presentation/li-huiba},
publisher = {USENIX Association},
month = jul
}

@misc{
bello*2017neural,
title={Neural Combinatorial Optimization with Reinforcement Learning},
author={Irwan Bello and Hieu Pham and Quoc V. Le and Mohammad Norouzi and Samy Bengio},
year={2017},
url={https://openreview.net/forum?id=rJY3vK9eg}
}

@article{gemma2,
  title={Gemma 2: Improving Open Language Models at a Practical Size},
  author={Gemma Team Morgane Riviere and Shreya Pathak and Pier Giuseppe Sessa and Cassidy Hardin and Surya Bhupatiraju and L'eonard Hussenot and Thomas Mesnard and Bobak Shahriari and Alexandre Ram'e and Johan Ferret and Peter Liu and Pouya Dehghani Tafti and Abe Friesen and Michelle Casbon and Sabela Ramos and Ravin Kumar and Charline Le Lan and Sammy Jerome and Anton Tsitsulin and Nino Vieillard and Piotr Stańczyk and Sertan Girgin and Nikola Momchev and Matt Hoffman and Shantanu Thakoor and Jean-Bastien Grill and Behnam Neyshabur and Alanna Walton and Aliaksei Severyn and Alicia Parrish and Aliya Ahmad and Allen Hutchison and Alvin Abdagic and Amanda Carl and Amy Shen and Andy Brock and Andy Coenen and Anthony Laforge and Antonia Paterson and Ben Bastian and Bilal Piot and Boxi Wu and Brandon Royal and Charlie Chen and Chintu Kumar and Chris Perry and Christoper A. Welty and Christopher A. Choquette-Choo and Danila Sinopalnikov and David Weinberger and Dimple Vijaykumar and Dominika Rogozi'nska and D. Herbison and Elisa Bandy and Emma Wang and Eric Noland and Erica Moreira and Evan Senter and Evgenii Eltyshev and Francesco Visin and Gabriel Rasskin and Gary Wei and Glenn Cameron and Gus Martins and Hadi Hashemi and Hanna Klimczak-Pluci'nska and Harleen Batra and Harsh Dhand and Ivan Nardini and Jacinda Mein and Jack Zhou and James Svensson and Jeff Stanway and Jetha Chan and Jin Zhou and Joana Carrasqueira and Joana Iljazi and Jocelyn Becker and Joe Fernandez and Joost R. van Amersfoort and Josh Gordon and Josh Lipschultz and Joshua Newlan and Junsong Ji and Kareem Mohamed and Kartikeya Badola and Kat Black and Katie Millican and Keelin McDonell and Kelvin Nguyen and Kiranbir Sodhia and Kish Greene and Lars Lowe Sjoesund and Lauren Usui and L. Sifre and Lena Heuermann and Leti-cia Lago and Lilly McNealus and Livio Baldini Soares and Logan Kilpatrick and Lucas Dixon and Luciano Luz Badini Martins and Machel Reid and Manvinder Singh and Mark Iverson and Martin Gorner and Mat Velloso and Mateo Wirth and Matt Davidow and Matt Miller and Matthew Rahtz and Matthew Watson and Meg Risdal and Mehran Kazemi and Michael Moynihan and Ming Zhang and Minsuk Kahng and Minwoo Park and Mofi Rahman and Mohit Khatwani and Natalie Dao and Nen-shad Bardoliwalla and Nesh Devanathan and Neta Dumai and Nilay Chauhan and Oscar Wahltinez and Pankil Botarda and Parker Barnes and Paul Barham and Paul Michel and Peng-chong Jin and Petko Georgiev and Phil Culliton and Pradeep Kuppala and Ramona Comanescu and Ramona Merhej and Reena Jana and Reza Ardeshir Rokni and Rishabh Agarwal and Ryan Mullins and Samaneh Saadat and Sara Marie Mc Carthy and Sarah Perrin and S{\'e}bastien M. R. Arnold and Se-bastian Krause and Shengyang Dai and Shruti Garg and Shruti Sheth and Sue Ronstrom and Susan Chan and Timothy Jordan and Ting Yu and Tom Eccles and Tom Hennigan and Tom{\'a}s Kocisk{\'y} and Tulsee Doshi and Vihan Jain and Vikas Yadav and Vilobh Meshram and Vishal Dharmadhikari and Warren Barkley and Wei Wei and Wenming Ye and Woohyun Han and Woosuk Kwon and Xiang Xu and Zhe Shen and Zhitao Gong and Zichuan Wei and Victor Cotruta and Phoebe Kirk and Anand Rao and Minh Giang and Ludovic Peran and Tris Warkentin and Eli Collins and Joelle Barral and Zoubin Ghahramani and Raia Hadsell and Daniel Sculley and Jeanine Banks and Anca Dragan and Slav Petrov and Oriol Vinyals and Jeffrey Dean and Demis Hassabis and Koray Kavukcuoglu and Cl{\'e}ment Farabet and Elena Buchatskaya and Sebastian Borgeaud and Noah Fiedel and Armand Joulin and Kathleen Kenealy and Robert Dadashi and Alek Andreev},
  journal={arXiv preprint arXiv:2408.00118},
  year={2024}
}

@inproceedings{minillm,
  title = {MiniLLM: Knowledge Distillation of Large Language Models},
  author = {Gu, Yuxian and Dong, Li and Wei, Furu and Huang, Minlie},
  booktitle={The Twelfth International Conference on Learning Representations},
  year = {2024}
}

\newpage
\appendix

\section*{Appendix}

\section{Author List and Acknowledgment}

\subsection{Author List}

Authors are listed alphabetically by their first name. 
Names marked with * denote individuals who have departed from our team. 

\noindent
\textbf{Research \& Engineering:} 
Anyi Xu,
Bangcai Lin,
Bing Xue,
Bingxuan Wang*,
Bingzheng Xu,
Bochao Wu,
Bowei Zhang,
Chaofan Lin,
Chen Dong,
Chengda Lu,
Chenggang Zhao,
Chengqi Deng,
Chenhao Xu,
Chenze Shao,
Chong Ruan*,
Conner Sun,
Damai Dai,
Daya Guo*,
Dejian Yang,
Deli Chen,
Donghao Li,
Erhang Li,
Fangyun Lin,
Fangzhou Yuan,
Feiyu Xia,
Fucong Dai,
Guangbo Hao,
Guanting Chen,
Guoai Cao,
Guolai Meng,
Guowei Li,
Han Yu,
Han Zhang,
Hanwei Xu,
Hao Li,
Haofen Liang,
Haoling Zhang,
Haoming Luo,
Haoran Wei*,
Haotian Yuan,
Haowei Zhang*,
Haowen Luo,
Haoyu Chen,
Haozhe Ji,
Honghui Ding,
Hongxuan Tang,
Huanqi Cao,
Huazuo Gao,
Hui Qu,
Hui Zeng,
J. Yang,
J.Q. Zhu,
Jia Yu,
Jialiang Huang,
Jiasheng Ye,
Jiashi Li,
Jiaxin Xu,
Jiewen Hu,
Jin Yan,
Jingchang Chen,
Jingli Zhou,
Jingting Xiang,
Jingyang Yuan,
Jingyuan Cheng,
Jinhua Zhu,
Jiping Yu,
Joseph Sun,
Jun Ran*,
Junguang Jiang,
Junjie Qiu,
Junlong Li*,
Junxiao Song,
Kai Dong,
Kaige Gao,
Kang Guan,
Kexing Zhou,
Kezhao Huang*,
Kuai Yu,
Lean Wang,
Lecong Zhang,
Lei Wang,
Li Zhang,
Liang Zhao,
Lihua Guo,
Lingxiao Luo,
Linwang Ma,
Litong Wang,
Liyu Cai,
Liyue Zhang,
Longhao Chen,
M.S. Di,
M.Y Xu,
Max Mei,
Mingchuan Zhang,
Minghua Zhang,
Minghui Tang,
Mingxu Zhou,
Panpan Huang,
Peixin Cong,
Peiyi Wang,
Qiancheng Wang,
Qihao Zhu,
Qingyang Li,
Qinyu Chen,
Qiushi Du,
Qiwei Jiang,
Rui Tian,
Ruifan Xu,
Ruijie Lu,
Ruiling Xu,
Ruiqi Ge,
Ruisong Zhang,
Ruizhe Pan,
Runji Wang,
Runqian Chen,
Runqiu Yin,
Runxin Xu,
Ruomeng Shen,
Ruoyu Zhang,
S.H. Liu,
Shanghao Lu,
Shangyan Zhou,
Shanhuang Chen,
Shaofei Cai,
Shaoheng Nie,
Shaoyuan Chen,
Shengding Hu,
Shengyu Liu,
Shiqiang Hu,
Shirong Ma,
Shiyu Wang,
Shuiping Yu,
Shunfeng Zhou,
Shuting Pan,
Shuying Yu,
Songyang Zhou,
Tao Ni,
Tao Yun,
Tian Jin,
Tian Pei,
Tian Ye,
Tianle Lin,
Tianran Ji,
Tianyi Cui,
Tianyuan Yue,
Tingting Yu,
Tun Wang,
W. Zhang,
Wangding Zeng,
Weilin Zhao,
Wen Liu,
Wenfeng Liang,
Wenjie Pang,
Wenjing Luo,
Wenjing Yao,
Wenjun Gao,
Wenkai Yang,
Wenlve Huang,
Wentao Zhang,
Wenting Ma,
Xi Gao,
Xiang He,
Xiangwen Wang,
Xiao Bi,
Xiaodong Liu,
Xiaohan Wang,
Xiaokang Chen,
Xiaokang Zhang,
Xiaotao Nie,
Xin Cheng,
Xin Liu,
Xin Xie,
Xingchao Liu,
Xingchen Liu,
Xingkai Yu,
Xingyou Li,
Xinyu Yang,
Xu Chen,
Xuanyu Wang,
Xuecheng Su,
Xuheng Lin,
Xuwei Fu,
Y.C. Yan,
Y.Q. Wang*,
Y.W. Ma,
Yanfeng Luo,
Yang Zhang,
Yanhong Xu,
Yanru Ma,
Yanwen Huang,
Yao Li,
Yao Li,
Yao Zhao,
Yaofeng Sun,
Yaohui Wang,
Yi Qian,
Yi Yu,
Yichao Zhang,
Yifan Ding,
Yifan Shi,
Yijia Wu,
Yiliang Xiong,
Ying He,
Ying Zhou,
Yingjia Luo,
Yinmin Zhong,
Yishi Piao,
Yisong Wang,
Yixiang Zhang,
Yixiao Chen,
Yixuan Tan,
Yixuan Wei,
Yiyang Ma,
Yiyuan Liu,
Yonglun Yang,
Yongqiang Guo,
Yongtong Wu,
Yu Wu,
Yuan Cheng,
Yuan Ou,
Yuanfan Xu,
Yuanhao Li,
Yuduan Wang,
Yuhan Wu,
Yuhao Meng,
Yuheng Zou,
YuKun Li,
Yunfan Xiong,
Yupeng Chen,
Yuqian Cao,
Yuqian Wang,
Yushun Zhang,
Yutong Lin,
Yuxian Gu,
Yuxiang Luo,
Yuxiang You,
Yuxuan Liu,
Yuxuan Zhou,
Yuyang Zhou,
Yuzhen Huang,
Z.F. Wu,
Zehao Wang,
Zehua Zhao,
Zehui Ren,
Zhangli Sha,
Zhe Fu,
Zhean Xu,
Zhenda Xie,
Zhengyan Zhang,
Zhewen Hao,
Zhibin Gou,
Zhicheng Ma,
Zhigang Yan,
Zhihong Shao,
Zhixian Huang,
Zhixuan Chen,
Zhiyu Wu,
Zhizhou Ren,
Zhuoshu Li,
Zhuping Zhang,
Zian Xu,
Zihao Wang,
Zihui Gu,
Zijia Zhu,
Zilin Li,
Zipeng Zhang*,
Ziwei Xie,
Ziyi Gao,
Zizheng Pan,
Zongqing Yao.

\noindent
\textbf{Business \& Compliance:}
Chenchen Ling,
Chengyu Hou,
Dongjie Ji,
Fang Wei,
Hengqing Zhang,
Jia Luo,
Jia Song,
Jialu Cai,
Jian Liang,
Jiangting Zhou,
Jieyu Yang,
Jin Chen,
Jingzi Zhou,
Junmin Zheng,
Leyi Xia,
Linyan Zhu,
Miaojun Wang,
Mingming Li,
Minmin Han,
Ning Wang,
Panpan Wang,
Peng Zhang,
Ruyi Chen,
Shangmian Sun,
Shaoqing Wu,
W.L. Xiao,
Wei An,
Wenqing Hou,
Xianzu Wang,
Xiaowen Sun,
Xiaoxiang Wang,
Xinyu Zhang,
Xueyin Chen,
Yao Xu,
Yi Shao,
Yiling Ma,
Ying Tang,
Yuehan Yang,
Yuer Xu,
Yukun Zha,
Yuping Lin,
Yuting Yan,
Zekai Zhang,
Zhe Ju,
Zheren Gao,
Zhongyu Wu,
Zihua Qu,
Ziyi Wan.

\subsection{Acknowledgment}

We would like to thank \href{https://www.zhihu.com/people/toyama}{Dolly Deng} and other testers for their valuable suggestions and feedback regarding the capabilities of \dsviv{} series models.

\section{Evaluation Details}
\begin{table}[htbp]
\centering
\caption{Agentic Search vs. Retrieval Augmented Search for \dsvivp{}.}
\label{tab:agentic_search}
\small
\setlength{\tabcolsep}{4pt}
\begin{tabular}{l l c | c c r rrr}
\toprule
 \textbf{Difficulty} & \textbf{Category} & \textbf{\#}
  & \textbf{Agent Win} & \textbf{RAG Win} & \textbf{Tie}
  & \textbf{Agent\%} & \textbf{RAG\%} & \textbf{Tie\%} \\
\midrule

% ===== V4 Retrieval Augmented Search =====

   \multirow{2}{*}{Easy} & Objective Q\&A (客观问答)  & 196 & 110 &  43 &  43 & 56.1 & 21.9 & 21.9 \\
                          & Subjective Q\&A (主观问答) & 321 & 198 &  56 &  67 & 61.7 & 17.4 & 20.9 \\
\cmidrule(lr){1-9}
   \multirow{2}{*}{Hard} & Objective Q\&A (客观问答)  & 168 & 102 &  33 &  33 & 60.7 & 19.6 & 19.6 \\
                          & Subjective Q\&A (主观问答) & 184 & 126 &  27 &  31 & 68.5 & 14.7 & 16.8 \\
\cmidrule(lr){1-9}
   \multicolumn{2}{c}{\textbf{Total (总计)~~~~~~~~}} & \textbf{869} & \textbf{536} & \textbf{159} & \textbf{174} & \textbf{61.7} & \textbf{18.3} & \textbf{20.0} \\
\bottomrule
\end{tabular}
\end{table}

\begin{table}[htbp]
\centering
\caption{Cost Comparison:Agentic Search vs. Retrieval Augmented Search (Mean) for \dsvivp{}. Most of the tool calls are parallel for Agentic Search.}
\label{tab:cost_comparison}
\small
\begin{tabular}{l rrr}
\toprule
\textbf{Version} & \textbf{Tool Calls} & \textbf{Prefill (tokens)} & \textbf{Output (tokens)} \\
\midrule
V4 Agentic Search & 16.2 & 13649 & 1526 \\
V4 Retrieval Augmented Search   &  --- & 10453 & 1308 \\
\bottomrule
\end{tabular}
\end{table}
\begin{table}[htbp]
\centering
\caption{Comparative Evaluation of \dsvivp{} and \dsviiiII{} on Search Q\&A Tasks.}
\label{tab:search_qa}
\small
\setlength{\tabcolsep}{4pt}
\begin{tabular}{ll r | rrr rrr}
\toprule
& & & \multicolumn{6}{c}{\textbf{Internal Evaluation (内部综合评估)}} \\
\cmidrule(lr){4-9}
\textbf{Category} & \textbf{Subcategory} & \textbf{\#}
  & \textbf{V4 win} & \textbf{V3.2 win} & \textbf{tie}
  & \textbf{V4\%} & \textbf{V3.2\%} & \textbf{tie\%} \\
\midrule

% ===== Objective Q&A =====
\multirow{4}{*}{\shortstack[l]{\textbf{Objective}\\\textbf{Q\&A}\\\textbf{(客观问答)}}}
  & Single-value Search (单值信息查找)   &  95 & 36 & 10 & 49 & 37.9 & 10.5 & 51.6 \\
  & Entity Search (实体信息查找)         &  99 & 24 &  7 & 68 & 24.2 &  7.1 & 68.7 \\
  & Enumerative Search (枚举型信息查找)  &  95 & 19 &  8 & 68 & 20.0 &  8.4 & 71.6 \\
\cmidrule(lr){2-9}
  & \textbf{Subtotal (小计)} & \textbf{289} & \textbf{79} & \textbf{25} & \textbf{185} & \textbf{27.3} & \textbf{8.7} & \textbf{64.0} \\
\midrule

% ===== Subjective Q&A =====
\multirow{8}{*}{\shortstack[l]{\textbf{Subjective}\\\textbf{Q\&A}\\\textbf{(主观问答)}}}
  & Causal Analysis (原因分析)           & 100 & 28 &  5 & 67 & 28.0 &  5.0 & 67.0 \\
  & Comparison (对比)                    &  96 & 28 & 20 & 48 & 29.2 & 20.8 & 50.0 \\
  & Advice Seeking (寻求建议)            &  92 & 23 &  8 & 61 & 25.0 &  8.7 & 66.3 \\
  & Recommendation (推荐)               &  95 & 26 & 19 & 50 & 27.4 & 20.0 & 52.6 \\
  & Planning \& Strategy (攻略计划)      &  92 & 32 & 11 & 49 & 34.8 & 12.0 & 53.3 \\
  & Opinion \& Evaluation (评价看法)     &  96 & 30 &  8 & 58 & 31.2 &  8.3 & 60.4 \\
  & Trend Analysis (趋势分析)            &  96 & 23 &  3 & 70 & 24.0 &  3.1 & 72.9 \\
\cmidrule(lr){2-9}
  & \textbf{Subtotal (小计)} & \textbf{667} & \textbf{190} & \textbf{74} & \textbf{403} & \textbf{28.5} & \textbf{11.1} & \textbf{60.4} \\
\midrule

% ===== TOTAL =====
\multicolumn{2}{c}{\textbf{TOTAL (总计)}} & \textbf{956} & \textbf{269} & \textbf{99} & \textbf{588} & \textbf{28.1} & \textbf{10.4} & \textbf{61.5} \\
\bottomrule
\end{tabular}
\end{table}

\begin{table}[t]
\centering
\caption{Comparative Analysis of \dsvivp{} and Gemini-3.1-Pro in Chinese Functional Writing.}
\label{tab:funceval_conclusion}
\small
\setlength{\tabcolsep}{4pt}
\begin{tabular}{ll r | rrr rrr}
\toprule
& & & \multicolumn{6}{c}{\textbf{Internal Evaluation (内部综合评估)}} \\
\cmidrule(lr){4-9}
\textbf{Category} & \textbf{Subcategory} & \textbf{\#}
  & \textbf{DS win} & \textbf{Gem win} & \textbf{Tie}
  & \textbf{DS\%} & \textbf{Gem\%} & \textbf{Tie\%} \\
\midrule

% ===== Business Writing =====
\multirow{10}{*}{\shortstack[l]{\textbf{Business}\\\textbf{Writing}\\\textbf{(办公文本)}}}
  & Report (报告)                   & 527 & 350 & 162 & 15 & 66.41 & 30.74 & 2.85 \\
  & Proposal (方案策划)             & 291 & 181 & 103 &  7 & 62.20 & 35.40 & 2.41 \\
  & Education (教育培训)            & 159 & 100 &  56 &  3 & 62.89 & 35.22 & 1.89 \\
  & Email \& Letter (邮件书信)      & 146 & 107 &  37 &  2 & 73.29 & 25.34 & 1.37 \\
  & Notice (通知公告)               &  72 &  43 &  24 &  5 & 59.72 & 33.33 & 6.94 \\
  & Professional (专业文本)         &  63 &  34 &  27 &  2 & 53.97 & 42.86 & 3.17 \\
  & Recruitment (招聘求职)          &  42 &  27 &  15 &  0 & 64.29 & 35.71 & 0.00 \\
  & Technical (技术文本)            &  29 &  22 &   7 &  0 & 75.86 & 24.14 & 0.00 \\
  & Review (介绍评价)               &  20 &  15 &   5 &  0 & 75.00 & 25.00 & 0.00 \\
\cmidrule(lr){2-9}
  & \textbf{Subtotal (小计)} & \textbf{1349} & \textbf{879} & \textbf{436} & \textbf{34} & \textbf{65.16} & \textbf{32.32} & \textbf{2.52} \\
\midrule

% ===== Media Writing =====
\multirow{9}{*}{\shortstack[l]{\textbf{Media}\\\textbf{Writing}\\\textbf{(媒体文本)}}}
  & Social Media (社交媒体文案)     & 267 & 156 & 101 & 10 & 58.43 & 37.83 & 3.75 \\
  & Ad Copy (广告商品文案)          & 214 & 109 &  98 &  7 & 50.93 & 45.79 & 3.27 \\
  & Long-form Content (内容平台长文)&  99 &  71 &  25 &  3 & 71.72 & 25.25 & 3.03 \\
  & News Report (新闻报道)          &  51 &  27 &  22 &  2 & 52.94 & 43.14 & 3.92 \\
  & Advertorial (营销软文)          &  17 &  12 &   4 &  1 & 70.59 & 23.53 & 5.88 \\
  & Headline (标题)                 &  11 &   7 &   4 &  0 & 63.64 & 36.36 & 0.00 \\
  & Narration Script (口播文案)     &   4 &   2 &   1 &  1 & 50.00 & 25.00 & 25.00 \\
  & Comment (评论)                  &   3 &   2 &   1 &  0 & 66.67 & 33.33 & 0.00 \\
\cmidrule(lr){2-9}
  & \textbf{Subtotal (小计)} & \textbf{666} & \textbf{386} & \textbf{256} & \textbf{24} & \textbf{57.96} & \textbf{38.44} & \textbf{3.60} \\
\midrule

% ===== Everyday Writing =====
\multirow{6}{*}{\shortstack[l]{\textbf{Everyday}\\\textbf{Writing}\\\textbf{(生活文本)}}}
  & Congratulatory (祝贺文本)       & 101 &  54 &  41 &  6 & 53.47 & 40.59 & 5.94 \\
  & Communication (沟通回复)        & 100 &  71 &  26 &  3 & 71.00 & 26.00 & 3.00 \\
  & Reflection (心得感想)           &  90 &  68 &  17 &  5 & 75.56 & 18.89 & 5.56 \\
  & Review (介绍评价)               &  55 &  44 &   9 &  2 & 80.00 & 16.36 & 3.64 \\
  & Comment (评论)                  &  44 &  34 &   8 &  2 & 77.27 & 18.18 & 4.55 \\
\cmidrule(lr){2-9}
  & \textbf{Subtotal (小计)} & \textbf{390} & \textbf{271} & \textbf{101} & \textbf{18} & \textbf{69.49} & \textbf{25.90} & \textbf{4.62} \\
\midrule

% ===== Oral Writing =====
\multirow{6}{*}{\shortstack[l]{\textbf{Oral}\\\textbf{Writing}\\\textbf{(口头文本)}}}
  & Speech (发言稿)                 & 226 & 135 &  85 &  6 & 59.73 & 37.61 & 2.65 \\
  & Narration Script (口播文案)     &  51 &  25 &  23 &  3 & 49.02 & 45.10 & 5.88 \\
  & Sales Script (话术)             &  31 &  22 &   6 &  3 & 70.97 & 19.35 & 9.68 \\
  & Dialogue (对话文本)             &  10 &   4 &   6 &  0 & 40.00 & 60.00 & 0.00 \\
  & Congratulatory (祝贺文本)       &   1 &   1 &   0 &  0 & 100.00 & 0.00 & 0.00 \\
\cmidrule(lr){2-9}
  & \textbf{Subtotal (小计)} & \textbf{319} & \textbf{187} & \textbf{120} & \textbf{12} & \textbf{58.62} & \textbf{37.62} & \textbf{3.76} \\
\midrule

% ===== Official Document =====
\multirow{6}{*}{\shortstack[l]{\textbf{Official}\\\textbf{Document}\\\textbf{(公文文本)}}}
  & Administrative Doc (事务文书)   & 117 &  60 &  53 &  4 & 51.28 & 45.30 & 3.42 \\
  & Personal Doc (个人文书)         &  73 &  45 &  27 &  1 & 61.64 & 36.99 & 1.37 \\
  & Government Doc (行政公文)       &  34 &  19 &  14 &  1 & 55.88 & 41.18 & 2.94 \\
  & Speech (发言稿)                 &   3 &   1 &   2 &  0 & 33.33 & 66.67 & 0.00 \\
  & Essay Writing (申论写作)        &   3 &   1 &   1 &  1 & 33.33 & 33.33 & 33.33 \\
\cmidrule(lr){2-9}
  & \textbf{Subtotal (小计)} & \textbf{230} & \textbf{126} & \textbf{97} & \textbf{7} & \textbf{54.78} & \textbf{42.17} & \textbf{3.04} \\
\midrule

% ===== Academic Writing =====
\multirow{5}{*}{\shortstack[l]{\textbf{Academic}\\\textbf{Writing}\\\textbf{(学术文本)}}}
  & Research Paper (学术论文)       & 104 &  67 &  32 &  5 & 64.42 & 30.77 & 4.81 \\
  & Coursework (课程作业)           &  90 &  53 &  35 &  2 & 58.89 & 38.89 & 2.22 \\
  & Academic Support (学术辅助)     &  15 &  11 &   3 &  1 & 73.33 & 20.00 & 6.67 \\
  & Science Outreach (专业科普)     &   7 &   6 &   1 &  0 & 85.71 & 14.29 & 0.00 \\
\cmidrule(lr){2-9}
  & \textbf{Subtotal (小计)} & \textbf{216} & \textbf{137} & \textbf{71} & \textbf{8} & \textbf{63.43} & \textbf{32.87} & \textbf{3.70} \\
\midrule
\multicolumn{2}{l}{\textbf{Total (总计)}} & \textbf{3170} & \textbf{1986} & \textbf{1081} & \textbf{103} & \textbf{62.65} & \textbf{34.10} & \textbf{3.25} \\

\bottomrule
\end{tabular}
\end{table}

\begin{table}[t]
\centering
\caption{
Comparative Analysis of \dsvivp{} and Gemini-3.1-Pro in Chinese Creative Writing.}
\label{tab:creative_writing}
\footnotesize
\setlength{\tabcolsep}{2.5pt}
\begin{tabular}{l r | rrr rrr | rrr rrr}
\toprule
& & \multicolumn{6}{c|}{\textbf{Instruction Following(指令遵循)}}
  & \multicolumn{6}{c}{\textbf{Writing Quality (写作质量)}} \\
\cmidrule(lr){3-8} \cmidrule(lr){9-14}
\textbf{Subcategory (文体)} & \textbf{\#}
  & \textbf{DS} & \textbf{Gem} & \textbf{Tie}
  & \textbf{DS\%} & \textbf{Gem\%} & \textbf{Tie\%}
  & \textbf{DS} & \textbf{Gem} & \textbf{Tie}
  & \textbf{DS\%} & \textbf{Gem\%} & \textbf{Tie\%} \\
\midrule

Fiction (小说故事)                & 836 & 504 & 323 &  5 & 60.58 & 38.82 & 0.60  & 672 & 157 &  3 & 80.77 & 18.87 & 0.36 \\
General Fiction (泛小说故事)      & 662 & 368 & 290 &  3 & 55.67 & 43.87 & 0.45  & 467 & 194 &  0 & 70.65 & 29.35 & 0.00 \\
Fan Fiction (同人文)              & 410 & 253 & 150 &  3 & 62.32 & 36.95 & 0.74  & 338 &  67 &  1 & 83.25 & 16.50 & 0.25 \\
General Fan Fic. (泛同人文)      & 202 & 111 &  90 &  1 & 54.95 & 44.55 & 0.50  & 161 &  40 &  1 & 79.70 & 19.80 & 0.50 \\
Narrative (记叙文)                & 171 & 115 &  54 &  2 & 67.25 & 31.58 & 1.17  & 141 &  30 &  0 & 82.46 & 17.54 & 0.00 \\
General Prose (泛散文)            & 124 &  83 &  40 &  1 & 66.94 & 32.26 & 0.81  &  88 &  36 &  0 & 70.97 & 29.03 & 0.00 \\
Prose (散文)                      & 112 &  74 &  38 &  0 & 66.07 & 33.93 & 0.00  &  92 &  20 &  0 & 82.14 & 17.86 & 0.00 \\
Writing Style (文笔)              & 112 &  81 &  31 &  0 & 72.32 & 27.68 & 0.00  &  86 &  26 &  0 & 76.79 & 23.21 & 0.00 \\
Classical Poetry (古诗文)         &  48 &  24 &  24 &  0 & 50.00 & 50.00 & 0.00  &  39 &   9 &  0 & 81.25 & 18.75 & 0.00 \\
Modern Poetry (现代诗)            &  43 &  23 &  20 &  0 & 53.49 & 46.51 & 0.00  &  32 &  11 &  0 & 74.42 & 25.58 & 0.00 \\
Lyrics (歌词)                     &  30 &   8 &  22 &  0 & 26.67 & 73.33 & 0.00  &  16 &  14 &  0 & 53.33 & 46.67 & 0.00 \\
Literary Appreciation (赏析)               &  27 &  20 &   7 &  0 & 74.07 & 25.93 & 0.00  &  18 &   9 &  0 & 66.67 & 33.33 & 0.00 \\
General Argument. (泛议论文)     &  24 &  15 &   9 &  0 & 62.50 & 37.50 & 0.00  &  17 &   7 &  0 & 70.83 & 29.17 & 0.00 \\
General Narrative (泛记叙文)      &  23 &  11 &  12 &  0 & 47.83 & 52.17 & 0.00  &  15 &   8 &  0 & 65.22 & 34.78 & 0.00 \\
General Classical (泛古文诗歌)    &   9 &   5 &   4 &  0 & 55.56 & 44.44 & 0.00  &   5 &   4 &  0 & 55.56 & 44.44 & 0.00 \\
Creative Writing (创意写作)       &   6 &   2 &   4 &  0 & 33.33 & 66.67 & 0.00  &   4 &   2 &  0 & 66.67 & 33.33 & 0.00 \\
Argumentative (议论文)            &   5 &   5 &   0 &  0 & 100.00 & 0.00 & 0.00  &   5 &   0 &  0 & 100.00 & 0.00 & 0.00 \\
General Mod. Poetry (泛现代诗)   &   2 &   1 &   1 &  0 & 50.00 & 50.00 & 0.00  &   2 &   0 &  0 & 100.00 & 0.00 & 0.00 \\

\midrule
\textbf{Total (总计)} & \textbf{2837} & \textbf{1703} & \textbf{1119} & \textbf{15} & \textbf{60.03} & \textbf{39.44} & \textbf{0.53} & \textbf{2198} & \textbf{634} & \textbf{5} & \textbf{77.48} & \textbf{22.35} & \textbf{0.18} \\

\bottomrule
\end{tabular}
\end{table}

\begin{table}[htbp]
\centering
\caption{\dsvivp{} vs. Claude-Opus-4.5 on Complex Instruction Following and Multi-Turn Writing.}
\label{tab:complex_multiturn}
\small
\setlength{\tabcolsep}{4pt}
\begin{tabular}{l r | rrr rrr}
\toprule
& & \multicolumn{6}{c}{\textbf{Internal Evaluation (内部综合评估)}} \\
\cmidrule(lr){3-8}
\textbf{Category} & \textbf{\#}
  & \textbf{DS} & \textbf{Opus} & \textbf{Tie}
  & \textbf{DS\%} & \textbf{Opus\%} & \textbf{Tie\%} \\
\midrule

Complex Inst. Following (复杂指令跟随) & 49 & 23 & 26 & 0 & 46.9\% & 53.1\% & 0.0\% \\
Multi-Turn Writing (多轮写作) & 147 & 67 & 76 & 4 & 45.6\% & 51.7\% & 2.7\% \\

\midrule

\textbf{Total (总计)} & \textbf{196} & \textbf{90} & \textbf{102} & \textbf{4} & \textbf{45.9\%} & \textbf{52.0\%} & \textbf{2.0\%} \\

\bottomrule
\end{tabular}
\end{table}

\begin{figure}[htbp]
    \centering
    \includegraphics[width=\linewidth]{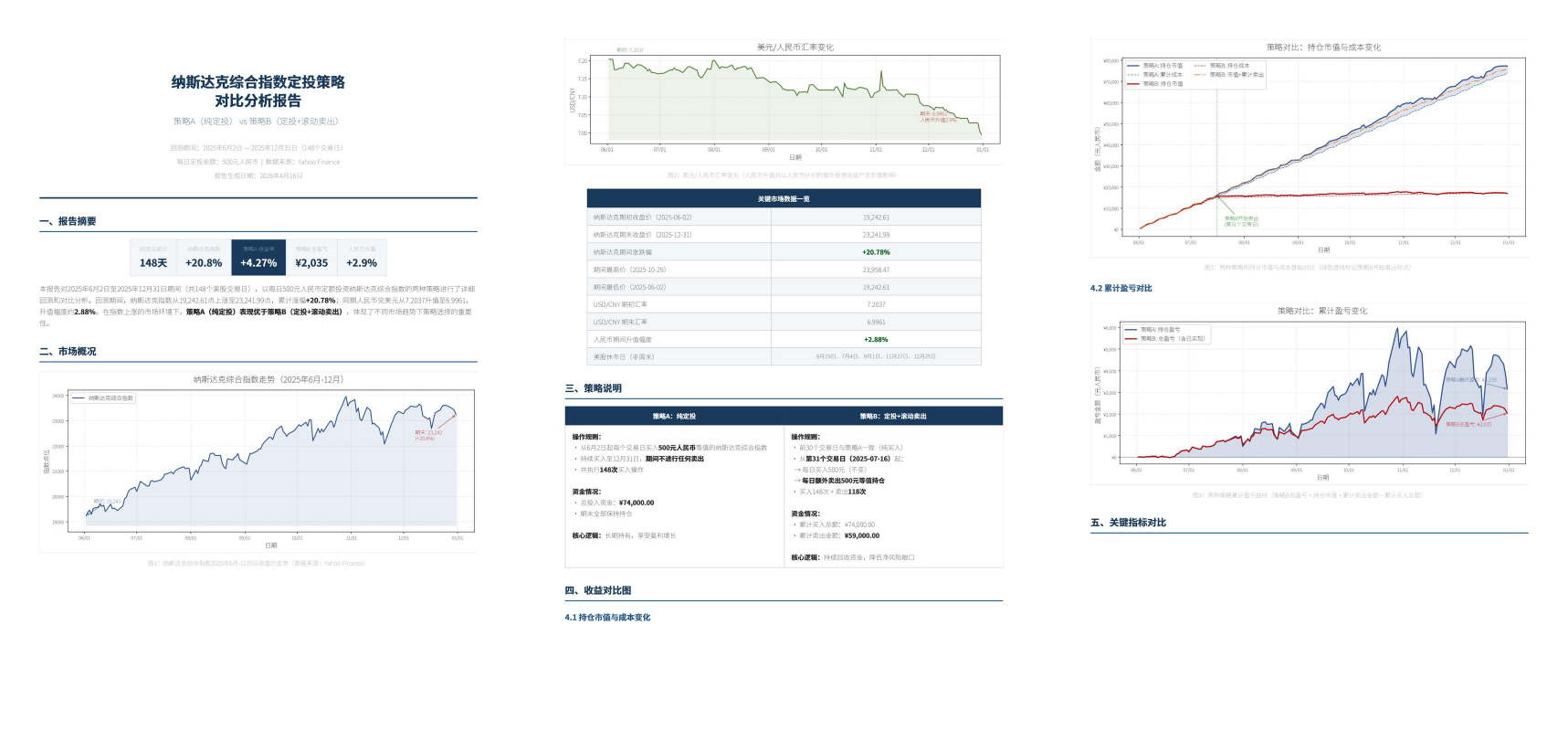} 
    \caption{Example output of a task that requires comparing two regular investment strategies for the NASDAQ.}
    \label{fig:case-nasdaq}
\end{figure}

\begin{figure}[htbp]
    \centering
    \includegraphics[width=\linewidth]{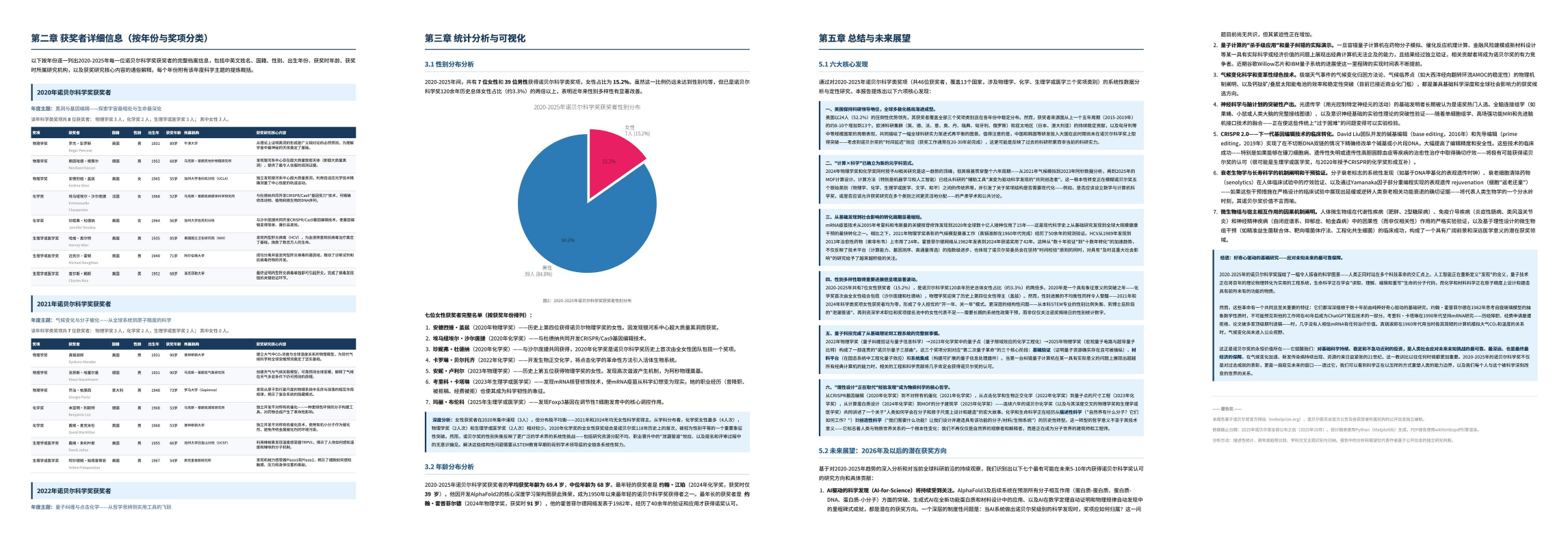} 
    \caption{Example output of a task which requires researching 2020-2025 Nobel Science Prizes and generating an analytical PDF report.}
    \label{fig:case-nobel}
\end{figure}

\end{CJK*}
\end{document}